%% file: combined.tex
\documentclass[10pt,twocolumn,letterpaper]{article}

\usepackage{wacv}              

\usepackage{graphicx}
\usepackage{amsmath}
\usepackage{amssymb}
\usepackage{booktabs}
\usepackage[accsupp]{axessibility}

\input{preamble}

\usepackage[pagebackref,breaklinks,colorlinks]{hyperref}

\usepackage[capitalize]{cleveref}
\crefname{section}{Sec.}{Secs.}
\Crefname{section}{Section}{Sections}
\Crefname{table}{Table}{Tables}
\crefname{table}{Tab.}{Tabs.}

\def\wacvPaperID{1905} 
\def\confName{WACV}
\def\confYear{2025}

\begin{document}

\input{PaperForReviewContent.tex}

\clearpage

\input{SupplementaryContent.tex}

\end{document}

%% file: preamble.tex
\usepackage[dvipsnames]{xcolor}

\usepackage{graphicx}
\usepackage{amsmath}
\usepackage{amssymb}
\usepackage{booktabs}
\usepackage{times}
\usepackage{epsfig}
\usepackage{color}
\usepackage[dvipsnames]{xcolor}
\usepackage{url}
\usepackage[acronym]{glossaries}
\usepackage{comment}
\usepackage{adjustbox}

\usepackage{cite}     %
\usepackage{xspace}

\makeglossaries{}
\newacronym{bop}{BoP}{Benchmark for 6D Object Pose Estimation}
\newacronym{sota}{SotA}{State-of-the-Art}
\newacronym{fig}{Fig.}{Figure}
\newacronym{tab}{Tab.}{Table}
\newacronym{vsd}{VSD}{Visible Surface Discrepancy}
\newacronym{mssd}{MSSD}{Maximum Symmetry-Aware Surface Distance}
\newacronym{mspd}{MSPD}{Maximum Symmetry-Aware Projection Distance}
\newacronym{sdf}{SDF}{Signed Distance Function}
\newacronym{dof}{DoF}{Degrees-of-Freedom}
\newacronym{icp}{ICP}{Iterative Closest Point}
\newacronym{mvs}{MVS}{Multi-View Stereo}
\newacronym{sfm}{SfM}{Structure-from-Motion}

\newcommand{\realcap}{Reality Capture}
\newcommand{\colmap}{Colmap}

\newcommand{\fig}{Fig.}

\newcommand{\PAR}[1]{\vskip4pt \noindent{\bf #1~}}

\makeatletter
\DeclareRobustCommand\onedot{\futurelet\@let@token\@onedot}
\def\@onedot{\ifx\@let@token.\else.\null\fi\xspace}

\def\eg{\emph{e.g}\onedot}
\def\ie{\emph{i.e}\onedot}

%% file: PaperForReviewContent.tex
\title{Comparative Evaluation of 3D Reconstruction Methods for Object Pose Estimation}

\author{
Varun Burde\thanks{Equal Contribution.} \qquad
Assia Benbihi$^*$ %
\qquad Pavel Burget %
\qquad Torsten Sattler \\
Czech Institute of Informatics, Robotics and Cybernetics, Czech Technical University in Prague\\
{\tt\small firstname.lastname@cvut.cz}
}
\maketitle

\begin{abstract}
Current generalizable object pose estimators, i.e., approaches that do not need to be trained per object, rely on accurate 3D models.
Predominantly, CAD models are used, which can be hard to obtain in practice.
At the same time, it is often possible to acquire images of an object.
Naturally, this leads to the question of whether 3D models reconstructed from images are sufficient to facilitate accurate object pose estimation.
We aim to answer this question by proposing a novel benchmark for measuring the impact of 3D reconstruction quality on pose estimation accuracy.
Our benchmark provides calibrated images suitable for reconstruction and registered with the test images of the YCB-V dataset for pose evaluation under the BOP benchmark format.
Detailed experiments with multiple state-of-the-art 3D reconstruction and object pose estimation approaches show that the geometry produced by modern reconstruction methods is often sufficient for accurate pose estimation.
Our experiments lead to interesting observations: (1) Standard metrics for measuring 3D reconstruction quality are not necessarily indicative of pose estimation accuracy, which shows the need for dedicated benchmarks such as ours. (2) Classical, non-learning-based approaches can perform on par with modern learning-based reconstruction techniques and can even offer a better reconstruction time-pose accuracy tradeoff. (3) There is still a sizable gap between performance with reconstructed and with CAD models. To foster research on closing this gap, the benchmark is made available at \url{https://github.com/VarunBurde/reconstruction_pose_benchmark}.
\end{abstract}

\section{Introduction}
\label{sec:intro}

Object pose estimators predict the position and orientation of an object in a given image and is a prerequisite to various applications including robotic manipulation and 
Augmented Reality (AR). 
Existing methods can be categorized into CAD-based and CAD-free methods based on whether the pose estimation involves a CAD model, \ie, a 3D model, of the object.
%
%
%
CAD-based methods usually lead the BOP benchmark for object pose estimation~\cite{hodan2018bop,hodavn2020bop,bopbenchmark} but their deployment is limited by their reliance on CAD models.
Often, such 3D models are generated by CAD applications in complex setups or reconstructed via range scanners~\cite{downs2022google} that produce highly accurate meshes.
Naturally, this process can be time-consuming.
Also, in many practical settings, \eg, service robots, it is infeasible to obtain such 3D models for each object class and object instance that might be encountered.
A more practical approach in such scenarios would be to reconstruct 3D object models on the fly from images. 

Generalization is another factor limiting the deployment of pose estimators: in the example where a robot faces a new object and generates a 3D model on the fly, most of the estimators still require training or finetuning on the new object.
This induces a computational overhead not only because of the network training but also because the robot must capture enough views for the training to converge.
Instead, deploying CAD-based estimators that generalize to unseen objects~\cite{labbe2022megapose,foundationposewen2024,ornek2023foundpose,caraffa2025freeze} is more efficient since they can be used without finetuning and can run with 3D reconstructions that work even with sparse views.

In this paper, we investigate using 3D reconstructions from RGB images for object pose estimation.
In particular, we want to quantify the object pose accuracy gap between using high-quality CAD models and 3D models obtained from images inside
generalizable object pose estimators~\cite{labbe2022megapose,foundationposewen2024}.
For this purpose, we propose a new benchmark.
Compared to existing benchmarks for 3D reconstruction~\cite{ETH3D_bench,knapitsch2017tanks,jensen2014large,tola2012efficient}, our benchmark does not treat 3D reconstruction as a task unto itself but rather evaluates the resulting 3D models inside a higher-level task (namely object pose estimation). 
Thus, we mainly \textit{measure 3D reconstruction performance by the accuracy of the estimated poses} rather than the accuracy of the resulting 3D model itself.
Results show there is no strong correlation between the two performances, \ie, one method can generate more accurate 3D models than another yet lead to less accurate pose estimates, demonstrating the need for the proposed benchmark.

Our benchmark is based on the widely adopted YCB-V~\cite{xiang2018posecnn} dataset for object pose evaluation, which consists of 21 objects selected from the YCB dataset~\cite{calli2015ycb}, including small objects, objects with low texture, complex shapes, high reflectance, and multiple symmetries. 
For each object, our benchmark consists of two components: 
\textbf{(1)}
A new set of object images captured to be suitable for 3D reconstruction, \ie, high resolution, object-focused, and occlusion-free images contrary to the YCB-V images.
\textbf{(2)} A set of test images for object pose estimation together with ground truth poses registered with the image sets from (1). 
Compared to the data for the first component, which we captured ourselves, we use the test images provided by the YCB-V~\cite{xiang2018posecnn} dataset for the second component.
We believe that registering the proposed novel reconstruction benchmark with an existing pose estimation dataset that is well-known to the community will make the adoption of our benchmark easier compared to also building an object pose estimation benchmark.
Given the two components, we evaluate multiple state-of-the-art 3D reconstruction methods, including both classical methods based on multi-view stereo~\cite{schonberger2016pixelwise,capturereality} and recent learning-based approaches using neural implicit representations~\cite{yariv2021volume,yu2022monosdf,yariv2023bakedsdf,oechsle2021unisurf,wang2021neus}.
Our experiments show that standard metrics for 3D reconstruction quality (both geometry-based and appearance-based) do not necessarily directly correlate with object pose estimation accuracy. 
This clearly shows a need for novel benchmarks such as ours that directly measure the impact of 3D reconstruction algorithms on object pose estimation's accuracy.  

In summary, this paper makes the following contributions: \textbf{(1)} We propose a novel benchmark to evaluate 3D reconstruction algorithms in the context of object pose estimation. 
Our benchmark aims to measure to what degree 3D models reconstructed from images instead of CAD models can be used for object pose estimation. 
\textbf{(2)} We evaluate a wide range of state-of-the-art 3D reconstruction algorithms on our benchmark in terms of object pose accuracy, data requirements to reach high accuracy and reconstruction times. 
We draw several insights that are very relevant to the reconstruction and pose estimation research communities, including: 
(a) There is a clear need for benchmarks that evaluate 3D reconstruction algorithms in the context of object pose estimation. 
(b) Classical, non-learning-based approaches can perform similar to modern learning-based reconstruction techniques while offering faster reconstruction times. 
(c) The top-performing reconstruction methods are the same for the two object pose estimators considered in our benchmark~\cite{labbe2022megapose,foundationposewen2024}, suggesting that all object pose estimators might potentially benefit from progress in the area of 3D reconstruction. 
(d) While it is possible to obtain a similar performance as with CAD models for some types of objects, there is still a sizable gap in performance for other types of objects, especially for objects with fine details and reflective surfaces. 
Clearly, the problem of 3D object reconstruction is not yet solved. 
\textbf{(3)} The benchmark data is available online
and includes calibrated images suitable for 3D reconstruction and 
the 3D reconstructed models.

\section{Related Work}
\label{sec:related_work}

\PAR{Object Pose Estimation} is addressed with multiple solutions including
feature matching~\cite{drost2010model,hinterstoisser2016going,collet2009object, martinez2010moped,rothganger20063d,manuelli2019kpam,sun2022onepose,ornek2023foundpose}, template matching~\cite{hinterstoisser2011gradient,wohlhart2015learning,hinterstoisser2013model,hinterstoisser2011multimodal},
pose regression~\cite{xiang2018posecnn,liu2022gen6d,he2022fs6d},
3D-coordinate-regression~\cite{hodan2020epos,wang2021gdr,haugaard2022surfemb,brachmann2014learning,brachmann2016uncertainty,li2019cdpn,rad2017bb8,tremblay2018deep,cai2020reconstruct},
and render-and-compare~\cite{li2018deepim,labbe2020cosypose,labbe2022megapose,shugurov2022osop,park2020latentfusion}.
One key property for the deployment of pose estimators is their generalization that falls into three categories: no generalization, generalization to object categories seen at training time~\cite{lin2022single,wang2019normalized,chen2020category,manuelli2019kpam}, and generalization to any object~\cite{labbe2022megapose,foundationposewen2024,lin2024sam,nguyen2024gigaPose,ornek2023foundpose,moon2024genflow,caraffa2025freeze}.
Pose estimators that generalize to any objects are the most suitable in practical scenarios as it is unrealistic to train the estimators on any object instance or category that might be encountered during deployment.
However, they often rely on CAD models and it is also impractical to generate CAD models for all objects beforehand.
One solution is to develop novel pose estimators that use alternative 3D representations such as \gls{sfm} point clouds~\cite{sun2022onepose,he2022onepose++}, implicit rendering~\cite{chen2023texpose,park2020latentfusion}, and 3D models generated with implicit functions~\cite{foundationposewen2024} or diffusion~\cite{nguyen2024gigaPose,long2024wonder3d}.
The proposed benchmark follows another avenue which is \textit{how well 3D reconstruction methods measure against CAD models in their use for pose estimation}.
To do so, we fix the pose estimation and analyze its relative performance with various 3D reconstructions.
To the best of our knowledge, there is no other benchmark that evaluates 3D reconstruction methods based on their performance on the downstream object pose estimation task.
The results drawn by the proposed benchmark draw novel insights relevant for the pose estimation community on the advantages and limitations of current reconstruction methods, \eg, for which types of objects 3D reconstruction can replace CAD models.
At the time of the writing, the RGBD-based FoundationPose~\cite{foundationposewen2024} sets the state-of-the-art 
which motivates its integration in the benchmark to evaluate the 3D reconstructions.
We also integrate Megapose~\cite{labbe2022megapose} for it adopts a similar render-and-compare approach but on RGB images.
Please refer to the supplementary for a detailed description of these methods.

\PAR{3D Reconstruction.}
Given a set of calibrated images with \textbf{known camera poses}, 
\gls{mvs} approaches establish dense correspondences between images, which in turn are used to reconstruct the 3D structure of the scene.
Traditionally, MVS approaches use \textbf{explicit} representations such as dense point clouds~\cite{furukawa2010accurate,lhuillier2005quasi}, depth maps~\cite{strecha2006combined,schonberger2016pixelwise,bleyer2011patchmatch,kanade1995development}, surface hulls~\cite{furukawa2006carved,sinha2004visual}, or voxels~\cite{seitz1999photorealistic,kutulakos2000theory,sinha2007multi,ulusoy2017semantic}. 
If desired, meshes can be extracted from these representations, \eg, using Poisson surface reconstruction~\cite{kazhdan2013screened}. 
In our benchmark, we evaluate two MVS-based 3D reconstruction approaches:  
the open-source COLMAP~\cite{schonberger2016structure,schonberger2016pixelwise} that commonly is used as a baseline for 3D reconstruction methods and the state-of-the-art commercial system Reality Capture~\cite{capturereality} known for its efficiency.

Alternatives to explicit reconstructions are \textbf{implicit} representations~\cite{faugeras1997level,curless1996volumetric,newcombe2011kinectfusion}, \ie, continuous functions defined over the 3D space that embed properties about the scene, \eg, \gls{sdf}.
Defining implicit functions can be complex so it is convenient to learn them instead~\cite{mescheder2019occupancy,park2019deepsdf,wang2021neus,sitzmann2019scene}.
With the advent of differentiable rendering, several implicit functions relevant for 3D reconstruction can be learned simply from calibrated-image-supervision including volumetric radiance field~\cite{mildenhall2020nerf}, occupancy network~\cite{oechsle2021unisurf}, and \gls{sdf}~\cite{yariv2020multiview,niemeyer2020differentiable}.
Volumetric methods~\cite{mildenhall2020nerf,lombardi2019neural}
represent the scene as a colored volume where a network outputs a volume density and a color for each point in the space.
Rendering the volume density~\cite{kajiya1984ray} produces depth maps suitable to generate 3D meshes using Poisson surface reconstruction~\cite{kazhdan2013screened}.
Volumetric methods are optimized for novel view synthesis only whereas optimizing SDF networks involves more geometric constraints~\cite{gropp2020implicit} to ensure that the network exhibits SDF properties such as zero-level set corresponding to the surface.

\PAR{Benchmarking 3D Reconstruction Algorithms.}
Existing datasets and benchmarks~\cite{strecha2008benchmarking,ETH3D_bench,knapitsch2017tanks,seitz2006comparison,scharstein2014high,scharstein2002taxonomy,hirschmuller2007evaluation,jensen2014large,aravecchia2024comparing}
are tailored to measure the accuracy of the 3D scene geometry itself.
In contrast, we are interested in determining to what degree 3D reconstruction algorithms can be used to aid object pose estimation.
Thus, we evaluate the performance of a 3D reconstruction method by the accuracy of the object pose estimates it facilitates.
In parallel, we also evaluate how reconstruction performance (as measured by existing benchmarks) correlates to the object pose performance, \ie, can one really on the reconstruction performance to assess whether the reconstructed 3D model is suitable for object pose estimation.
The results show that one reconstruction method that is better according to existing reconstruction benchmarks can lead to worse pose estimation than another method. 
This clearly shows the need for our novel benchmark, which evaluates 3D reconstruction methods based on the resulting pose accuracy.

\PAR{Object Pose Estimation Benchmarks.}
Pose evaluation datasets provide object CAD models, images depicting those objects in various positions
and the object poses~\cite{hinterstoisser2013model,brachmann2014learning,hodan2017t,drost2017introducing,kaskman2019homebreweddb,tyree20226,xiang2018posecnn,doumanoglou2016recovering,rennie2016dataset,tejani2014latent}.
These many datasets and benchmarks are unified in the common BOP benchmark by Hodan~\etal~\cite{hodan2018bop,hodavn2020bop} that sets the data format, metrics, and evaluation tools.
We follow their guidelines to evaluate object pose estimation accuracy.
Contrary to the BOP benchmark that focuses on the pose estimators, 
our benchmark focuses on the 3D reconstruction methods and how to improve them as to make the deployment of existing pose estimation algorithms easier in practice.

We chose to base the proposed benchmark on the YCB-V dataset~\cite{xiang2018posecnn} for two reasons:
(1) it is an established dataset for object pose estimation and it displays objects with various sizes and textures relevant to the evaluation.
(2) The objects shown in the dataset are available\footnote{\url{https://www.ycbbenchmarks.com/}}, allowing us to capture new images of them suitable for 3D reconstruction.
Such images should show the objects without occlusions (to enable reconstructing the full object), focus on the object (as not to lose pixels / resolution to the background), and be of high resolution (to facilitate high-quality reconstruction).
The evaluation datasets in the BOP benchmark, including YCB-V~\cite{xiang2018posecnn}, provide images with the opposite properties, which makes for interesting pose estimation challenges, but the images are unsuitable for high-quality 3D reconstruction.
One exception is the T-LESS~\cite{hodan2017t} dataset but it is limited to small objects with grey uniform and Lambertian texture, which would severely limit the insights drawn by the evaluation as shown in the results.
The combination of (1) and (2) ensures that we do not have to create a new benchmark for object pose estimation and instead can use an existing one to evaluate 3D reconstructions.
We believe that using a dataset from an existing pose estimation benchmark that is well-known to the community will make the adoption of the proposed benchmark easier.
To the best of our knowledge, there is currently no other existing dataset satisfying both conditions.

\section{Dataset}
\label{sec:benchmark}

%
%
%
The images used for the reconstruction are captured with a camera mounted on a robotic arm in an indoor environment.
The reconstruction algorithms run on these calibrated images and the resulting
meshes are compared to the reference CAD models on their use for object
pose estimation.
Our benchmark thus consists of two components: (1) a set of novel images of a
variety of different objects with known intrinsic and extrinsic parameters
that can be used to build 3D models; (2) The set of BOP test images to evaluate the meshes on their integration into the object pose estimation pipeline. The test images show the objects in
different scenes and provide ground truth poses for the objects used to evaluate object pose estimation.
\textbf{Released Data:} The dataset (1) contains calibrated and undistorted images for all 21 objects, object masks, and object 3D reconstructions.

\PAR{Data Acquisition.}
The images are collected with a Basler Ace camera
acA2440-20gc~\cite{basler}
mounted on the flange of a 7 \gls{dof} KUKA LBR IIWA 14 R820 on a sphere around the object (see Sec. A in the Supp.).
The positions are evenly distributed on the sphere with latitude and longitude spanning over [0$^{\circ}$,150$^{\circ}$] and [0$^{\circ}$,360$^{\circ}$] respectively, with intervals of 10$^{\circ}$.
The number of images per object varies between 397 and 505 images with an average of 480.
The image resolution is 2448x2048 pixels and the field of view covers the whole object.
The camera exposure remains fixed during the object scan.
The camera and the hand-eye calibrations are run with OpenCV~\cite{bradski2000opencv} and MoveIt~\cite{chitta2012moveit,moveitcalibration}
with an error of up to 5mm in translation and 0.1$^{\circ}$ in rotation.
The resulting camera calibration is used to undistort the images using
COLMAP~\cite{schonberger2016structure}.
The camera poses obtained from the kinematic chain of the robot are later refined with an offline optimization (see Sec.4).
%
The motion planning and the collision avoidance are operated with the IIWA stack~\cite{hennersperger2017towards} and  MoveIt~\cite{chitta2012moveit} with the CHOMP planning algorithm~\cite{ratliff2009chomp}.

\section{3D Object Reconstruction Benchmark}
\label{sec:eval}

The proposed benchmark answers the question of how well can the 3D models generated by current 3D reconstruction methods replace the CAD models commonly used in object pose estimation.
To this end, we compare the performance of two \gls{sota} object pose estimators~\cite{labbe2022megapose,foundationposewen2024} when using the 3D reconstructed models against the use of classical CAD models.
This section describes the benchmark setup including the evaluated pose and reconstruction methods, and the registration of the reconstructions with the YCB-V~\cite{xiang2018posecnn} CAD models.

\PAR{Evaluated Pose Estimators.}
In this work, we are interested in pose estimators that generalize to objects unseen during training and with high performance to facilitate the deployment of pose estimators in real-world scenarios.
At the time of the writing, the state-of-the-art is led by the CAD-based FoundationPose~\cite{foundationposewen2024} that operates in a render-and-compare fashion on RGB-D images.
We also evaluate Megapose~\cite{labbe2022megapose} for it adopts a similar render-and-compare strategy but on RGB-only images, which can make it more practical in real-world scenarios where RGB cameras are cheaper to deploy and used more frequently.

\PAR{Evaluated Reconstruction Methods.}
Given the expansive literature on neural implicit representations, it is unrealistic to evaluate them all, especially given the computational complexity of the benchmark.
It roughly takes 8 days on 8 Nvidia A-100 GPUs to evaluate the following methods with a single pose estimator, which is a non-negligible cost.
Instead, we select gold-standard methods or methods representative of new optimization paradigms.
We release the data used in the benchmark and invite external contributions to extend the benchmark to more methods.

UniSURF~\cite{oechsle2021unisurf} is a surface representation that approximates the SDF with an occupancy network, contrary to the majority of methods, such as NeuS~\cite{wang2021neus}, that approximates the SDF directly with a neural network to reach better reconstruction accuracy.
MonoSDF~\cite{yu2022monosdf} integrates off-the-shelf geometric and depth cues in the SDF learning process to further improve the reconstruction quality.
In parallel, VolSDF~\cite{yariv2021volume} improves the geometry learned with volumetric representations by constraining the volumetric density to be a function of the scene's surface geometry defined by an \gls{sdf}.
BakedSDF~\cite{yariv2023bakedsdf} builds on top of VolSDF to learn an implicit representation that is then baked into a triangle mesh, \ie, it is stored in the mesh, which is more efficient to render.
The mesh is augmented with spherical Gaussians at each vertex that are further trained to represent the view-dependent appearance of the mesh.
NeuralAngelo~\cite{li2023neuralangelo} improves the resolution of the mesh's surface in a memory-efficient way by adopting the multi-resolution feature hash-grids introduced by iNGP~\cite{muller2022instant} on top of which tiny-MLPs predict the surface field.
A mesh is obtained from the SDF using the Marching Cubes algorithm~\cite{lorensen1987marching}.
We also evaluate Plenoxels~\cite{fridovich2022plenoxels} that represents the scene with a feature grid with density values and spherical harmonics optimized with differentiable rendering.
The absence of a neural network makes the training faster than the traditional NeRF~\cite{mildenhall2020nerf} methods.
As for the volumetric methods, we evaluate iNGP~\cite{muller2022instant} as a representative of fast feature-grids-based methods and the Nerfacto~\cite{nerfstudio} implementation of Nerf~\cite{mildenhall2020nerf} that integrates several contributions, including~\cite{muller2022instant}, that the authors have found to work well in real-case scenarios. 
One relevant property of Nerfacto~\cite{nerfstudio} is its adaptive sampling along the ray that typically helps the training focus on the useful parts of the scene, \ie, where the object is located.
Last, we evaluate the \gls{mvs} methods COLMAP~\cite{schonberger2016pixelwise,schonberger2016structure} and Reality Capture~\cite{capturereality}.

The meshes thus obtained are used in place of the original YCB-V~\cite{xiang2018posecnn} CAD models to generate object renderings used in the render-and-compare pose estimators.
Another way to generate rendering from implicit representations is to synthesize novel views directly from the trained network.
While this has the advantage of bypassing the mesh generation~\cite{lorensen1987marching,kazhdan2013screened}, it remains, for now, slower than mesh rendering for which efficient implementations are readily available.
Still, we evaluate how well can implicit renderings serve pose estimation with iNGP~\cite{muller2022instant} for its fast rendering compared to the prohibitive runtime of other methods (\eg, Megapose with Nerfacto~\cite{nerfstudio} takes close to 2 minutes on A100 GPU for a single image).

\PAR{Pose Refinement.}
All reconstructions run on the undistorted images with the camera intrinsics and extrinsics collected during the data acquisition (see Sec.~\ref{sec:benchmark}).
The camera extrinsics are obtained from the kinematic chain of the robot which can be noisy, thus we refine them in two steps: first with the iNGP~\cite{muller2022instant} pose refinement then with bundle adjustment~\cite{schonberger2016structure} (see Sec. A.1). 
After this step, the camera extrinsics remain fixed.
Although some evaluated methods have pose refinement capabilities, we turn them off so that all reconstruction methods share the same coordinate frame to avoid introducing biases in the evaluation.

\PAR{Reconstruction and Texturing.}
We use the default parameters for COLMAP~\cite{schonberger2016structure,schonberger2016pixelwise}, which are already optimized for reconstruction purposes.
Before running the Poisson surface reconstruction~\cite{kazhdan2013screened}, we filter out the room's background by roughly cropping the dense point cloud around the object. %
A similar processing is applied for Reality Capture~\cite{capturereality}.
The learning-based methods undergo the pre-processings described by their respective authors (see Sec. B.1 in the supp. mat.).
The texture of learning-based methods is generated by querying the network for the color at the position of the mesh's faces and Reality Capture~\cite{capturereality} uses proprietary code.
For methods that do not have texturing capabilities (COLMAP~\cite{schonberger2016structure} and iNGP~\cite{muller2022instant}), we evaluate the colored mesh since this leads to better pose estimates than if the mesh was textured with an off-the-shelf texturing software~\cite{Waechter2014Texturing} (see Sec. C.4 in the supp. mat.).

\PAR{Mesh Registration.}
To enable the pose evaluation on the YCB-V~\cite{xiang2018posecnn} images, the reconstructions must be registered to the original YCB-V meshes because the ground-truth object poses are defined in the coordinate frame of the original meshes.
The rigid transform between the coordinate frame of the original mesh and the world coordinate frame where we placed the captured objects is estimated with 
\gls{icp}~\cite{besl1992method} initialized with a user-defined coarse alignment.
Once the 3D models are aligned, we further crop them to filter out any remaining background in the reconstructions.

Some of the original YCB-V~\cite{xiang2018posecnn} objects were discontinued so we use the look-alike alternatives recommended by the YCB benchmark~\cite{ycbbench}. 
Thus, the objects collected in this benchmark may differ in texture and scale,
with an average scale change of 4\% of the object's dimensions.
Given the small object's dimensions, such changes often fall below the distance thresholds used in the metrics. %
Therefore, we choose to keep the scale of the collected objects to be consistent with the current object's reality.

\section{Experimental Evaluation}
\label{secresults}

We compare the performance of \gls{sota} object pose estimators~\cite{labbe2022megapose,foundationposewen2024} when using the 3D reconstructed models against the use of classical CAD models.
We follow the object pose evaluation guidelines defined in the BOP
benchmark~\cite{hodan2018bop,hodavn2020bop} that provides accurate CAD models.
These CAD models are used as a reference for the reconstruction evaluation and to estimate the baseline object poses (\textbf{Oracle}).
We recall the metrics used to evaluate the reconstructions and the object poses.
We then analyze how the quality of the estimated poses is impacted by the 3D reconstructions, the object properties, the texturing, and the amount of images used in the reconstruction.

\begin{figure*}[t]
   \centering
    \begin{subfigure}[b]{0.27\textwidth}
        \centering
   \includegraphics[width=\linewidth,height=\textheight,keepaspectratio]
  {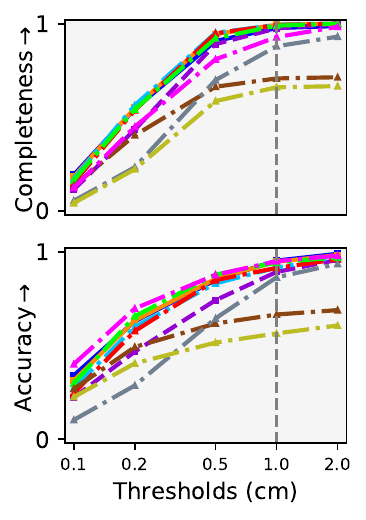}
    \end{subfigure}%
    \begin{subfigure}[b]{0.34\textwidth}
        \centering
   \includegraphics[width=\linewidth,height=\textheight,keepaspectratio]
   {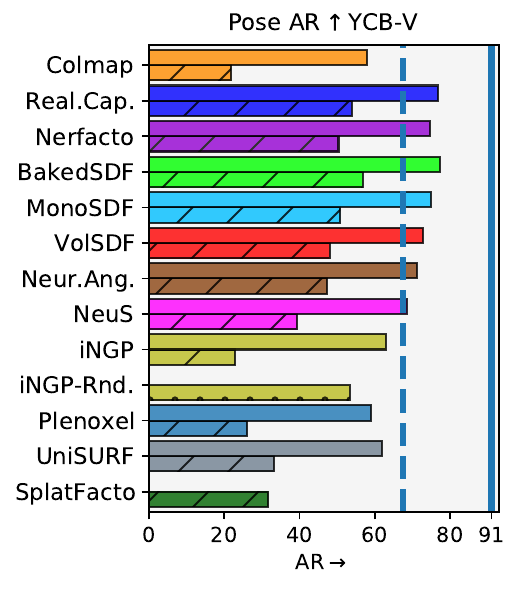}
    \end{subfigure}%
        \begin{subfigure}[b]{0.33\textwidth}
        \centering
      \includegraphics[width=\linewidth,height=0.13\textheight,keepaspectratio]
        {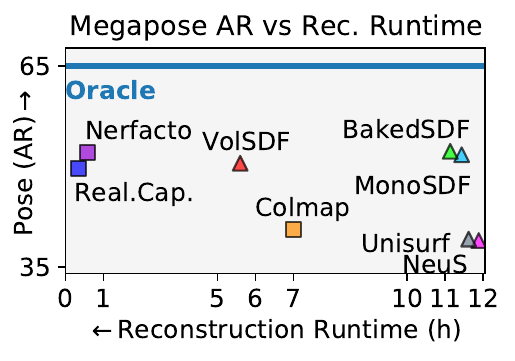}
    \includegraphics[width=\linewidth,height=0.16\textheight,keepaspectratio]
        {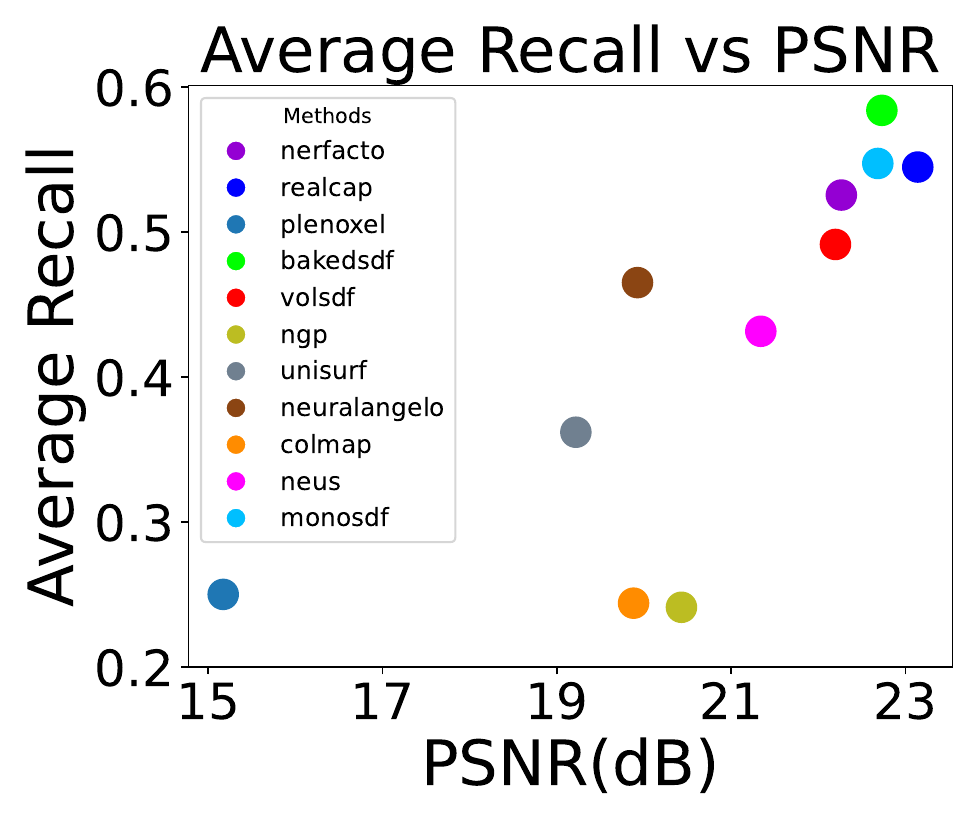}
        \label{fig:ar_vs_incomplete_rec}
    \end{subfigure}%
   \caption{
     \textbf{3D Reconstruction Evaluation.}
     Left: Most reconstructions exhibit high geometric quality. 
     Center: 3D reconstructions evaluated with the Average Recall (AR) of Megapose~\cite{labbe2022megapose} (dashed and dotted) and FoundationPose~\cite{foundationposewen2024} (uniform) when replacing the CAD models with the reconstructions.
     The pose evaluation is more discriminative than the geometric one (left) and the relative ranking of the methods is consistent between the two pose estimators.
     Right: Runtime, mesh appearance quality and pose quality~\cite{labbe2022megapose}. Nerfacto~\cite{nerfstudio} and Reality Capture~\cite{capturereality} achieve the best performances when accounting for both the pose recall and the runtimes.
   }
   \label{fig:results_rec}
\end{figure*}

\PAR{Implementation Details.}
The Nerfstudio~\cite{nerfstudio} and SdfStudio~\cite{Yu2022SDFStudio} codebases
are used to run Nerfacto~\cite{nerfstudio}, MonoSDF~\cite{yu2022monosdf}, VolSDF~\cite{yariv2021volume}, UniSURF~\cite{oechsle2021unisurf}, NeuS~\cite{wang2021neus}, and BakedSDF~\cite{yariv2023bakedsdf}.
We use the author's codebase for all other methods and for the pose estimation~\cite{labbe2022megapose,foundationposewen2024}, the MVS-texturing~\cite{Waechter2014Texturing}, the reconstruction evaluation~\cite{schops2017multi} and the pose evaluation~\cite{boptoolkit}.
The mesh registration and post-processing are run with Open3D~\cite{Zhou2018}.
All methods run on a single NVIDIA A100 GPU except for RealityCapture that runs on a GeForce RTX 3060 for it requires a Windows installation. 

\PAR{Reconstruction Metrics.}
The reconstructed geometries are evaluated with the completeness and
accuracy~\cite{schops2017multi} over distance thresholds ranging from 1mm to 2cm.
The \textit{completeness} measures how much of the reference mesh is modeled by the
reconstructed one: it is the ratio of reference points which nearest-neighbor
in the reconstructed mesh is within a distance threshold. 
The \textit{accuracy} assesses how close the reconstructed mesh is to the reference one with the ratio of reconstructed points within a distance threshold to the reference points.
These metrics are sensitive to the mesh's density so we first
subsample the reconstructed meshes to a fixed-size set of points before evaluation.

\PAR{Object Pose Estimation and Metrics.}
The object poses are estimated with the \gls{sota} RGB Megapose~\cite{labbe2022megapose} and RGB-D FoundationPose~\cite{foundationposewen2024}.
The evaluation follows the protocol of the standard BOP benchmark~\cite{hodan2018bop,hodavn2020bop} and reports the \textbf{Average Recall (AR)} on the three errors defined next.
These errors do not measure the object pose translation and rotation errors but measure the misalignment of the object when it is positioned with the estimated pose and with the ground-truth pose.
Such assessment not only accounts for the poses' discrepancy but also the impact of said discrepancy on the perception or manipulation of the object, which is a relevant indicator in industrial applications.

The \gls{vsd} measures the distance between the surfaces of the object at the ground-truth and estimated poses.
The \gls{mssd}
measures the maximum surface distance
between the two poses.
This metric is symmetry-aware, \ie, it does not penalize the estimated pose for ambiguous symmetries.
The \gls{mspd} measures the visual discrepancy in pixel distance induced by an incorrect pose when rendering the object.
The fraction of images for which the errors are below a given threshold defines the recall. It is averaged over several error thresholds and over images to form the \textbf{AR} for each error. The resulting three ARs are averaged to form the global AR that we report next.
\textit{Interpretation: }
the VSD and MSSD are good indicators of the object pose quality for navigation or object-manipulation applications while the MSPD is better suited for applications that exploit object rendering, such as Augmented and Virtual Reality~\cite{hodan2018bop}. 

\PAR{Reconstruction and Pose Scores.}
\label{res:rec_perf}
\fig~\ref{fig:results_rec} shows quantitative results on the geometric quality of the reconstructions 
(left), their performance when integrated into FoundationPose~\cite{foundationposewen2024} and Megapose~\cite{labbe2022megapose} (center), and the correlation between the estimated poses~\cite{labbe2022megapose}, the appearance quality of the mesh, and the reconstruction runtime (right).
The results demonstrate the practical advantage of the MVS-based Reality Capture~\cite{capturereality} that achieves \gls{sota} reconstruction performance under the shortest runtime (recall that RealityCapture is run on a weaker GPU).
Overall, there remains a 10-20\% gap in quality between CAD models and the best 3D reconstructions
(\fig~\ref{fig:results_rec}-center).
Given that the relative pose performances of the reconstructions are consistent between the two pose estimators, one can expect advances in 3D reconstruction to close the gap for other pose estimators.

\begin{figure*}[t]
   \centering
\includegraphics[width=0.84\linewidth,height=\textheight,keepaspectratio]
{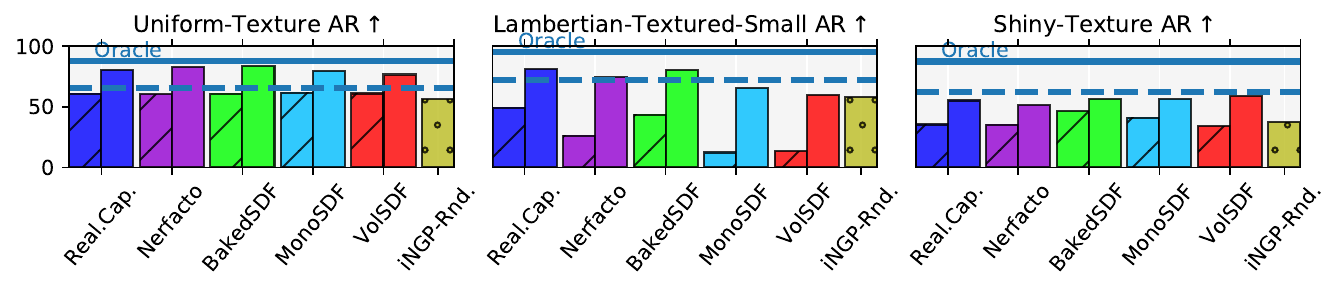}
   \caption{\textbf{Pose Evaluation on Object Categories.}
   Average Recall (AR) of Megapose~\cite{labbe2022megapose} (dashed and dotted bar) and FoundationPose~\cite{foundationposewen2024} (uniform bar) when replacing the CAD models with the reconstructions.
    Some categories expected to be challenging (uniform-texture objects) are not whereas
    small and shiny objects are more challenging.
   }
   \label{fig:ar_vs_categories1}
\end{figure*}

The reconstructions best for pose estimation usually exhibit high geometric and appearance quality.
Still, there is no direct correlation between the reconstruction metrics and the pose scores: for example the geometric completeness of UniSURF~\cite{oechsle2021unisurf} and %
iNGP~\cite{muller2022instant}
is relatively low, which is reflected in the pose performance.
At the same time,  RealityCapture~\cite{capturereality} performs comparably to BakedSDF~\cite{yariv2023bakedsdf} in terms of completeness and accuracy but leads to slightly worse poses.
Conversely, the reconstructed geometry of Nerfacto~\cite{nerfstudio} is worse than MonoSDF~\cite{yu2022monosdf} but their pose evaluation is on par.
One interesting observation is that the geometric evaluation (\fig~\ref{fig:results_rec}-left) does not discriminate between VolSDF and MonoSDF (superimposed red and cyan curves) whereas the pose evaluation does.
This supports the necessity for the proposed benchmark and validates the proposed reconstruction evaluation that not only measures the accuracy of the reconstruction itself but also the quality of the pose estimation it enables.

There is also no immediate correlation between the quality of the mesh's appearance and the pose quality (\fig~\ref{fig:results_rec}-right).
For example, Reality Capture reaches a higher PSNR than BakedSDF yet the latter leads to better poses.
As for Colmap~\cite{schonberger2016structure} and NeuralAngelo~\cite{li2023neuralangelo}, they exhibit similar PSNR but the former leads to much worse pose estimation.

Another remarkable result is the high performance when Megapose~\cite{labbe2022megapose} uses the iNGP~\cite{muller2022instant} implicit renderings instead of the iNGP mesh's ones.
This shows that an off-the-shelf implicit representation can be integrated with a pose estimator, which completes recent contributions that train pose estimation together with implicit renderings~\cite{foundationposewen2024,chen2023texpose}.
This also suggests that pose estimators can perform well even on objects hard to reconstruct as long as the novel view synthesis is good.

\PAR{Finer Pose Evaluation.}
The BOP benchmark defines multiple metrics to assess how suitable the pose estimator, hence the 3D reconstruction, is for a given application such as augmented and virtual reality (MSPD), navigation (VSD), and object manipulation (MSSD).
Due to the page limit, we report the plots in Sec. C.1 of the supp. mat.
The VSD recall of the reconstructions remains close to the baseline so objects are overall well-positioned, which is useful for global object reasoning such as object avoidance and navigation.
However, there is a drop in MSSD recall that is particularly sensitive to small misplacements when objects exhibit high curvature regions.
This is typical of grasping points so this signals a limitation when using reconstructed meshes for object manipulation.
Another limitation relates to virtual reality because of the MSPD recall drop that indicates visual dissonance when rendering objects.

\textbf{Object Properties.}
We split the YCB-V~\cite{xiang2018posecnn} objects into 8 groups based on their shape, texture, material properties, and the degree of change between the collected objects and the original ones. We show the plots for 3 groups in Fig.~\ref{fig:ar_vs_categories1} and the rest in Sec. C.3 of the supp. mat. The pose performance variations across groups are telling of some of the current 3D reconstruction challenges.

For simple objects such as large objects with lambertian texture, whether uniform or not, 3D reconstructions can replace CAD models without a significant drop in performance for both pose estimators.
These results on objects with little or uniform texture suggest that the reconstructed geometry is reliable for pose estimation: since the texture has little information, it marginally contributes to the pose estimation contrary to the geometry.
Also, this suggests that imperfect texturing is not necessarily prohibitive for pose estimation on objects with little texture.

\begin{figure*}[t]
   \centering
    \begin{subfigure}[b]{0.62\textwidth}
        \centering
        \includegraphics[width=\linewidth,height=\textheight,keepaspectratio]
  {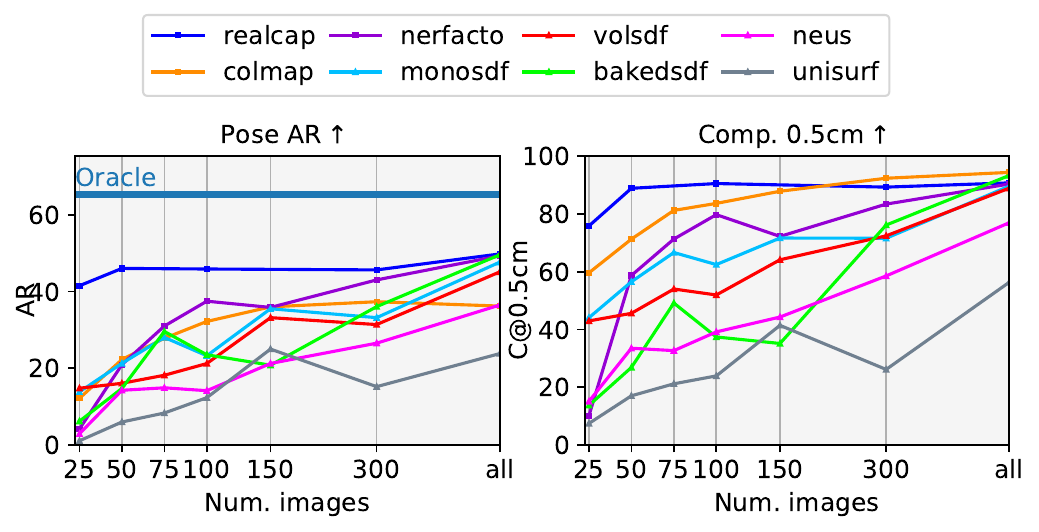}
    \end{subfigure}%
    \begin{subfigure}[b]{0.30\textwidth}
        \centering
        \includegraphics[width=\linewidth,height=\textheight,keepaspectratio]{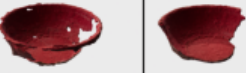}
        \includegraphics[width=\linewidth,height=\textheight,keepaspectratio]{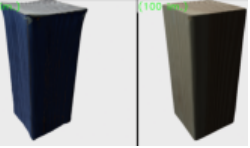}
        \label{fig:ar_vs_incomplete_rec}
    \end{subfigure}%
    \caption{
    \textbf{Reconstruction from image sets with increasing size} (left).
     Average Recall (AR) of Megapose~\cite{labbe2022megapose} when replacing the object's CAD model with the reconstruction; and 
     geometric evaluation of the reconstruction with completeness under a 5mm error threshold.
     Note that the performance of RealityCapture~\cite{capturereality} remains stable when the number of images goes down and runs extremely fast: on average, it takes less than 1 min. on 25 images and 2 min. on 50 images.
    \textbf{Subpar reconstructions and good pose estimation} (right).
    The reconstructed bowls~\cite{capturereality} 
     both miss parts yet the left one reaches oracle performance (AR=38) whereas the one on the right is subpar (AR=15).
    Both wood blocks
    (AR=39) outperform the oracle (AR=27) even though the left texture is incorrect.
    }
    \label{fig:ar_vs_num_images}
\end{figure*}

Conversely, the reconstruction of objects with shiny or small texture elements performs worse than other categories, independently of the pose estimators.
Several factors contribute to this drop.
First, several of the objects in these categories have been discontinued or got their texture updated. %
The texture inconsistency between the test images and our reconstructions impedes the pose estimation, which is confirmed by the higher pose performance on the `legacy' objects, \ie, the objects that did not change, than on the updated ones.
Another factor 
is that small and shiny textures are harder to generate than Lambertian ones.
When it comes to small objects ($\le$ 8cm), the reconstruction generates a low-resolution texture, which decreases the amount of visual information relevant to disambiguating the estimated poses.
This is all the more critical as those objects exhibit symmetries that can only be differentiated through their texture (07-pudding-box, 06-tuna-fish-can).
As for shiny objects, the light reflection induces a visual discrepancy between the target texture and the reconstructed one. 

None of the top-performing methods exhibit an advantage over the others for the more challenging object types.
Given their ability to model view-dependent effects such as reflections, one would have expected NeRF-based methods~\cite{nerfstudio,muller2022instant} to significantly outperform RealityCapture~\cite{capturereality} which does not model view-dependent effects, but this is not the case.
These results suggest that the texture generation of the 3D reconstruction has room for improvement.

\textbf{Number of Images.}
Besides the quality of the pose estimation, an important property of the reconstruction methods is their data requirement.
\fig~\ref{fig:ar_vs_num_images}-left shows the geometric and Megapose evaluation~\cite{labbe2022megapose} of reconstructions generated from varying numbers of images sampled uniformly around the object~\cite{gonzalez2010measurement}. %
As expected, the results get better with the number of images and 75-100 images is enough to generate reconstructions close to the best ones.
However, we observe that using sparse views can hinder the convergence of some reconstructions.
We also observe that the performance does not increase linearly with the data
which we attribute to non-uniform noise in the camera poses.
One source of such noise is in the robotic arm's positional encoders: they are known to be noisier in configurations that are close to degenerate, \eg, when the arm is extended.
These results show that another avenue for improvement includes robustness to noisy poses and reconstructions from limited views.
In parallel, qualitative results in \fig~\ref{fig:ar_vs_num_images}-right illustrate examples where reconstructions are visually subpar yet informative enough for satisfying pose estimation: the two instances of the object lead to similar pose estimation performance.
This suggests that incomplete meshes can be relevant for object pose estimation as long as the discriminative parts of the objects are reconstructed such as the outer edge of the bowl.

\section{Conclusions}
In this work, we have considered the problem of reconstructing 3D models from images such that they can be used for object pose estimation. 
In particular, we aimed to answer the question of whether state-of-the-art 3D reconstruction algorithms can be readily used to remove the need for CAD models during training and testing of object pose estimators. 
We have created a new benchmark, based on the YCB-V~\cite{calli2015ycb,xiang2018posecnn} dataset, that can be used to measure the impact of replacing CAD models with reconstructed meshes. 
We have evaluated multiple state-of-the-art reconstruction methods. 
Our results show that reconstructed models or their renderings can often readily replace CAD models. 
Still, there is a considerable gap between CAD and image-based models for certain types of objects that are hard to reconstruct, \eg, small or shiny objects. 

Several interesting observations can be drawn from our results: 
(1) The classical, handcrafted Reality Capture often performs similarly to state-of-the-art learning-based approaches.
It is also very efficient and able to reconstruct scenes within a few minutes, which makes it a strong baseline for future work.
(2) The ranking between different 3D reconstruction approaches is quite consistent between the two evaluated pose estimators.
There thus is hope that all pose estimators will automatically benefit from advances in the area of 3D reconstruction.
(3) Incomplete reconstructions, where parts of the object cannot be reconstructed, are not necessarily problematic as the resulting models can still enable accurate pose estimation as long as enough parts are reconstructed. 
This observation validates our idea of evaluating reconstruction systems inside a higher-level task as classical reconstruction-based metrics, \eg, accuracy and completeness, are falsely indicating the resulting models as low-quality reconstructions.
(4) There is not a clear relation between the quality of the estimated 3D scene geometry and pose accuracy. For example, the renderings produced directly by the iNGP network achieve quite good performance with Megapose whereas the renderings of the iNGP mesh fall behind.
This is a promising result as it suggests that objects that are hard to reconstruct can still allow accurate pose estimation if the novel view synthesis performs well.
This also supports the necessity for the proposed benchmark that evaluates 3D reconstruction on their integration with pose estimators rather than with the standard reconstruction evaluations.
(5) In practice, one would ideally use as few images as possible for the reconstruction to minimize both capture and reconstruction time. 
Our results indicate room for improvement in the few-image regime. 

While our results show that using reconstructed models is a promising direction for removing the need for CAD models, we also observe that the problem is far from solved. 
We believe that our benchmark will help drive research on approaches that can accurately reconstruct challenging objects from as few images as possible.

\PAR{Acknowledgements.} 
This work was supported by the Czech Science Foundation (GACR) EXPRO (grant no. 23-07973X),
the Ministry of Education, Youth and Sports of the Czech Republic through the e-INFRA CZ (ID:90254),
the Grant Agency of the CTU in Prague, grant No.SGS23/172/OHK3/3T/13,
the EU under the project Robotics and advanced industrial production - ROBOPROX (reg. no. CZ.02.01.01/00/22\_008/0004590),
the EU Horizon 2020 project RICAIP (grant agreement No 857306).

%% file: SupplementaryContent.tex
The supplementary material is organized as follows: Sec.~\ref{supp:sec_dataset} describes the data acquisition and the information related to the release as announced in Sec.3 of the main paper.
Sec.~\ref{supp:sec_eval_setup} provides implementation details relative to the reconstructions, the pose estimation, and the evaluation, as announced in Sec. 4 of the main paper.
Sec.\ref{supp:sec_results} reports further quantitative results that support the conclusions drawn in Sec. 5 of the main paper.

\section{Dataset}
\label{supp:sec_dataset}

The proposed benchmark answers the question ``To what degree 3D models reconstructed by current 3D reconstruction algorithms can replace the CAD models commonly used for object pose estimation?'' 
To this end, we compare the performance of two \gls{sota} object pose estimators, FoundationPose and Megapose~\cite{labbe2022megapose,foundationposewen2024}, when using the 3D reconstructed models against the use of classical CAD models.
Thus, we mainly measure 3D reconstruction performance by the accuracy of the estimated poses rather than the accuracy of the resulting 3D models itself.

\begin{figure*}[t]
   \centering
  \includegraphics[width=\linewidth,keepaspectratio]
  {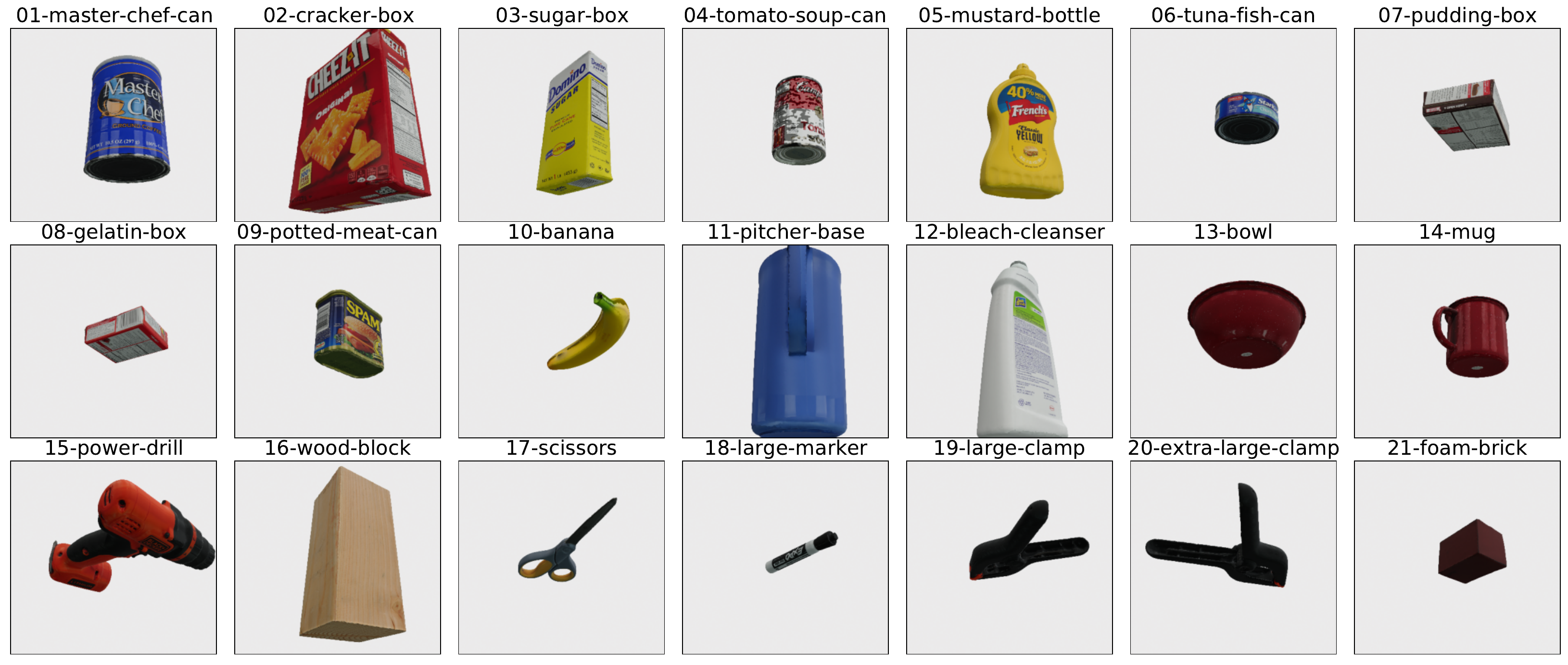}
   \caption{\textbf{The 21 YCB-V~\cite{calli2015ycb,xiang2018posecnn} objects} rendered from original CAD models.
   }
   \label{fig:ycbv_objects}
\end{figure*}

The evaluation of the pose estimators runs on the classic YCB-V~\cite{xiang2018posecnn} pose estimation dataset and follows the standard guidelines defined in the BOP
benchmark~\cite{hodan2018bop,hodavn2020bop}.
The YCB-V dataset~\cite{xiang2018posecnn} is made of 21 objects (Fig.~\ref{fig:ycbv_objects}) selected from the YCB dataset~\cite{calli2015ycb}, including small objects, objects with low texture, complex shapes, high reflectance, and multiple symmetries. 
For each object, our benchmark consists of two components: 
\textbf{(1)} Images of the object captured by a robot arm that can be %
{used for 3D reconstruction.}
\textbf{(2)} A set of test images for object pose estimation with ground truth poses registered with the image sets from (1).
Compared to the data for the first component, which we captured ourselves, we use the test images provided by the YCB-V~\cite{xiang2018posecnn} dataset for the second component. 
Given the two components, we evaluate multiple state-of-the-art 3D reconstruction methods.

In this section, we describe how the images in \textbf{(1)} were collected (Sec.~\ref{supp:data_collection}) and how the data is released (Sec.~\ref{supp:dataset_release}).

\begin{figure}[t]
   \centering
\includegraphics[width=\linewidth,keepaspectratio]
  {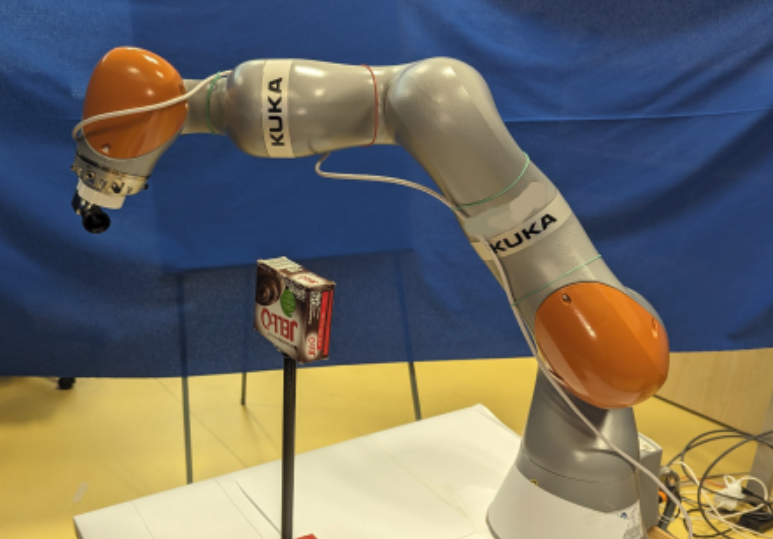}
   \caption{\textbf{Data Acquisition.}
   The objects are mounted on a tripod.
   A Basler Ace camera mounted on a 7DoF KUKA arm autonomously captures views from all sides of the YCB-V~\cite{calli2015ycb,xiang2018posecnn} objects.
   }
   \label{fig:setup}
\end{figure}

\subsection{Data Collection Setup}
\label{supp:data_collection}

\PAR{Data Acquisition.}
The images are collected with a Basler Ace camera
acA2440-20gc\footnote{https://docs.baslerweb.com/aca2440-20gc}
mounted on the flange of a 7 \gls{dof} KUKA LBR IIWA 14 R820 (Fig.~\ref{fig:setup}).
The image resolution prior to undistortion is 2448x2048 pixels and the field of view covers the whole object.
The acquisition is done in an open-floor indoor environment exposed to neon lighting and varying sunlight through ceiling-high windows, which is typical of daily life and industrial scenarios.
The camera exposure remains fixed during the object's scan.

The camera is calibrated using open-source solutions: the OpenCV routine for the camera calibration~\cite{bradski2000opencv} and the MoveIt~\cite{chitta2012moveit} routine for the hand-eye calibration\footnote{https://github.com/ros-planning/moveit\_calibration}.
Afterward, the camera can be positioned with an error of up to 5mm in translation and 0.1$^{\circ}$ in rotation.
The resulting camera calibration is used to undistort the images using the open-source COLMAP~\cite{schonberger2016structure} library.
The camera poses obtained from the kinematic chain of the robot are later refined with an offline optimization (see the paragraph `Pose refinement' below).

The objects are positioned on a tripod such that the centroid of the object is approximately at the center of the reachability domain of the robot.
For each object, dense outside-in views are captured by positioning the camera on a sphere centered on the object's centroid and with a radius $\sim$30cm.
The positions are evenly distributed on the sphere with latitude and longitude spanning over [0$^{\circ}$,150$^{\circ}$] and [0$^{\circ}$,360$^{\circ}$] respectively, with intervals of 10$^{\circ}$.
The lowest camera position at latitude 150$^{\circ}$ allows to capture views of the objects from all sides.
Since the object is a source of collisions, the number of reachable views varies depending on the object's shape and size.
The number of images per object varies between 397 and 505 images with an average of 480.
The motion planning and the collision avoidance are operated with the proprietary IIWA stack~\cite{hennersperger2017towards} and the open-source MoveIt~\cite{chitta2012moveit} framework using the CHOMP planning algorithm~\cite{ratliff2009chomp}.

\PAR{Image Registration.}
The YCB-V dataset~\cite{xiang2018posecnn} provides test images depicting the objects and their ground-truth poses.
These poses are the rotation and the translation between the camera coordinate frame and the object's coordinate frame, which is the coordinate frame of the original CAD model.
This coordinate frame is different from the `world' coordinate frame where we collect the images and reconstruct the objects.
Thus, the YCB-V ground-truth poses are not compatible with the coordinate frame of the 3D reconstructions. 
Consequently, we register the collected images to the original CAD model for each object. 
To do so, we first generate 3D reconstructions for all objects in their coordinate frame with COLMAP~\cite{schonberger2016structure} then estimate the rigid transform between the reconstruction's coordinate frame and the CAD model's coordinate frame with \gls{icp}~\cite{besl1992method,Zhou2018}.
The ICP is initialized with user-defined coarse alignments.

\PAR{Pose Refinement.}
The camera extrinsics of the collected images are obtained from the kinematic chain of the robot that can be relatively noisy.
One source of such noise is in the robotic arm's positional encoders: they are known to be noisier in configurations that are close to degenerate, \eg, when the arm is extended.
This can affect the quality of the reconstruction methods so we refine the camera extrinsics in two steps:
i) we use sequentially the pose refinement capabilities of iNGP~\cite{muller2022instant} that update the poses consistently with the trained 3D reconstruction; ii) we then use the iNGP-refined poses to generate a sparse 3D reconstruction and refine the camera poses again with bundle adjustment using COLMAP~\cite{schonberger2016structure}.
After this step, the camera extrinsics remain fixed.
Although some evaluated methods have pose refinement capabilities, we turn them off so that all reconstruction methods share the same coordinate frame.
Without this, the reconstruction's coordinate frames may be slightly different for each method depending on the amplitude of the pose refinement, which would introduce inconsistencies in the evaluation.

\PAR{Object masks.}
Once the images are registered with the original YCB-V~\cite{xiang2018posecnn}'s mesh, we generate object masks by projecting the mesh onto the image.
We sample a dense set of points on the mesh and project them on the image using the intrinsics and the registered extrinsics.
The pixels where the projected points fall form the object's mask.

\PAR{Subset Generation.}
In addition to the full set of captured images, we define image sets of different sizes as, in certain applications, data capture and 3D reconstruction times can be important. 
Providing smaller subsets of images allows us to simulate such scenarios. 
We provide subsets of the following sizes: 25, 50, 75, 100, 150, and 300.
The subsets are generated from the full set of images with Fibonacci sampling~\cite{gonzalez2010measurement} to ensure that the subset views cover the object uniformly.

\subsection{Data Release}
\label{supp:dataset_release}

The released data is available on the project's website:

\begin{center}
    \href{https://github.com/VarunBurde/reconstruction_pose_benchmark}{\texttt{reconstruction\_pose\_benchmark}}
\end{center}
The webpage contains instructions on how to download the data from an Apache server, either through a web page or a simple command line tool, \eg, \texttt{wget}.

\PAR{Released Data and Format.} %
We release the calibrated undistorted images of the 21 YCB-V~\cite{xiang2018posecnn} objects collected in Sec.~\ref{supp:data_collection}, the objects' masks, and the meshes generated by the reconstruction methods evaluated in the benchmark.

For each object, the undistorted images, their extrinsics, the camera intrinsics, and the object masks are packed in a single zip file.
The images and the masks are in \texttt{png} format, the extrinsics and intrinsics are released under both the Nerfstudio~\cite{nerfstudio} camera pose 
convention\footnote{docs.nerf.studio/quickstart/data\_conventions.html}
and the COLMAP~\cite{schonberger2016structure} camera pose convention\footnote{colmap.github.io/format.html\#text-format}.
These are two of the most commonly used camera pose conventions for 3D reconstruction.
Dataparsers for these two formats are available at the respective repositories.

For each reconstruction method, a single zip file contains all reconstructed objects.
The 3D reconstructions are saved under the standard \texttt{obj} format usually used by 3D processing libraries, \eg, Meshlab~\cite{LocalChapterEvents:ItalChap:ItalianChapConf2008:129-136}, Open3D~\cite{Zhou2018}, trimesh~\cite{trimesh}, and Pytorch3D~\cite{ravi2020pytorch3d}.



%

\PAR{License.}
The collected images and the reconstructed meshes are released under the CC BY 4.0 license: the license is conditioned on the license of the YCB~\cite{calli2015ycb} objects depicted in the images.
The YCB objects are released under the Creative Commons Attribution 4.0 International (CC BY 4.0)
~\footnote{http://ycb-benchmarks.s3-website-us-east-1.amazonaws.com/}
so the data is released under the same license.




\PAR{Benchmark Reproducibility.}
The benchmark uses open-source code for the reconstructions, the pose estimation, and the evaluation.
The only exception may be the code of the reconstruction method RealityCapture~\cite{capturereality}, which is free except for companies with high earnings\footnote{https://www.capturingreality.com/pricing-changes}.
Unless mentioned otherwise in Sec.B1, the default parameters of the reconstruction methods are used.

\section{Evaluation Setup}
\label{supp:sec_eval_setup}

Compared to existing benchmarks for 3D reconstruction, our benchmark does not treat 3D reconstruction as a task unto itself but rather evaluates the resulting 3D models inside a higher-level task, \ie, object pose estimation.
Thus, we mainly measure 3D reconstruction performance by the accuracy of the estimated poses rather than the accuracy of the resulting 3D models itself.
To this end, we compare the performance of two \gls{sota} object pose estimators~\cite{labbe2022megapose,foundationposewen2024} when using the 3D reconstructed models against the use of classical CAD models.

In the rest of this section, we report the implementation details relative to the evaluated 3D reconstructions (Sec~\ref{supp:rec_setup}), the pose estimators used for the evaluation (Sec~\ref{supp:pose_setup}), the pose evaluation metrics (Sec.~\ref{supp:pose_metrics}), and the nature of the objects in the evaluation dataset (Sec~\ref{subsec:object-categories}).

\subsection{Reconstruction Implementation Details}
\label{supp:rec_setup}

We describe the experimental setup related to the 3D reconstruction.

\PAR{\colmap~\cite{schonberger2016structure,schonberger2016pixelwise}.}
We use the default parameters of COLMAP's triangulation~\cite{schonberger2016structure,schonberger2016pixelwise}, which are already optimized for the purpose of reconstruction.
The feature extraction returns RootSIFT~\cite{arandjelovic2012three,Lowe04IJCV} features and the feature matching runs only between covisible image pairs: two images are covisible if they form an angle smaller than 45$^{\circ}$ with respect to the object's center.
Before the Poisson~\cite{kazhdan2013screened} mesh reconstruction, we filter out the room's background by roughly cropping the dense point cloud around the object.
We run the Poisson surface reconstruction with parameters \texttt{depth=10, trim=1}.

\PAR{\realcap~\cite{capturereality}.}
Given the known camera extrinsics and intrinsics, we first obtain a sparse point cloud model using the "draft" mode, which downsamples the images by a factor of two. 
For the following processing steps, a reconstruction region is defined as a bounding box (roughly) around the camera positions. 
All scene parts outside the region are ignored. 
A 3D mesh is then computed for this region in "normal detail", which also uses images downsampled by a factor of two. 
This reconstruction stage first computes depth maps to obtain a dense point cloud, from which a mesh is then extracted. 
Typically, there are artifacts in the form of additional connected components. 
We only keep the largest connected component, which in all cases corresponded to (parts of) the object. 
The remaining mesh is then textured.

\PAR{Implicit Methods.}
Nerfacto~\cite{nerfstudio} is trained for 30K iterations without pose refinement, with normal prediction, and with the default parameters set by Nerfstudio~\cite{nerfstudio}.
iNGP~\cite{muller2022instant} is trained with the default parameters set by the authors for 30K iterations.
A mesh is extracted from the resulting models using the Poisson surface reconstruction~\cite{kazhdan2013screened}.

MonoSDF~\cite{yu2022monosdf}, VolSDF~\cite{yariv2021volume}, BakedSDF~\cite{yariv2023bakedsdf}, and Neus~\cite{wang2021neus} are trained with the default configuration set in SdfStudio~\cite{Yu2022SDFStudio}, NeuralAngelo~\cite{li2023neuralangelo} and Plenoxel~\cite{fridovich2022plenoxels} with the configurations set by the authors.
VolSDF and Neus are trained for 100k steps, MonoSDF for 200k iterations, BakedSDF for 250k iterations, NeuralAngelo for 500K iterations, and Plexoxel for 128K iterations, as recommended by their respective authors.
These methods output an \gls{sdf} from which the mesh is extracted using the marching cube algorithm~\cite{lorensen1987marching} with SDF values sampled on a 1024x1024x1024 grid and a subsampling factor of 8.
The NeuralAngelo~\cite{li2023neuralangelo} mesh is extracted with a 2048-resolution grid, and the Plexoxel~\cite{fridovich2022plenoxels} one with a 256-resolution grid,
 as recommended by their respective authors. %
For all methods, the SDF is initialized with a unit sphere.
The same setup is adopted for Unisurf~\cite{oechsle2021unisurf}, except that it outputs an occupancy grid instead of an SDF.

\PAR{iNGP~\cite{muller2022instant} implicit rendering.}
In addition to evaluating the mesh derived with iNGP~\cite{muller2022instant}, we also evaluate how well the novel view synthesis of iNGP can replace the rendering of CAD models for pose estimation.
We used the default settings of the iNGP renderer, except for the number of samples per pixel that is decreased to 1.
We edit the Megapose~\cite{labbe2022megapose} codebase to replace the CAD renderings with the iNGP implicit renderings.

\PAR{Hardware.}
To provide fair runtime comparisons, all reconstruction methods are run on a single NVIDIA
A100 GPU, 32x Intel Xeon-SC 8628, 24 cores, 2,9 GHz with 256GB of RAM.

One exception though is for RealityCapture~\cite{capturereality} that runs on a GeForce RTX 3060 with 12 GB of RAM, \ie, a weaker GPU than the other methods.
This exception is because RealityCapture required a Windows installation that we had available only on a machine equipped with GeForce.

\subsection{Pose Estimators: FoundationPose and Megapose}
\label{supp:pose_setup}

We recall how the two pose estimators used in the evaluation operate and refer the reader to their respective papers for more details.

\PAR{FoundationPose}~\cite{foundationposewen2024}\footnote{https://github.com/NVlabs/FoundationPose} is a generalizable render-and-compare pose estimator that takes as input an RGBD image with an object of interest and a 3D representation of that object to output a pose.

The pose estimation proceeds in 2 steps: it first generates a set of pose hypotheses all around the object that are later refined by a network.
This refinement network is given both the RGBD test image and the RGBD renderings from the pose hypotheses generated with the 3D representation.
A second network then ranks the refined pose hypotheses using the RGBD test image and the RGBD rendering produced from the refined poses.
The best-ranked pose is the final output.

One of the main contributions of FoundationPose is that the 3D representation can take two forms as long as color and depth renderings can be generated from it: a traditional CAD model or an implicit representation.
The pose estimation is agnostic to the 3D representation as it only uses the RGBD renderings.
Also, it is jointly trained with renderings from the CAD models and the implicit representations, which reduces any possible distribution shift between the two types of renderings.

Whenever the 3D representation of the object at hand is not available, FoundationPose generates a 3D implicit representation of the object at test time.
Thus FoundationPose can be deployed on any object even when the 3D representation is not known beforehand, which reflects the practical conditions in which pose estimators are deployed.

The 3D implicit model can generate RGBD renderings in two ways: either with novel-view synthesis directly or by first extracting a mesh out of the implicit representation and then rastering the mesh.
The latter is computationally more efficient, as observed by the FoundationPose authors~\cite{foundationposewen2024}.
We adopt a similar derivation where we extract a mesh out of the evaluated implicit 3D reconstruction methods to produce RGBD renderings.

\begin{figure*}[t]
   \centering
\includegraphics[width=\linewidth,height=\textheight,keepaspectratio]
  {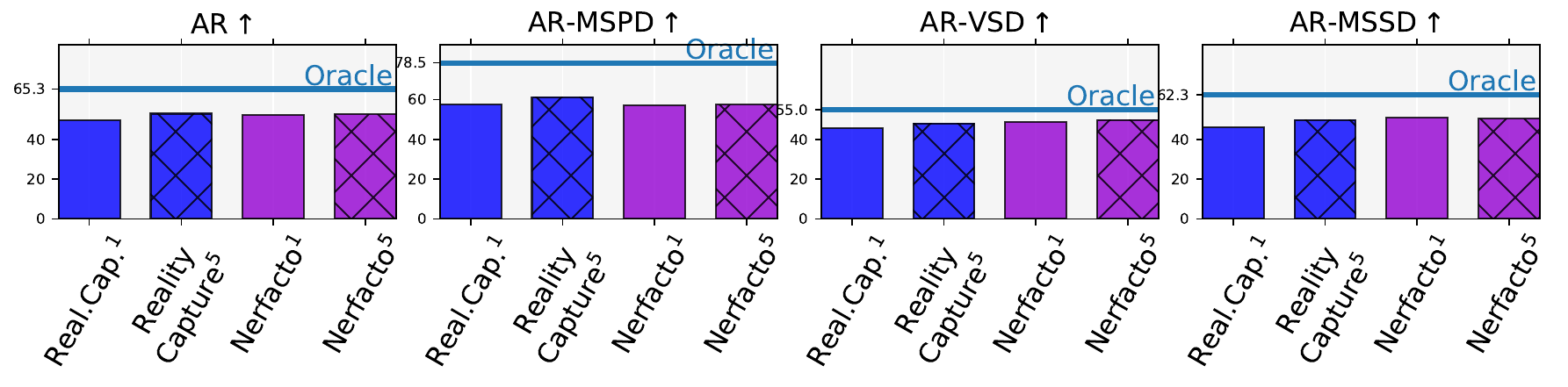}
   \caption{
     \textbf{Megapose Parameter Tuning}: comparison of the pose performance when keeping the top-1 (uniform bar) and top-5 (crossed bar) pose hypothesis for pose refinement.
     The experiment runs on two of the best-performing reconstruction methods RealityCapture~\cite{capturereality} and 
     Nerfacto~\cite{nerfstudio}.
     The results show that using a higher number of coarse hypotheses marginally
     improves the scores for RealityCapture (+3\%) and barely affects Nerfacto.
   }
   \label{fig:avg_ar_h1vsh5}
\end{figure*}

\PAR{Megapose}\cite{labbe2022megapose}\footnote{https://github.com/megapose6d/megapose6d} is a also generalizable render-and-compare pose estimator that takes as input an RGB image depicting an object and a 3D model of that object and outputs the pose of the object.
Megapose is made of two modules: a coarse pose estimator and a pose refiner.
Given a test image cropped around the object, %
\textbf{the coarse module} generates  $n$ pose hypothesis all around the objects from which renderings are generated.
The concatenation of the test image and the renderings are fed to a classification network which logits are interpreted as a score for each pose hypothesis.
The top-$K$ coarse poses are kept and refined: for each coarse pose, the refiner renders the object's mesh from that coarse pose and from three additional views. 
The views aim at creating parallax and are generated by moving the camera laterally while ensuring that the camera-z axis goes through the object's centroid. 
\textbf{The refiner network} is a regression network that takes as input the test image and the four renders defined previously and outputs a pose update to apply to the coarse pose. 
The final pose is the composition of the coarse pose with the pose update. This final pose can also be refined by repeating the refinement step.
We adopt the author's recommendation and set the number of refinement iterations to 5. 
We feed the coarse module with the maximal number of initial pose hypothesis $n=576$ and we keep the top-1 coarse pose, \ie, setting $K=1$, since we observe marginal improvement with $K=5$ (see \fig~\ref{fig:avg_ar_h1vsh5}) and it runs faster.

Before being fed to the network, the image is cropped around the object of interest.
One can either use a box detector or the ground-truth box to define the region of interest and we use the latter for both Megapose and FoudationPose.

\subsection{BOP Evaluation}
\label{supp:pose_metrics}

As is custom for object pose estimation, we report the three standard pose errors defined in the BOP benchmark~\cite{hodan2018bop,hodavn2020bop}.
The metrics reported are the Visible Surface Discrepancy (VSD), the Maximum Symmetry-Aware Surface Distance (MSSD), and the Maximum Symmetry-Aware Projection Distance (MSPD). 
We refer the reader to the well-documented BOP benchmark methodology for details\footnote{https://bop.felk.cvut.cz/challenges/bop-challenge-2019/\#evaluationmethodology}.
We recall the definitions here for the sake of completeness.
\textbf{Note that all these errors measure the pose error, hence the lower, the better.}

The \textbf{Maximum Symmetry-Aware Surface Distance (MSSD)}
measures the maximum surface misalignment between the surface of the object when it is positioned with the ground-truth pose and when it is positioned with the estimated pose. In both cases, the object is positioned with respect to the camera coordinate frame.
As indicated in the name, the MSSD is symmetry-aware, \ie, it does not penalize
the estimated pose for ambiguous symmetries (\eg a textureless sphere has an
infinite number of non-resolvable symmetries). 
\textit{Derivation details:} 
Let $M$ be the CAD object model, %
$\hat{\mathbf{P}}$ the estimated pose, $\bar{\mathbf{P}}$ the ground-truth pose, $V_M$ a set of points sampled on the reference CAD model $M$ (\ie the YCB-V~\cite{xiang2018posecnn,calli2015ycb} one), $S_M$ the set of symmetries of the object. The MSSD is computed as:
%
%
\begin{multline*}
    e_{\text{MSSD}}(\hat{\mathbf{P}}, \bar{\mathbf{P}}, S_M, V_M) = \\
    \text{min}_{\mathbf{S}\in S_M} \text{max}_{\mathbf{x}\in V_M} \|\hat{\mathbf{P}}\mathbf{x} - \bar{\mathbf{P}} \mathbf{S} \mathbf{x} \|_2 \enspace .
\end{multline*}
The metric is computed over the set of points $V_M$ sampled on the reference mesh positioned with the ground-truth pose and with the estimated pose. 
Corresponding points form pairs between which the distance is computed.
The MSSD reports the maximum distance over all pairs.
To account for symmetries, the derivation computes several distances for each pair of points where the ground-truth point is additionally transformed with an isometry to account for the symmetry ($\text{min}_{\mathbf{S}\in S_M}$).
The actual distance between two points in the minimum over all symmetries.
The symmetries are provided by the evaluation dataset as annotations.
\textbf{Interpretation:} as an upper bound on the surface misalignment, the MSSD is very sensitive as even a small pose error can induce high surface discrepancies around high-curvature
regions of the object.
In practice, such high-curvature regions are suitable grasping points for \textbf{robotic manipulation}, so a high AR-MSSD (Average Recall-MSSD, see below) suggests that the object pose estimation is suitable for robotic grasping.

The \textbf{Maximum Symmetry-Aware Projection Distance (MSPD)}
measures the maximum pixel displacement induced by an inaccurate
pose estimate when rendering the object.
\textit{Derivation details:}
The notation is the same as for the MSSD metric.
Let $M$ be the CAD object model, $\hat{\mathbf{P}}$ the estimated pose, $\bar{\mathbf{P}}$ the ground-truth pose, $V_M$ a set of points sampled on the reference CAD model $M$ (\ie the YCB-V~\cite{calli2015ycb,xiang2018posecnn} one), and $S_M$ the set of symmetries of the object.
The MSPD is computed as:
%
\begin{multline*}
    e_{\text{MSPD}}(\hat{\mathbf{P}}, \bar{\mathbf{P}}, S_M, V_M)  = \\ 
    \text{min}_{\mathbf{S}\in S_M} \text{max}_{\mathbf{x}\in V_M}
    \|\text{proj}(\hat{\mathbf{P}}\mathbf{x}) - \text{proj}(\bar{\mathbf{P}} \mathbf{S} \mathbf{x}) \|_2 \enspace .
\end{multline*}
As for the MSSD, the metric is computed on the set of points $V_M$ sampled on the reference CAD model when it is positioned with the ground-truth pose and with the estimated poses, respectively.
The points are projected onto the image plane and the MSPD reports the maximum distance between the projected points over all pairs.
\textbf{Interpretation:}
Note that the projective nature of the MSPD makes it unsuitable for applications that require physical interactions with the world.
Instead, this perceptual error is better suited for vision applications such as \textbf{Augmented and Virtual Reality} that exploit the rendering of positioned objects.

\begin{table*}[t]
\footnotesize
\begin{center}
\begin{tabular}{p{1.5in}p{3.5in}}
\toprule
     Category & Objects \\
    \midrule
    Lambertian-Textured-Large & 02-cracker-box, 03-sugar-box, 16-wooden-block \\
     \midrule
    Lambertian-Textured-Small & 07-pudding-box, 08-gelatin-box, 18-large-marker \\
     \midrule
     Shiny-Textured &  01-master-chef-can, 04-tomato-soup-can, 06-tuna-fish-can, 09-potted-meat-can,13-bowl, 14-mug \\
     \midrule
     Uniform-Texture & 10-banana, 11-pitcher-base, 17-scissors, 19-large-clamp, 20-extra-large-clamp, 21-foam-brick \\
    \midrule
    Low-Texture & 05-mustard-bottle, 12-bleach-cleanser \\
    \midrule
     Scissors-Like & 17-scissors, 19-large-clamp, 20-extra-large-clamp \\
    \midrule \midrule
     Legacy-Objects & 03-sugar-box, 04-tomato-soup-can, 10-banana, 12-bleach-cleaner, 13-bowl, 14-mug, 16-wooden-block, 17-scissors, 18-large-marker, 19-large-clamp, 20-extra-large-clamp, 21-foam-brick\\
    \midrule
    Updated-Objects & 01-master-chef-can, 02-cracker-box, 05-mustard-bottle, 06-tuna-fish-can, 07-pudding-box, 08-gelatin-box, 09-potted-meat-can, 11-pitcher-base, 15-power-drill\\
\bottomrule
\end{tabular}
  \caption{\textbf{Object Categories.} The categories are characteristic of
  object properties such as size, shape, texture, and materials.
  We also distinguish between the objects that are the same between the YCB-V datasets~\cite{calli2015ycb,xiang2018posecnn} and our data collection (`legacy') and the ones that got updated by their manufacturer ('updated').
  }
\label{tab:object_categories}
\end{center}
\end{table*}

The \textbf{Visible Surface Discrepancy (VSD)} measures the average misalignment of the visible surface of the object along the camera-z axis.
\textit{Derivation details:} 
Let $\hat{V}$ and $\bar{V}$ be visibility masks, \ie, the set of pixels where the reference model $M$ is visible in the image when the model is positioned at the estimated pose $\hat{P}$ and the ground-truth pose $\bar{P}$.
Let $\hat{D}$ and $\bar{D}$ be the distance maps obtained by rendering the reference object model $M$ in the estimated pose $\hat{P}$ and the ground-truth pose $\bar{P}$, respectively. The VSD is computed as:
%
\begin{multline*}
  e_{\text{VSD}}(\hat{D}, \bar{D}, \hat{V}, \bar{V}, \tau)= \mathrm{avg}_{p\epsilon \hat{V} \cup  \bar{V}}\\
      \begin{cases}
        0 &\text{if}~p \in \hat{V} \cap \bar{V} \wedge \| \hat{D}(p) - \bar{D}(p)\| \leq \tau \\
        1 & \text{otherwise }
      \end{cases} \enspace .
\end{multline*}
The VSD is also a perceptual metric but the definition of the distance is slightly different.
It measures a distance in 3D but contrary to the MSSD, it does not measure it between the corresponding points sampled on the mesh. 
Instead, the pairs are formed by taking a visible point from each mesh that projects onto the same pixel.
In practice, the CAD model is rendered from the two poses to generate distance maps (\ie, depth maps) and the VSD 
corresponds to the average number of misplaced rendered pixels.
Misplacement can either mean an incorrect distance map value or an incorrect position in the image, \eg, if the estimated pose translates the object too much in the camera-x or camera-y directions, the rendered pixels will be disjoint on the image space. 
\textbf{Interpretation:} a low VSD indicates that the object is
well positioned, yet can miss to report errors in the estimated pose's rotation.
If the object is mostly symmetric, a low VSD can indicate that the centroid of the object is well estimated but it provides less information on the rotation's error.
For example, take the 14-mug object and assume that the estimated pose is the ground-truth pose flipped so that the mug is upside-down.
Then the position of the CAD models positioned with the ground-truth pose and the estimated pose differ only on the side in which the 14-mug handle is. 
Instead, the surface of the 14-mug's "body" will be aligned.
Thus, the distance maps of these points will be mostly equal and will dominate the metric, indicating a good position.
Thus, the VSD is better suited \textbf{for applications with tolerance to such edge-cases such as robotic navigation.}

\textbf{Throughout the evaluation, the metrics are derived on the same model $M$ which is the original CAD model provided in the YCB-V~\cite{calli2015ycb,xiang2018posecnn} dataset, \ie, the reconstructed meshes are not used when computing the metrics.} 
This ensures that the results for poses obtained from different meshes are comparable. 

Note that the BOP benchmark does not report these errors as is but reports the
\textbf{Average Recall (AR)} on these errors. %
Given an error threshold, the recall is the fraction of the estimated poses for which the 
error falls below the threshold. 
\textbf{Since these metrics quantify the pose error, the lower, the better.}
The recall is computed separately over each metric and as usual \textbf{for recalls, the higher is better}.
The Average Recall (AR) for a given metric, which we report as AR-VSD, AR-MSPD,
AR-MSSD, is the average of several recalls computed over a range of error
thresholds.
Finally, we also report the \textbf{global average recall AR} computed as the average
of the three average recalls AR-VSD, AR-MSPD, AR-MSSD. As specified in the BOP
benchmark, it measures a method's overall performance.

The recall averaging used in the benchmark slightly differs from the averaging of the BOP benchmark.
Instead of averaging the metrics over the set of all test images, we averaged the recall over the test images depicting a given object instance (\eg 02-cracker-box) to output the object-specific AR-MSPD, AR-MSSD, AR-VSD, AR.
The global performance of the method is the average of these metrics over the set of objects. 
The performance of the methods of specific object categories is the average of the subset of objects that make that category.
Because of the change in the averaging order, the metrics we report can be slightly lower than the ones reported in the BOP benchmark by about 5\% on average.
This edit is not critical as we are interested in the relative pose performances rather than the absolute ones.

\subsection{Object Categorization}
\label{subsec:object-categories}

We split the YCB-V objects into categories with specific properties depending on their size, their texture and their materials (Tab.~\ref{tab:object_categories}).

\textbf{Lambertian-Textured-Large / Small} refers to large / small objects strongly textured, with diffuse reflections.

\textbf{Shiny-Textured} objects are medium and small objects which material has high reflectivity.

\textbf{Uniform-Texture} objects have no texture, \ie, they are color-uniform whereas \textbf{Low-texture} objects have both high-texture areas and textureless areas. 

\textbf{Scissors-like}: Objects with scissors-like shapes that may have holes inside the mesh and have symmetries from different views. %

\textbf{Updated Objects:} 
The YCB-V~\cite{xiang2018posecnn} are made of objects commonly found in grocery stores.
However, brands regularly update their product packaging, which means that the object's texture gets updated.
Some of the original YCB-V~\cite{xiang2018posecnn} objects are not available on the market anymore so we use the alternatives suggested by the official YCB~\cite{calli2015ycb} website\footnote{https://www.ycbbenchmarks.com/object-set/}.
The newer versions of the objects differ in texture and scale, with an average change estimated at 4\% of the object's dimensions. 
As for the texture update, the 01-master-chef-can, 11-pitcher-base, and 15-power-drill underwent full texture updates and other objects have only minor updates.

\textbf{Legacy Objects} are the objects that are the same between the YCB-V~\cite{xiang2018posecnn} dataset and our data collection, \ie, all objects minus the updated objects.

\begin{figure*}[t]
   \centering
  \includegraphics[width=\linewidth,height=\textheight,keepaspectratio]
  {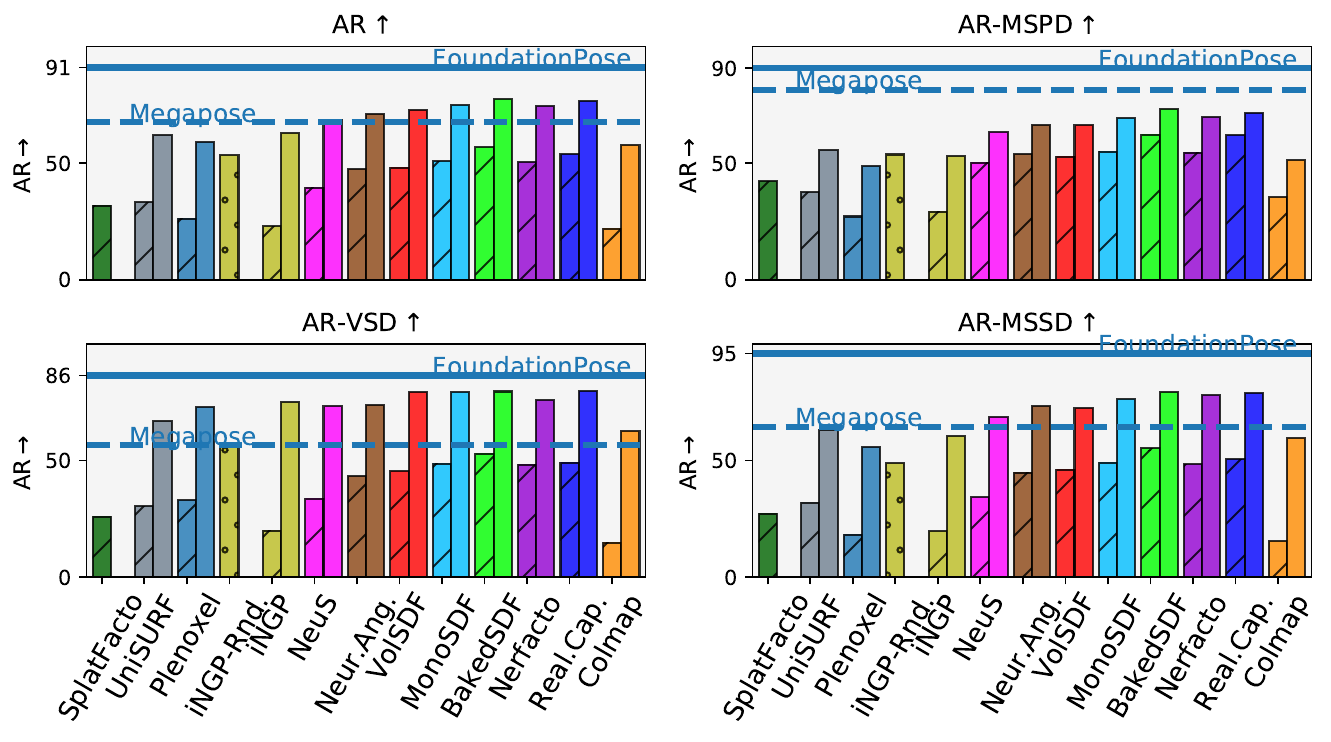}
   \caption{
     \textbf{Detailed Pose Evaluation of the 3D reconstructions.}
   We measure the performance of FoundationPose~\cite{foundationposewen2024} (uniform bar) and Megapose~\cite{labbe2022megapose} (lined bar) when replacing the CAD models with the 3D reconstructions.
   The blue lines indicate the performance of the pose estimators when using the original CAD models (FoundationPose: \textbf{\color{RoyalBlue}full line}, Megapose: \textbf{\color{RoyalBlue}dashed line}).
Overall, a gap remains between CAD models and 3D reconstructions for pose estimation.
This gap is smaller on the VSD, which indicates that 3D reconstructions can be suitable for pose estimation in applications with tolerance to pose errors.   
   }
   \label{fig:avg_ar_detail}
\end{figure*}

\section{Additional Results}
\label{supp:sec_results}

We report the quantitative results that support the conclusions drawn in the main paper and that could not fit due to the page limit: 
the average recall on each of the three pose errors (Sec.~\ref{supp:ar}),
the influence of the size of the training data on the performance (Sec.~\ref{supp:pose_vs_size}),
further results on the relation between the object categories and the pose performance (Sec.~\ref{supp:pose_vs_object}),
the importance of the texture for pose estimation (Sec~\ref{supp:texture}),
the detail of the pose performances per objects (Sec~\ref{supp:ar_vs_object}),
and qualitative results showing renderings from the reconstructed meshes (Sec~\ref{supp:quali}).

\subsection{Pose Evaluation: Average Recall}
\label{supp:ar}

The BOP benchmark defines multiple errors relevant to assess how suitable a reconstruction is for a given application such as augmented and virtual reality (MSPD), navigation (VSD), and object manipulation (MSSD)~\cite{hodan2018bop,hodavn2020bop,bopbenchmark}.
In the main paper, the pose performances are measured with the Average Recall (AR) averaged over all three VSD, MSSD, and MSPD errors
We now report the AR for each error separately (Fig.~\ref{fig:avg_ar_detail}) to support the conclusions of the main paper (Sec.5 "Finer Pose Evaluation").

The performance gap between Megapose~\cite{labbe2022megapose} and FoundationPose~\cite{foundationposewen2024} when they are evaluated with the original CAD models is relatively smaller for the MSPD than for the other errors.
One interpretation is the improvement of FoundationPose over Megapose is particularly useful for 3D spatial tasks, as measured by the VSD and the MSSD, \eg, for navigation and object manipulation.
The improvement is smaller for tasks involving object rendering for which Megapose is close to FoundationPose.

The main paper mentions the performance drop induced when replacing the original CAD models with the reconstructed ones and how it is consitent between the two pose estimators for the best reconstruction methods, \ie, RealityCapture~\cite{capturereality}, Nerfacto~\cite{nerfstudio}, VolSDF~\cite{yariv2021volume}, MonoSDF~\cite{yu2022monosdf}, BakedSDF~\cite{yariv2023bakedsdf}, and NeuralAngelo~\cite{li2023neuralangelo}.
This remains the case for the individual ARs and we also observe that the relative performances of the reconstruction methods are consistent across the different errors.
So one can expect that improving the 3D reconstructions might benefit all three pose estimation errors, hence various applications involving object pose estimation.

One interesting observation is the gap between the original CAD models and the best reconstruction methods is relatively small on the VSD compared to the other errors.
The VSD indicates that the object is overall well positioned so this suggests that 3D reconstructions can be suitable replacements for the CAD model when the pose estimator is used in tasks that are tolerant to the pose errors, \eg, visual navigation.
However, the gaps on the MSSD and MSPD remain large so there is room for improvement for 3D reconstructions to replace CAD models in tasks with low tolerance in the pose errors, \eg, object manipulation.

\subsection{Reconstruction with varying training size}
\label{supp:pose_vs_size}

\begin{figure*}[t]
   \centering
    \includegraphics[width=0.90\linewidth,height=\textheight,keepaspectratio]
    {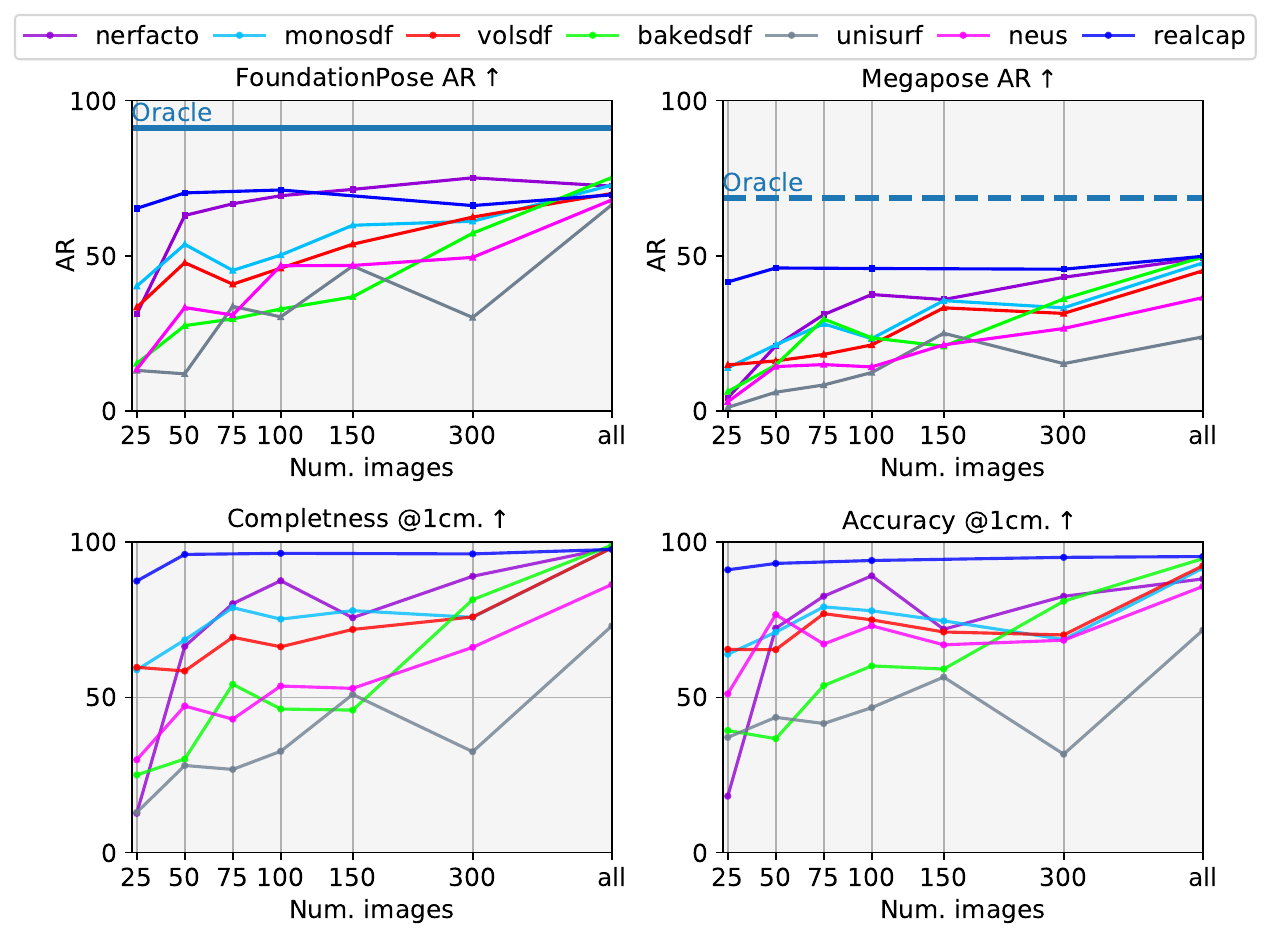}
   \caption{
    \textbf{Pose Evaluation of the 3D reconstructions from subsets of images.}
   \textbf{Top:} We measure the performance of FoundationPose~\cite{foundationposewen2024} and Megapose~\cite{labbe2022megapose} when replacing the CAD models with the 3D reconstructions generated from image sets of varying size.
   The blue lines indicate the performance of the pose estimators when using the original CAD models (FoundationPose: \textbf{\color{RoyalBlue}full line}, Megapose: \textbf{\color{RoyalBlue}dashed line}).
     As expected, the more data the better the results and 75-100 images are enough to get reasonable reconstructions.
     The performance of Reality Capture~\cite{capturereality} is relatively stable as the number of images decreases: \textbf{the reconstruction from 50 images takes as little as 2 minutes} and is already close to the best RealityCapture performance.
     \textbf{Bottom:} Evaluation of the reconstructed geometry with the traditional completeness and accuracy metrics.
   }
   \label{fig:ar_vs_num_images}
\end{figure*}

Besides the quality of the pose estimation, an important property of the reconstruction methods is their data requirement.
To evaluate the `data-greediness' of the reconstruction methods, we reconstruct the YCB-V~\cite{xiang2018posecnn} objects from subsets of images sampled uniformly around the object~\cite{gonzalez2010measurement}.
\fig~\ref{fig:ar_vs_num_images} complete the results in the main paper with the pose performance of FoundationPose~\cite{foundationposewen2024} with reconstructions from image subsets and the 3D reconstruction's geometric accuracy.

As for Megapose~\cite{labbe2022megapose}, the performance of FoundationPose~\cite{foundationposewen2024} improves as more images are used for the 3D reconstruction (top-left).
Nerfacto~\cite{nerfstudio} plateaus with only 75-100 images and the performance of RealityCapture~\cite{capturereality} remains relatively stable as the number of images decreases to 25.
This further demonstrates the practical advantage of RealityCapture that is both fast and data-efficient over other reconstruction methods: the reconstruction takes less than 1 min. on 25 images and $\sim$2 min. on 50 images.
Still, there remains a gap between the 3D reconstructions and the original CAD models for pose estimation that calls for further improvements in 3D reconstruction.

We observe that FoundationPose~\cite{foundationposewen2024} seems more resilient to innacurate 3D reconstructions than Megapose~\cite{labbe2022megapose}.
For example, the geometric quality of Nerfacto~\cite{nerfstudio} drops slightly at 150 images which is reflected in the Megapose performance but not in the FoundationPose ones.
A similar observation holds for BakedSDF~\cite{yariv2023bakedsdf} at 100 and 150 images.

\subsection{Object Categories}
\label{supp:pose_vs_object}

\begin{figure*}[t]
   \centering
\includegraphics[width=\linewidth,height=\textheight,keepaspectratio]
{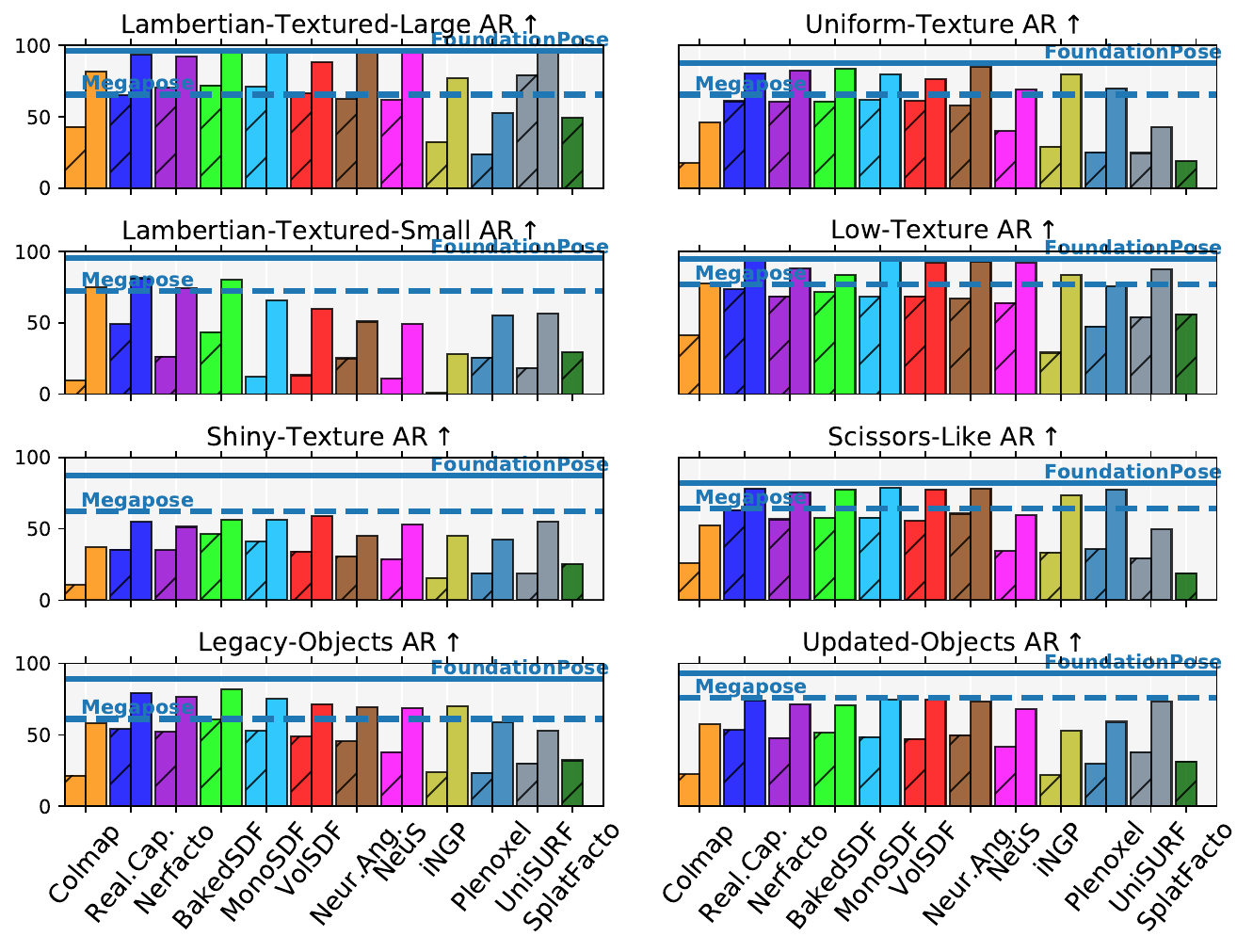}
   \caption{
   \textbf{Pose Evaluation of the 3D reconstructions for various Object Categories.}
   We measure the performance of FoundationPose~\cite{foundationposewen2024} (uniform bar) and Megapose~\cite{labbe2022megapose} (lined bar) when replacing the CAD models with the 3D reconstructions.
   The blue lines indicate the performance of the pose estimators when using the original CAD models (FoundationPose: \textbf{\color{RoyalBlue}full line}, Megapose: \textbf{\color{RoyalBlue}dashed line}).
    Some categories we expected to be challenging, such as objects with uniform texture are not.
   Instead, the most difficult categories are small and shiny objects.  
   }
   \label{fig:ar_vs_categories2}
\end{figure*}

We split the YCB-V~\cite{xiang2018posecnn} objects into 8 groups defined in Sec.~\ref{subsec:object-categories}  based on their shape, texture, material properties, and the degree of change between the collected objects and the original ones.
We report results for all object categories in \fig~\ref{fig:ar_vs_categories2} (including the 3 reported in the main paper) and observe that the relative pose performance across object categories is telling of some of the current challenges in 3D reconstruction.

The variation in pose performance is relatively large between the most simple categories and the most challenging ones.
Simple object categories include large objects with Lambertian texture and objects with little or uniform texture.
For these objects, replacing the CAD model with the 3D reconstructions has little or no impact on the performances.
We also note that some of the reconstruction methods that perform relatively lower than others, \eg Neus~\cite{wang2021neus}, NeuralAngelo~\cite{li2023neuralangelo}, achieve very good results on some of the simple object categories.
Another interesting result is the high performance of the MVS-based RealityCapture~\cite{capturereality} on objects with little texture: one could have expected lower results since MVS relies on feature matching which accuracy usually drops when the image content exhibits little texture.

As mentioned in the main paper, the most challenging object categories are small textured objects and objects with a `shiny' texture, \ie, a texture with high reflectance.
For these categories, the performance bottleneck often lies in the texture.
The resolution of small reconstructed objects is relatively lower than for larger objects and that loss of information can impede the pose estimation.
As for shiny objects, the reflection of light sources on the object during the data collection can cause visual discrepancies between the reconstruction and the object in the test images.

Since the YCB-V dataset~\cite{calli2015ycb,xiang2018posecnn} is made of objects commonly found in grocery stores, whenever a brand updates its packaging, the texture gets updated.
The YCB-V objects we captured hence fall into two categories: the `legacy' objects which texture did not change and the `updated' objects which texture has changed.
Only the coffee box `01-master-chef-can', the '11-pitcher-base, and the '15-power-drill' have undergone a full texture update and the other objects have undergone relatively minor updates.
Still, the gap between the 3D reconstructions and the CAD models is lower on the `legacy' objects than the `updated ones'.
BakedSDF~\cite{yariv2023bakedsdf} is even on par with the original CAD models with Megapose~\cite{labbe2022megapose}.
There still remains a gap between 3D reconstructions and CAD models though.
Note that the integration of a given 3D reconstruction with FoundationPose~\cite{foundationposewen2024} often outperforms or is on par with Megapose~\cite{labbe2022megapose} running with the original CAD model: this suggests that the gap between CAD models and 3D reconstructions could be closed not only by improving 3D reconstructions but also pose estimators.

\subsection{Influence of the Texturing Algorithm}
\label{supp:texture}

\begin{figure*}[t]
   \centering
\includegraphics[width=\linewidth,height=\textheight,keepaspectratio]
{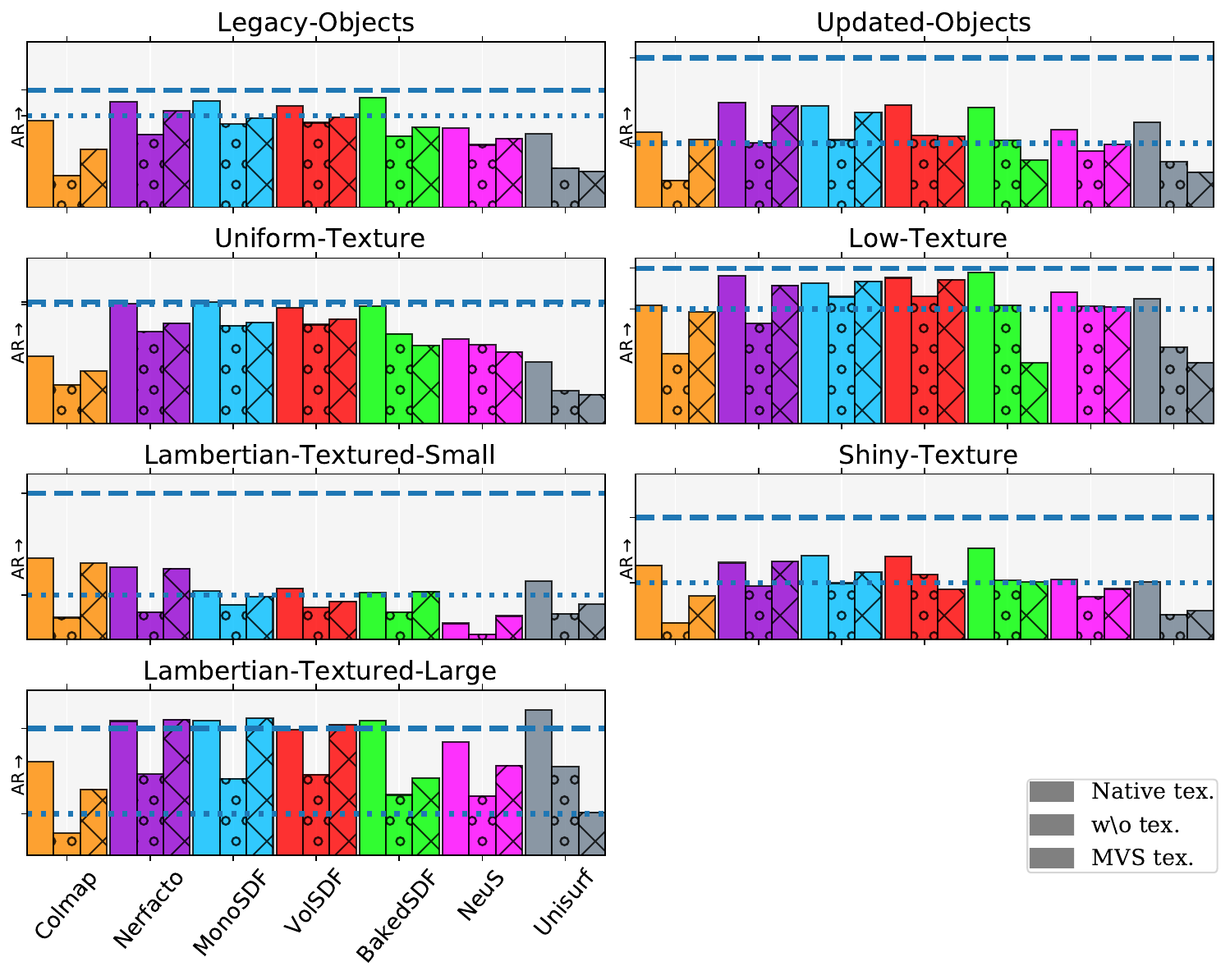}
   \caption{
   \textbf{Pose Evaluation of the 3D reconstructions under different texturings.}
    We measure the performance of Megapose~\cite{labbe2022megapose} when replacing the CAD models with 3D reconstructions.
   The reconstructions are compared to the original CAD models with texture {\color{RoyalBlue}\textbf{(dashed line)}} and without texture {\color{RoyalBlue}\textbf{(dotted line)}}.
   The performances of the textureless CAD models and the textureless 3D reconstructions are on par, which suggests that the 3D reconstructions have geometry suitable for pose estimation.
   }
   \label{fig:ar_vs_texture}
\end{figure*}

Previously, we observed that the performance of the pose estimation drops on small objects or shiny objects and we argued that the performance bottleneck lies in the texturing.
So we now analyze the influence of the texture on pose estimation (\fig~\ref{fig:ar_vs_texture}).
In the first experiment, we compare the \textbf{pose performance of the 3D reconstructions with and without texture}.
A second experiment assesses the influence of the texturing algorithm by \textbf{comparing the 3D reconstruction with native texturing and MVS-texturing}~\cite{Waechter2014Texturing}.
These reconstructions are compared to the original CAD models with texture {\color{RoyalBlue}\textbf{(dashed blue line)}} and without texture {\color{RoyalBlue}\textbf{(dotted blue line)}}.

We run this ablation on Megapose~\cite{labbe2022megapose} with a subset of methods:
Nerfacto~\cite{nerfstudio}, VolSDF~\cite{yariv2021volume}, MonoSDF~\cite{yu2022monosdf} and BakedSDF~\cite{yariv2023bakedsdf}, the colored but textureless reconstructions from COLMAP~\cite{schonberger2016structure,schonberger2016pixelwise}, Neus~\cite{wang2021neus} and UniSURF~\cite{oechsle2021unisurf}.

We compare the reconstructions under three texturings: the texturing \textbf{native} to each reconstruction method, \textbf{no texturing} at all, and texturing with the \textbf{MVS-texturing} algorithm~\cite{Waechter2014Texturing}.
Note that COLMAP does not texture the models but only provides vertex colors so for COLMAP we compare the mesh colored with the point cloud colors, the colorless mesh, and the mesh textured with MVS-texturing.
Native texturing refers to the texturing algorithm as implemented in NerfStudio~\cite{nerfstudio} and SDFStudio~\cite{Yu2022SDFStudio}: the UV map is generated by querying the network for the color at the position of the mesh’s faces.
The textureless mesh has exactly the same geometry as the textured one but without the texture map and with no color. 
Finally, the MVS-textured mesh is the output of the MVS-texturing algorithm~\cite{Waechter2014Texturing} run on the geometry of the reconstructed mesh.

\PAR{With vs. Without Texture (circles).}
For all object categories, the gap between the textureless CAD model (dotted line) and the textureless reconstruction (circle bars) is relatively small or non-existent.
This is the case even for the object categories that are the most challenging under the regular pose evaluation (small and shiny), which supports the assumption that the performance bottleneck for these categories is the texture.
This suggests that the geometric quality of the 3D reconstructions and the CAD models are comparable as measured by pose estimation performance.
This is in line with the results obtained with traditional geometric evaluation of the 3D reconstructions measured with the completeness and accuracy metrics.

When the texture is removed, whether from the CAD models or the 3D reconstructions, the pose performance drops less for objects with low or uniform texture than for textured objects.
A reasonable explanation is that the pose estimation relies more on the geometry than on the texture for objects with little texture information.
Hence, the absence of texture or even incorrect texturing may not be prohibitive for pose estimation on such objects as long as the reconstructed geometry is good enough.

\begin{figure}
    \centering
    \includegraphics[height=0.24\textheight,keepaspectratio]
    {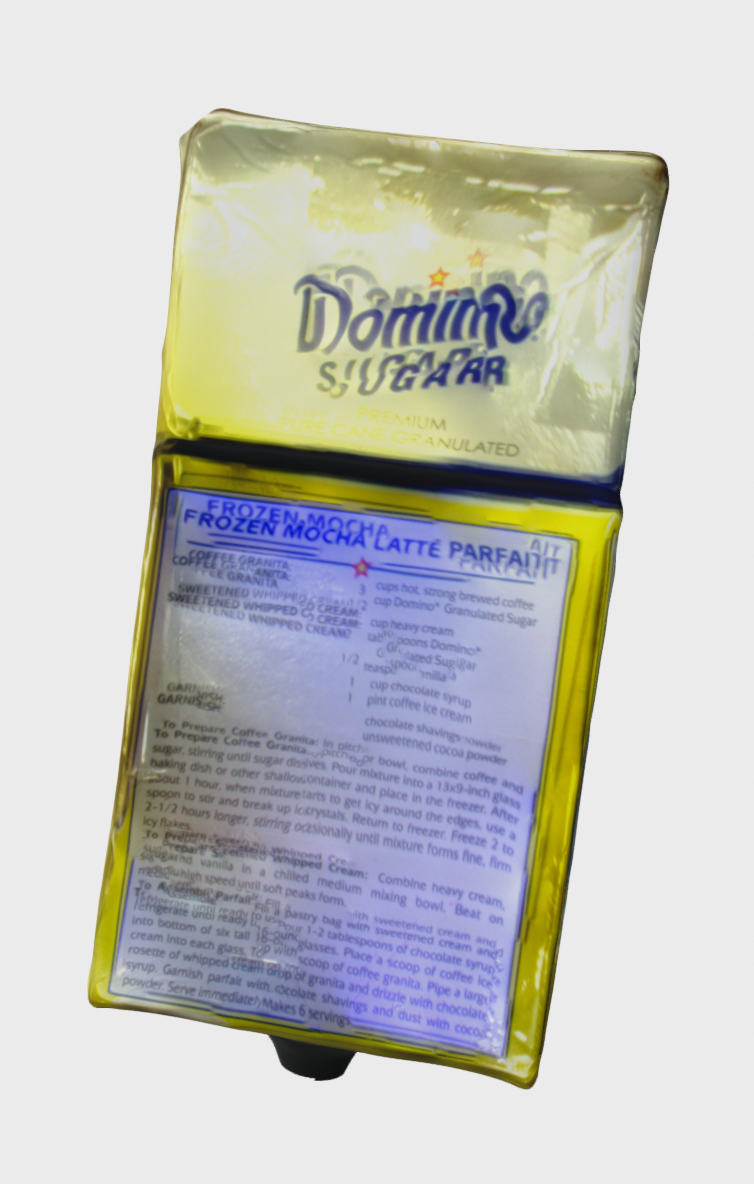}
      \includegraphics[width=0.74\linewidth,height=\textheight,keepaspectratio]{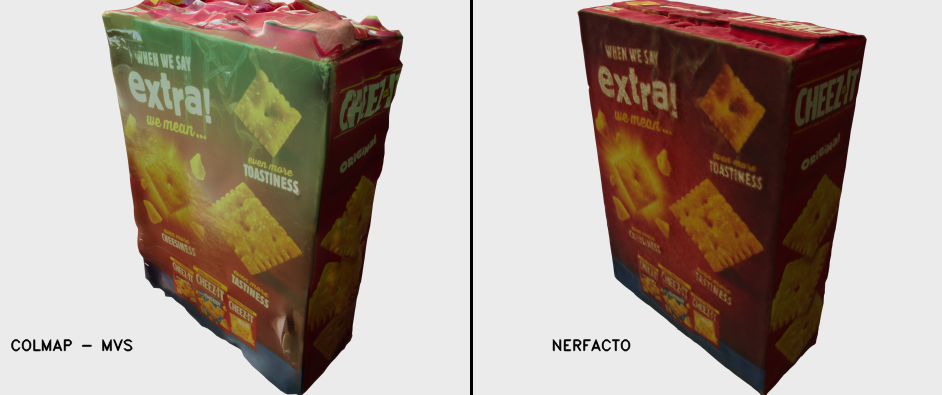}
    \caption{Examples of texturing artifacts generated with MVS-texturing~\cite{Waechter2014Texturing}.
    \textbf{Left}: Illustration of MVS-Texturing mirage artifact.
    The text in the 03-sugar-box is distorted and repeated. We also observe color saturation on the blue separation line between the top of the box and the bottom.
    \textbf{Right}: Illustration of the MVS-Texturing color artifacts. The top
   of the 02-cracker-box on the left (COLMAP + MVS-Texturing) exhibit green areas that do not exist in the original model and are not produced by native texturing
   of other methods (\eg Nerfacto on the right).}
    \label{fig:texrecon}
\end{figure}

\PAR{Native vs. MVS-Texturing.}
The texturing native to each method (uniform bar) leads to either comparable or better results than MVS-texturing~\cite{Waechter2014Texturing} (crossed bar).
Even the textureless but colored meshes of COLMAP~\cite{schonberger2016structure,schonberger2016pixelwise} leads to better pose performance than with the MVS-texturing.
This may appear counter-intuitive since MVS-Texturing~\cite{Waechter2014Texturing} produces sharper and more visually appealing textures (see \fig~\ref{fig:texrecon}-right).
However, render-and-compare methods usually operate on low-resolution renderings so that they can be fed efficiently to the network, so the lower resolution of the native texturing is not penalized.
Also, MVS-Texturing~\cite{Waechter2014Texturing} can generate strong color artifacts, such as color saturation or color mismatch that decrease the texture's quality (see the green tint as shown in~\fig~\ref{fig:texrecon}-right).
That could be due to multiple reasons:
noise in the camera pose which leads to blurred texture (\fig~\ref{fig:texrecon}-left), the surface extracted from SDFs may not been extracted with the optimum isosurface value, or MVS-Texturing can not handle well reflections. 

\subsection{Per-Object Pose Evaluation}
\label{supp:ar_vs_object}

\begin{figure*}[t]
   \centering
\includegraphics[width=0.75\linewidth,height=\textheight,keepaspectratio]
{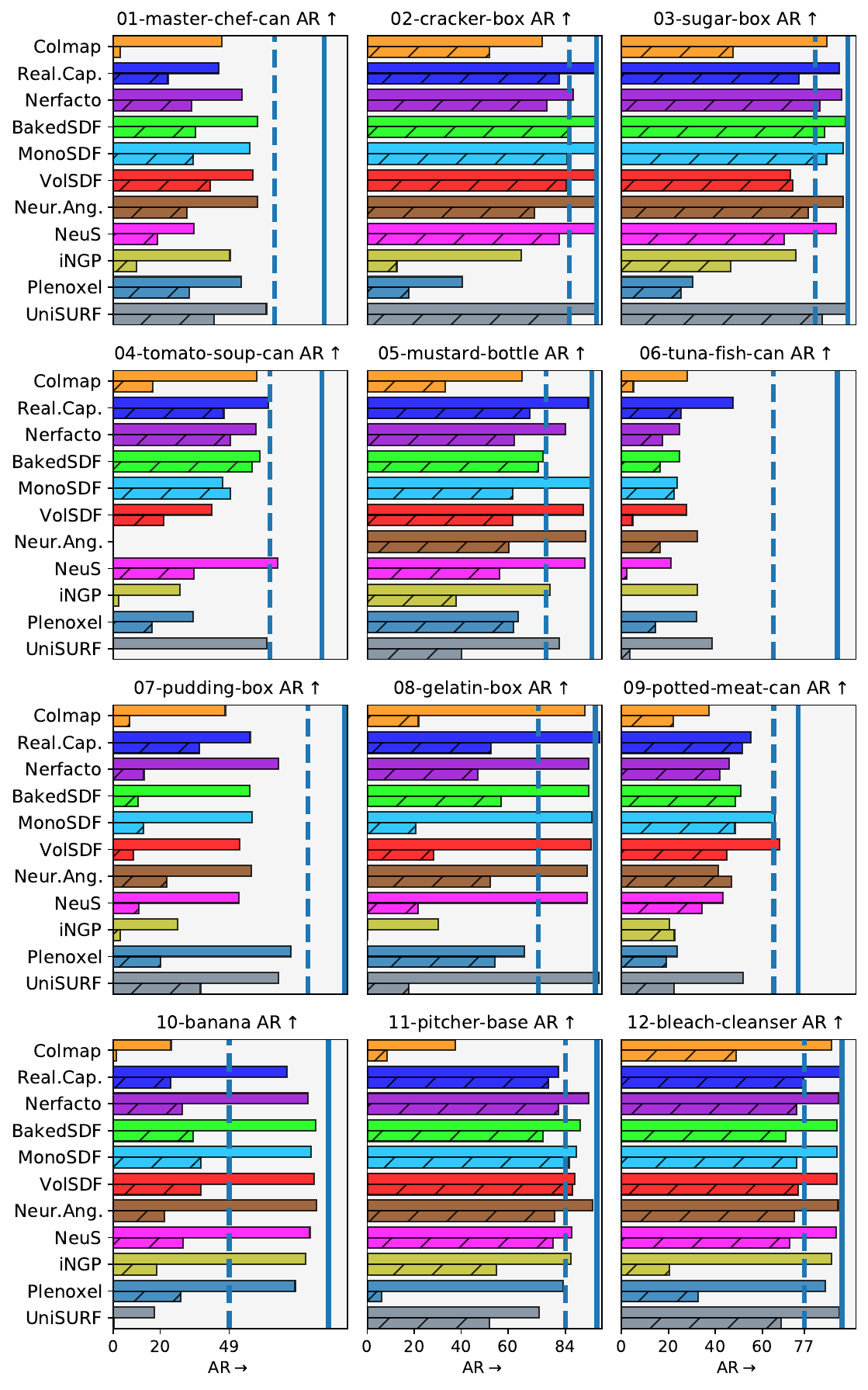}
   \caption{\textbf{Pose Evaluation of the 3D reconstructions for each YCB-V object~\cite{xiang2018posecnn} (1/2).}
   The 3D reconstructions are evaluated based on the performance of two pose estimators, \textbf{FoundationPose~\cite{foundationposewen2024} (uniform bar)} and \textbf{Megapose~\cite{labbe2022megapose} (lines bar)}, when the CAD model is replaced with 3D reconstructions.
   The blue vertical lines indicate the performance of the pose estimators when using the original CAD models (FoundationPose: \textbf{\color{RoyalBlue}full line}, Megapose: \textbf{\color{RoyalBlue}dashed line}).
   The objects for which it is the most challenging to replace the CAD model are the small and / or shiny objects, \eg, 06-tuna-fish-can, 07-pudding-box.
   For a given 3D reconstruction, the use of a high-performance pose estimator can compensate for some of the limitations of the 3D reconstruction: for example, FoundationPose~\cite{foundationposewen2024} induces a strong boost on 07-pudding-box compared to Megapose~\cite{labbe2022megapose}. 
   }
   \label{fig:ar_vs_obj_megapose_foundationpose_0}
\end{figure*}

\begin{figure*}[t]
   \centering
\includegraphics[width=0.75\linewidth,height=\textheight,keepaspectratio]
{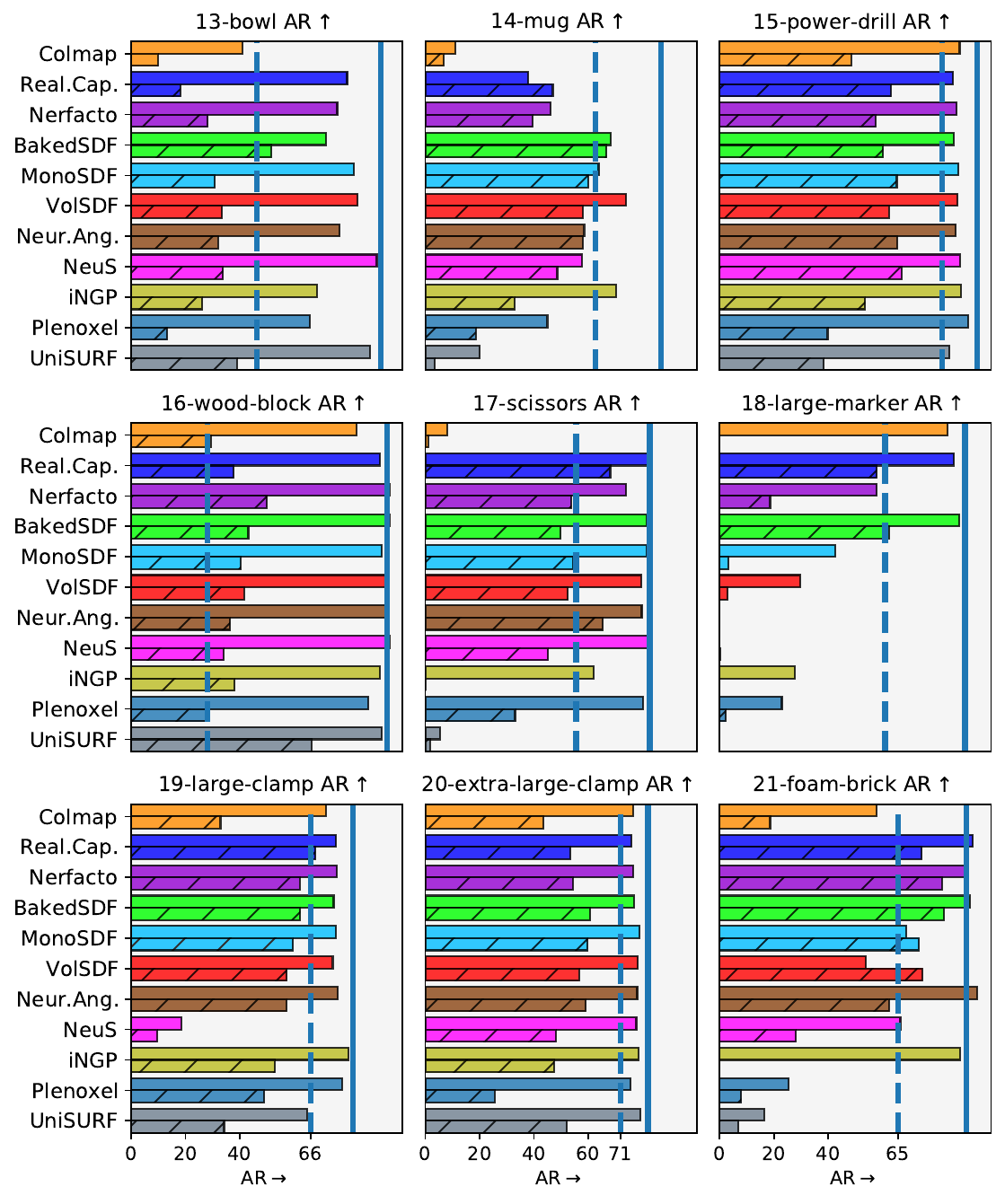}
   \caption{\textbf{Pose Evaluation of the 3D reconstructions for each YCB-V object~\cite{xiang2018posecnn} (1/2).}
   The 3D reconstructions are evaluated based on the performance of two pose estimators, \textbf{FoundationPose~\cite{foundationposewen2024} (uniform bar)} and \textbf{Megapose~\cite{labbe2022megapose} (lines bar)}, when the CAD model is replaced with 3D reconstructions.
   The blue vertical lines indicate the performance of the pose estimators when using the original CAD models (FoundationPose: \textbf{\color{RoyalBlue}full line}, Megapose: \textbf{\color{RoyalBlue}dashed line}).
   }
   \label{fig:ar_vs_obj_megapose_foundationpose_1}
\end{figure*}

We report the pose performances separately for each YCB-V object~\cite{calli2015ycb,xiang2018posecnn}, 3D reconstruction, and pose estimators in Fig.~\ref{fig:ar_vs_obj_megapose_foundationpose_0} and Fig.~\ref{fig:ar_vs_obj_megapose_foundationpose_1}.
The results are consistent with the previous conclusions that the most challenging objects are the small ones or shiny ones, \eg, 06-tuna-fish-can or the 18-large-marker.

As observed previously, the combination of FoundationPose~\cite{foundationposewen2024} and the 3D reconstructions is often better than Megapose~\cite{labbe2022megapose} with the original CAD models, \eg, 02-cracker-box, 10-banana.
This suggests that bridging the gap between 3D reconstructions and CAD models for pose estimations has two directions for improvements: improving the 3D reconstruction itself and improving pose estimators.

\subsection{Qualitative Results}
\label{supp:quali}

\begin{figure*}
   \centering   \includegraphics[width=\linewidth,height=\textheight,keepaspectratio]{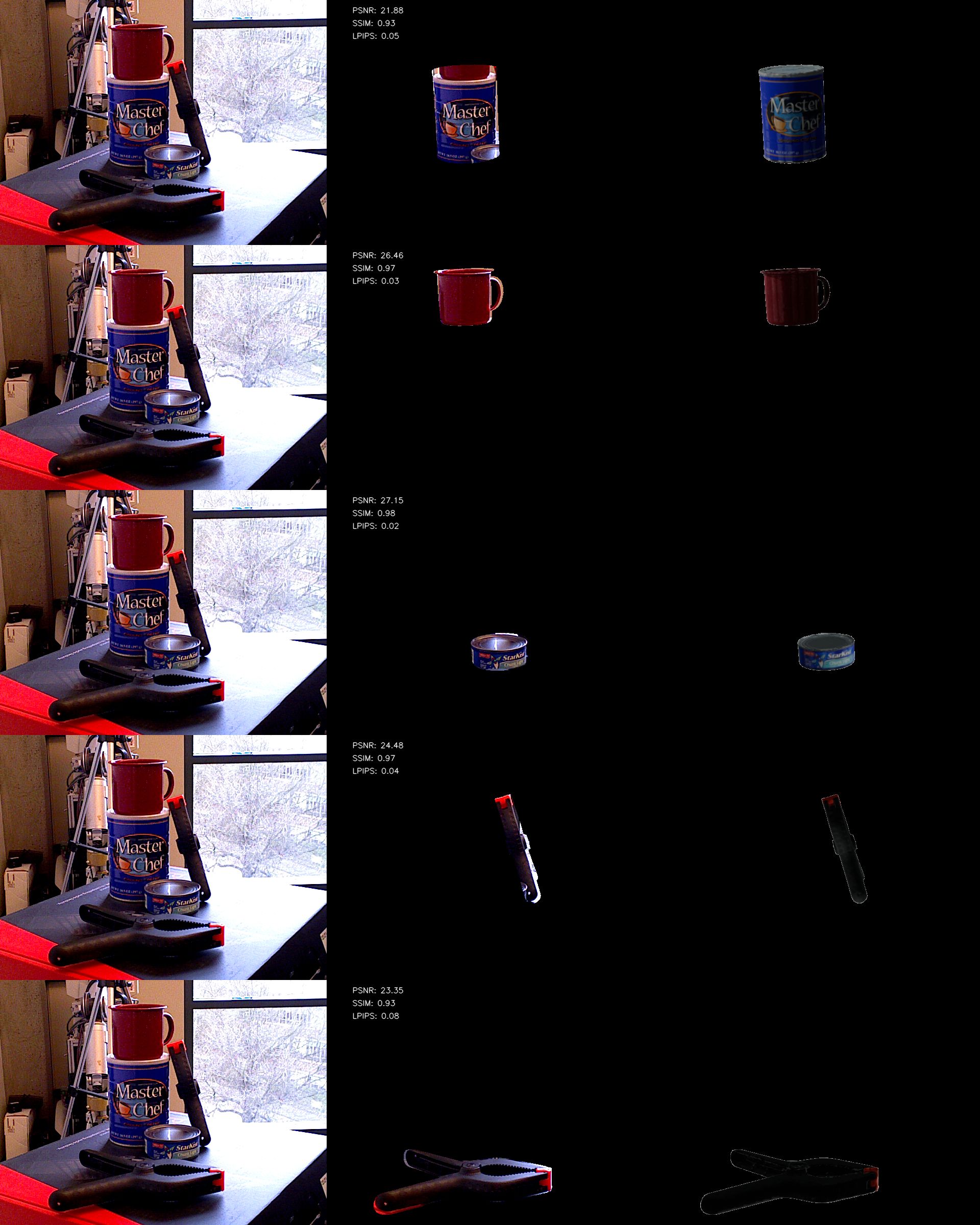}
   \caption{Methodology for evaluating SSIM, PSNR, and LPIPS metrics. On the left is the test YCB-V image, in the middle is the masked YCB-V object, and on the right is the rendering of the reconstructed mesh overlay on the masked image.}
   \label{fig:SSIM_PSNR}
\end{figure*}

The SSIM, PSNR, and LPIPS~\cite{zhang2018perceptual} metrics are evaluated by comparing the rendering of the CAD model and the masked YCB-V images, as illustrated in Figure \ref{fig:SSIM_PSNR}. We report the average values across all test images.

\begin{figure*}
   \centering   \includegraphics[width=\linewidth,height=\textheight,keepaspectratio]{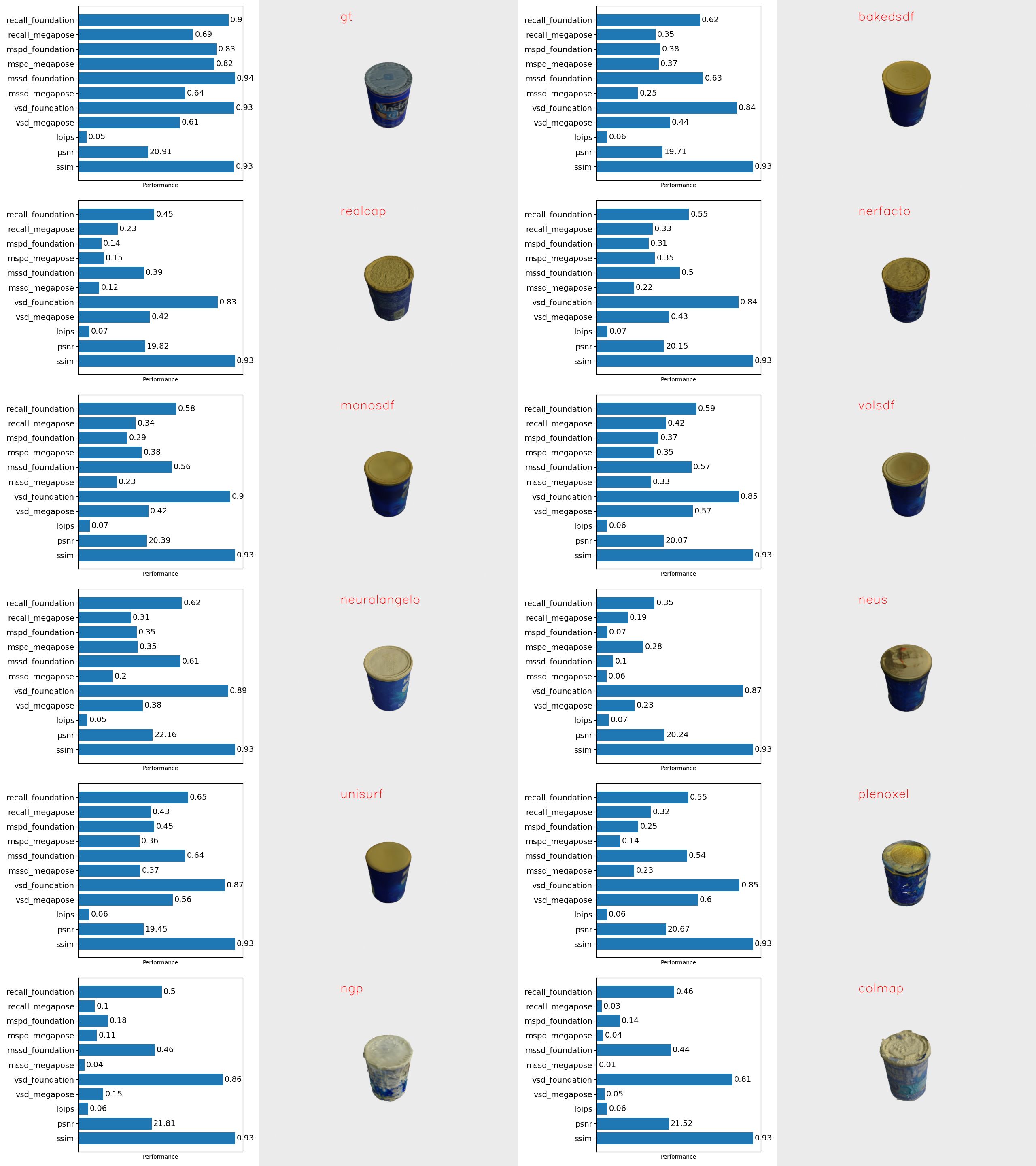}
   \caption{Pose performance and texture scores (left) and object renderings (right) of 01-master-chef-can reconstructed with various reconstructed methods.}
   \label{fig:01_master_chef_can}
\end{figure*}

\begin{figure*}
   \centering   \includegraphics[width=\linewidth,height=\textheight,keepaspectratio]{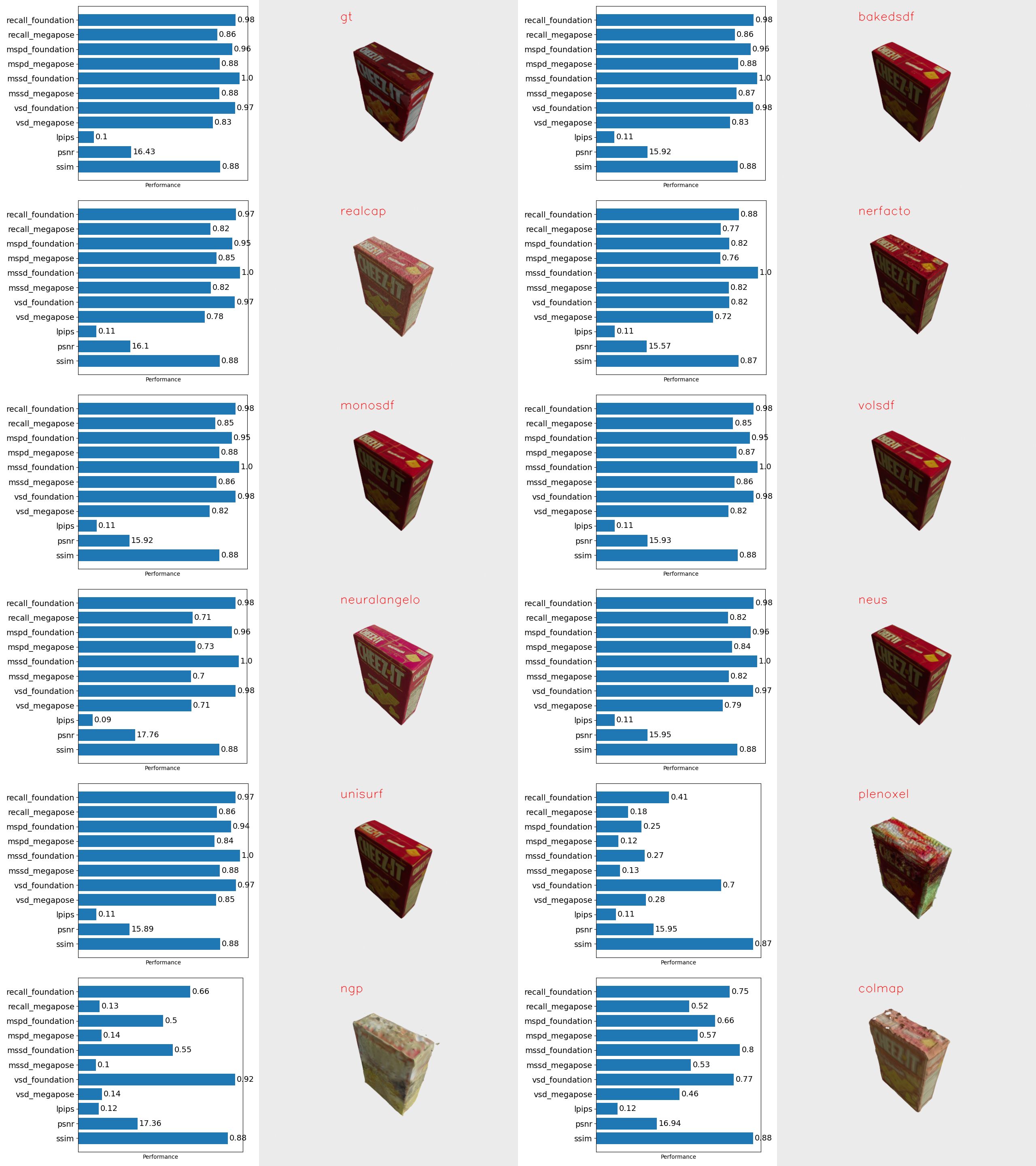}
   \caption{Pose performance and texture scores (left) and object renderings (right) of 02-cracker-box reconstructed with various reconstructed methods.}
   \label{fig:02_cracker_box}
\end{figure*}

\begin{figure*}
   \centering   \includegraphics[width=\linewidth,height=\textheight,keepaspectratio]{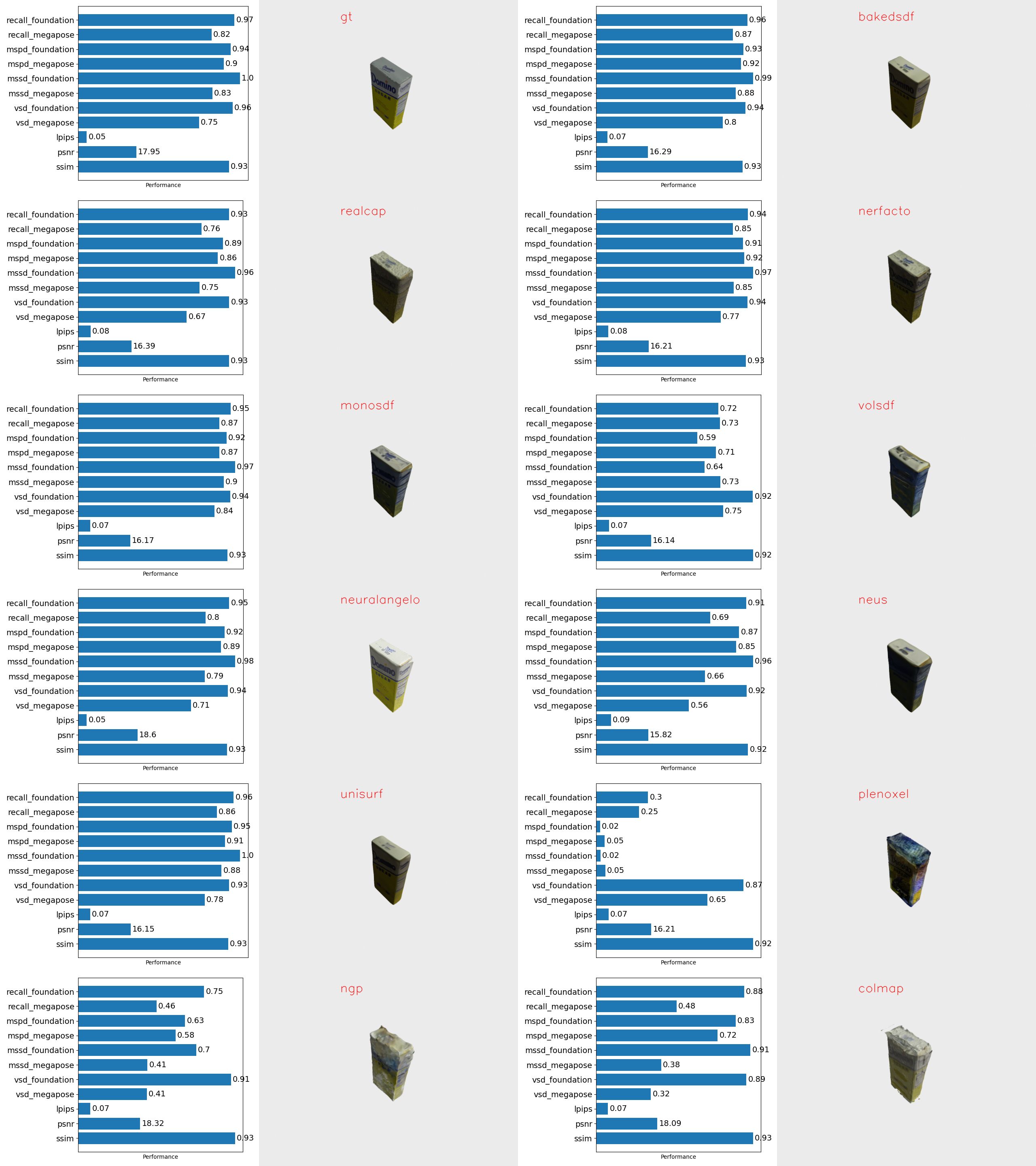}
   \caption{Pose performance and texture scores (left) and object renderings (right) of 03-sugar-box reconstructed with various reconstructed methods.}
   \label{fig:03_sugar_box}
\end{figure*}

\begin{figure*}
   \centering   \includegraphics[width=\linewidth,height=\textheight,keepaspectratio]{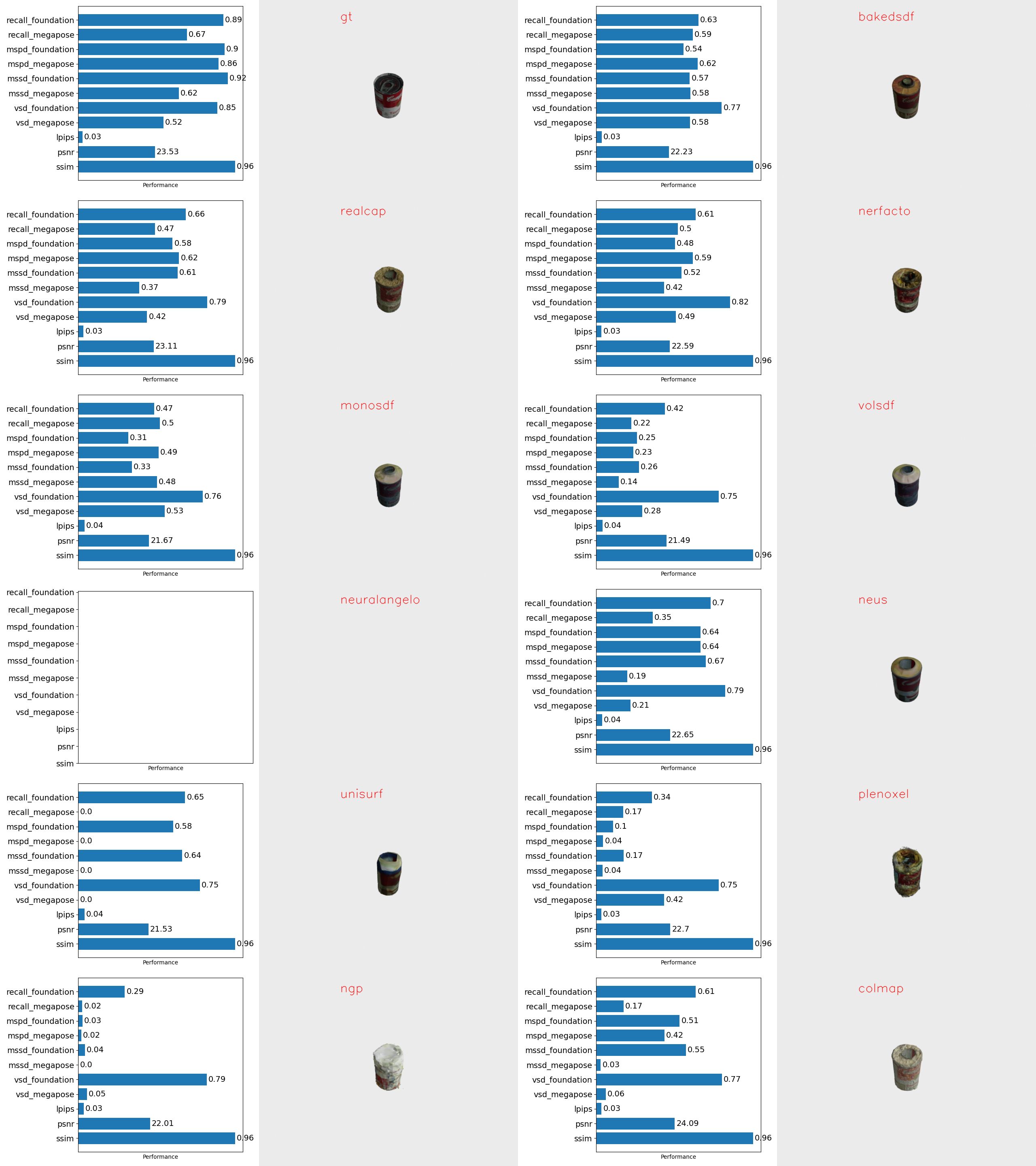}
   \caption{Pose performance and texture scores (left) and object renderings (right) of 04-tomatoe-soup-can reconstructed with various reconstructed methods.}
   \label{fig:04_tomatoe_soup_can}
\end{figure*}

\begin{figure*}
   \centering   \includegraphics[width=\linewidth,height=\textheight,keepaspectratio]{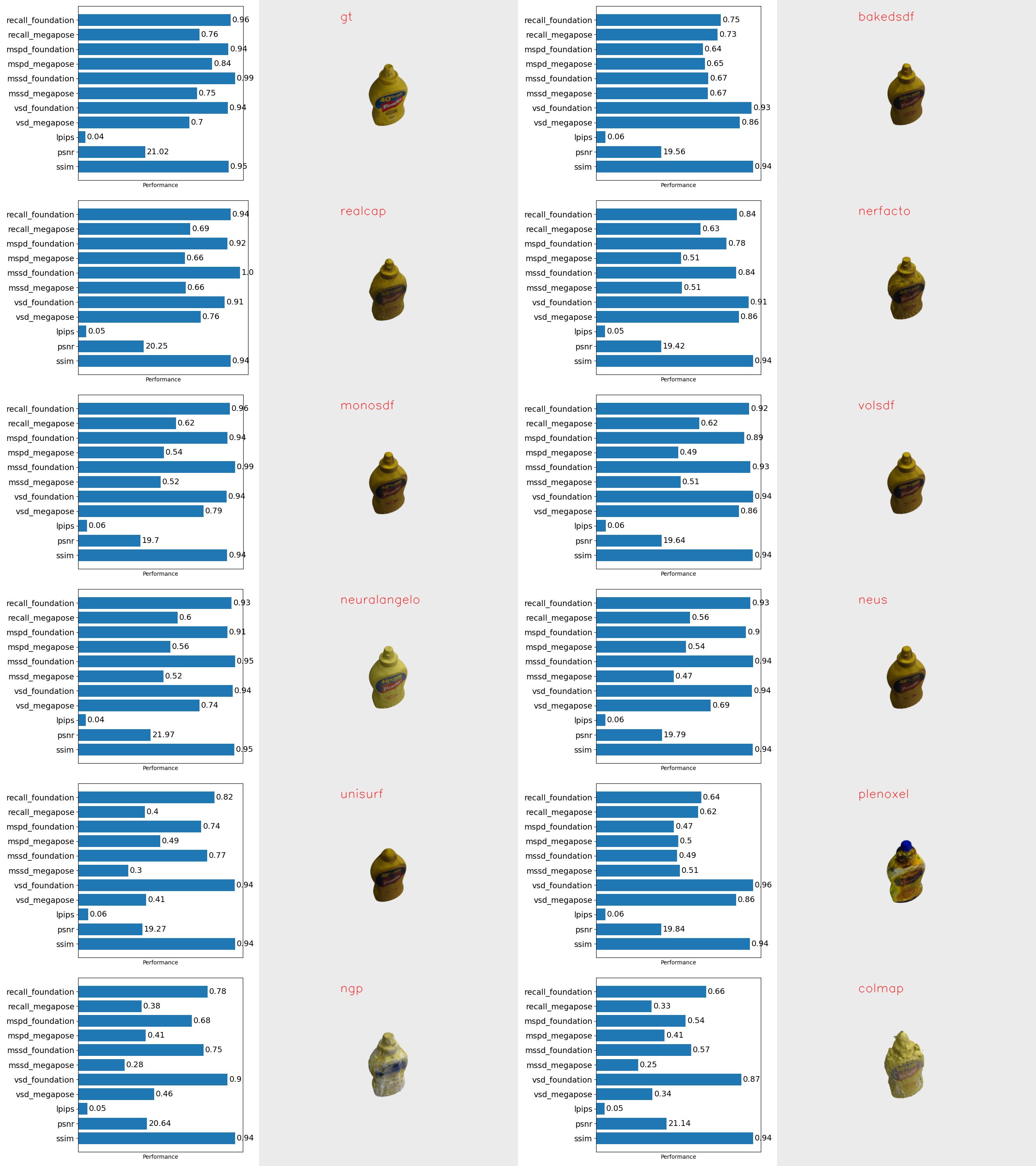}
   \caption{Pose performance and texture scores (left) and object renderings (right) of 05-mustard-bottle reconstructed with various reconstructed methods.}
   \label{fig:05_mustard_bottle}
\end{figure*}

\begin{figure*}
   \centering   \includegraphics[width=\linewidth,height=\textheight,keepaspectratio]{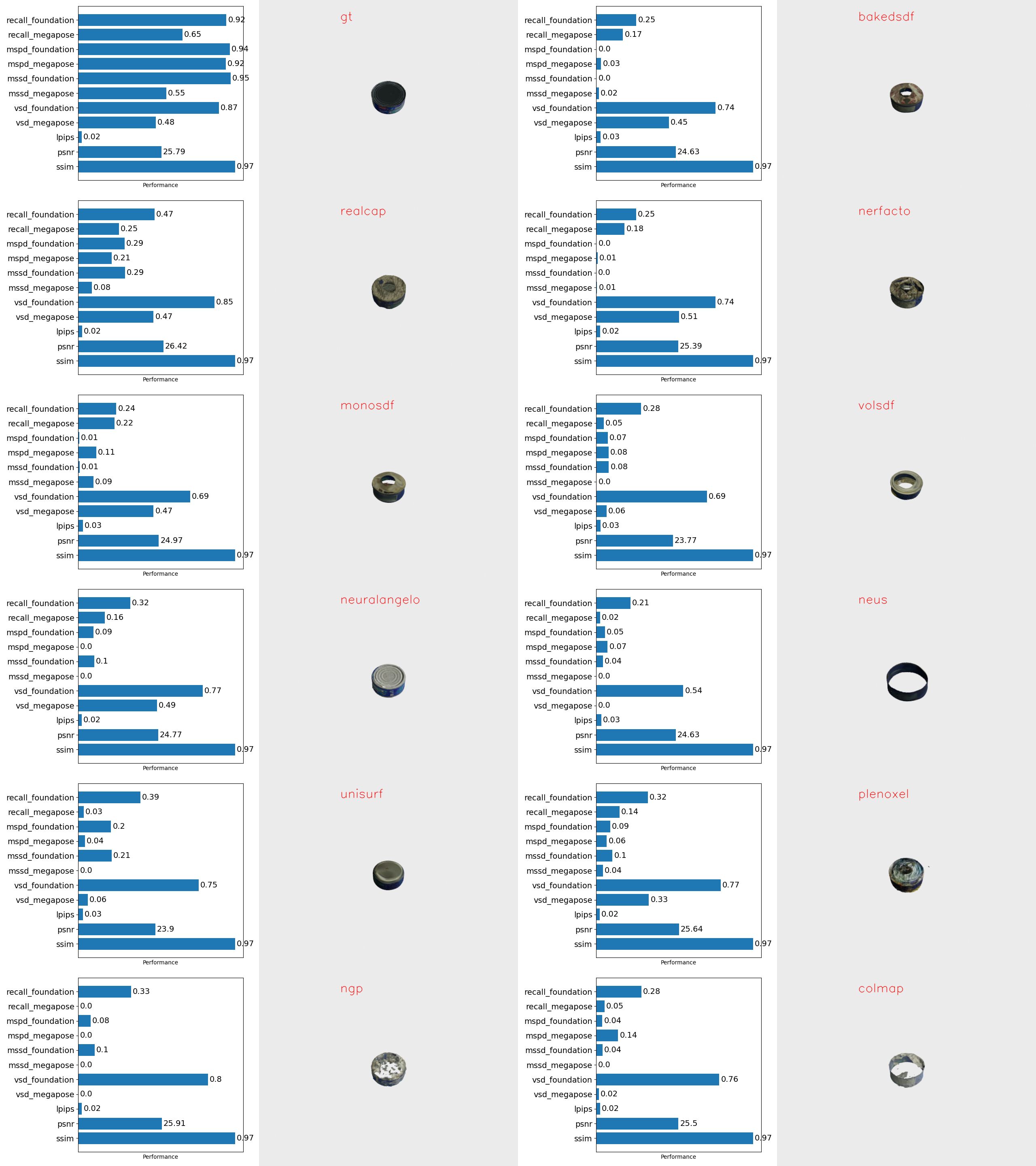}
   \caption{Pose performance and texture scores (left) and object renderings (right) of 06-tuna-fish-can reconstructed with various reconstructed methods.}
   \label{fig:06_tuna_fish_can}
\end{figure*}

\begin{figure*}
   \centering   \includegraphics[width=\linewidth,height=\textheight,keepaspectratio]{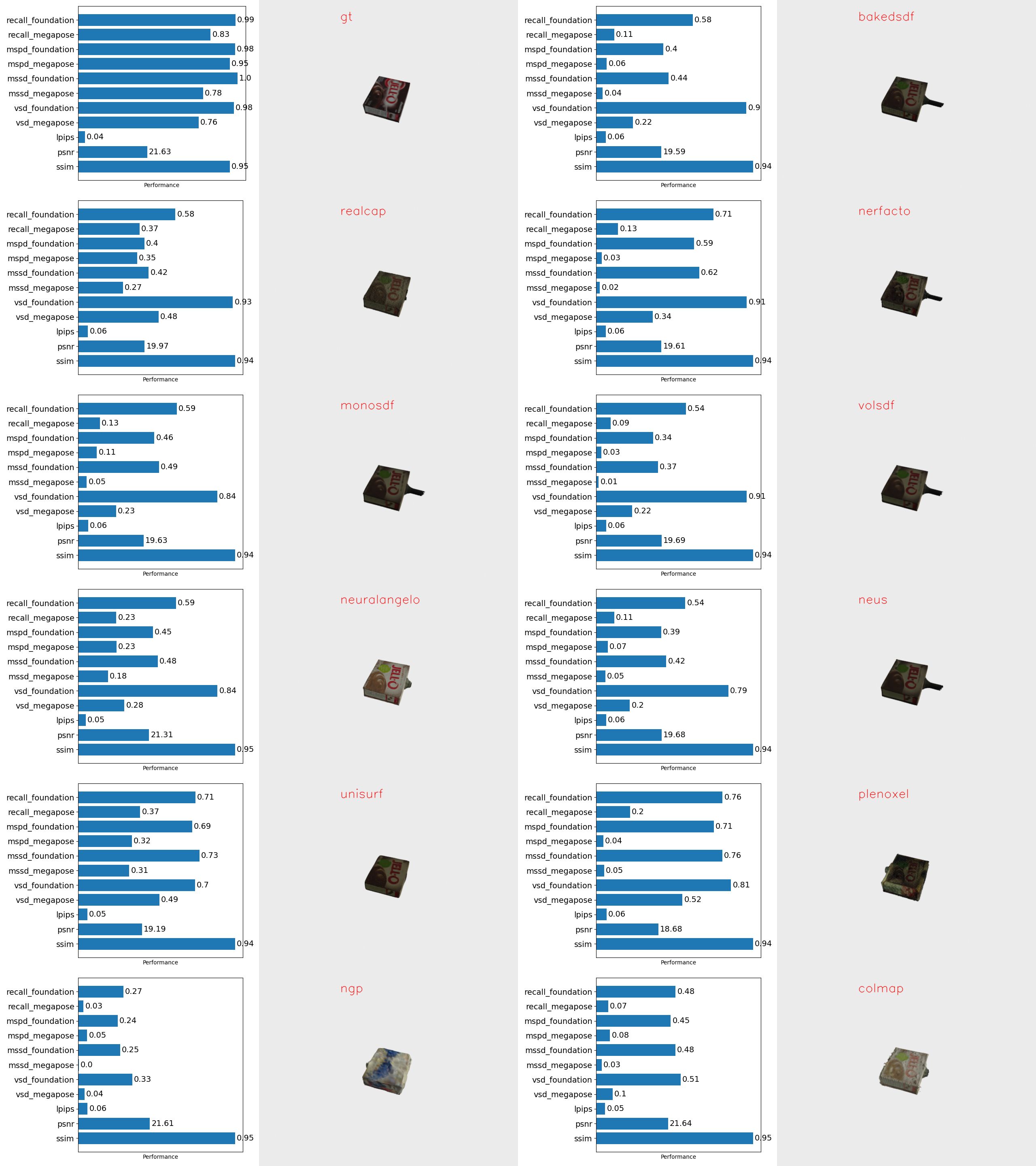}
   \caption{Pose performance and texture scores (left) and object renderings (right) of 07-pudding-box reconstructed with various reconstructed methods.}
   \label{fig:07_pudding_box}
\end{figure*}

\begin{figure*}
   \centering   \includegraphics[width=\linewidth,height=\textheight,keepaspectratio]{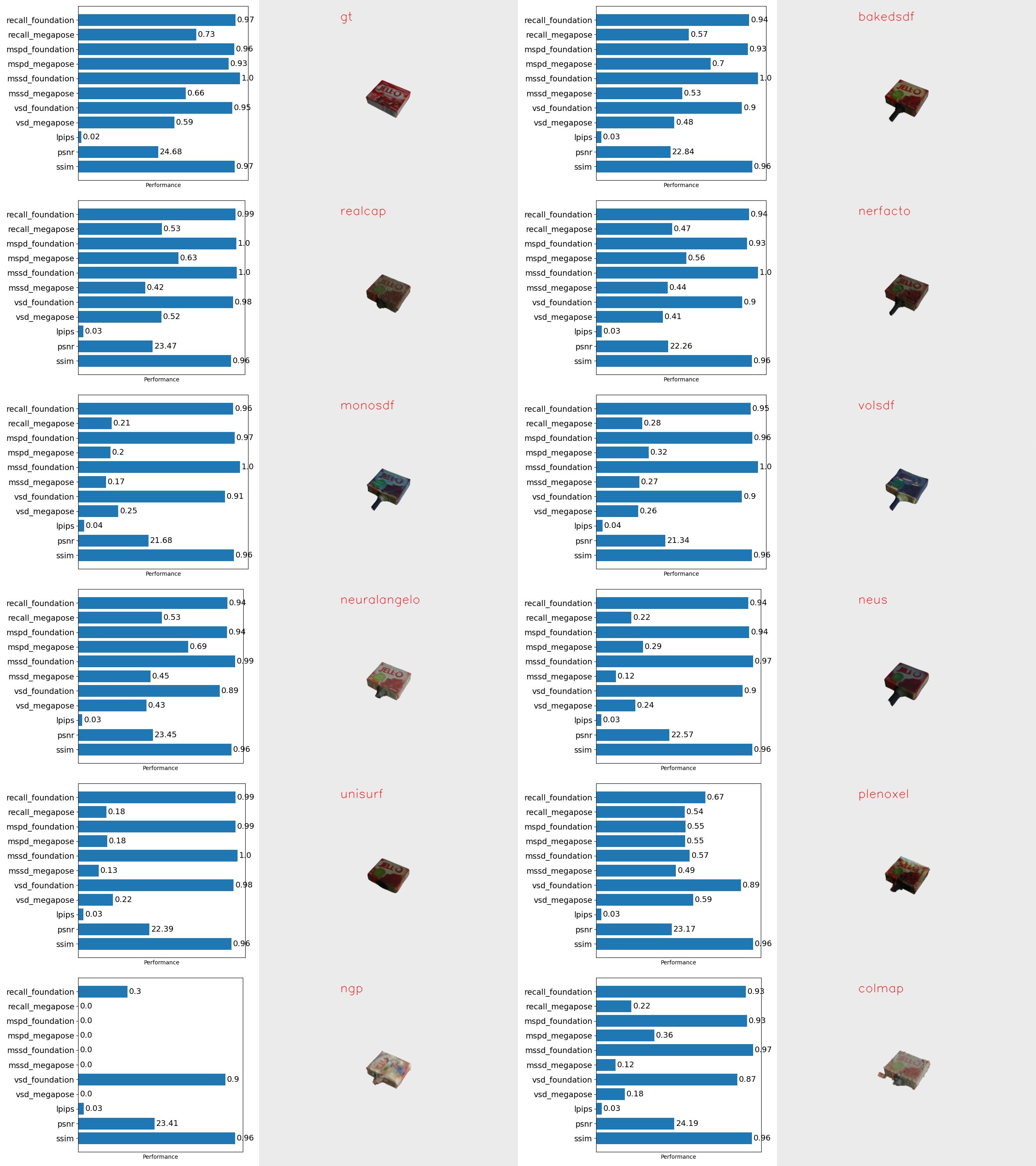}
   \caption{Pose performance and texture scores (left) and object renderings (right) of 08-gelatin-box reconstructed with various reconstructed methods.}
   \label{fig:08_gelatin_box}
\end{figure*}

\begin{figure*}
   \centering   \includegraphics[width=\linewidth,height=\textheight,keepaspectratio]{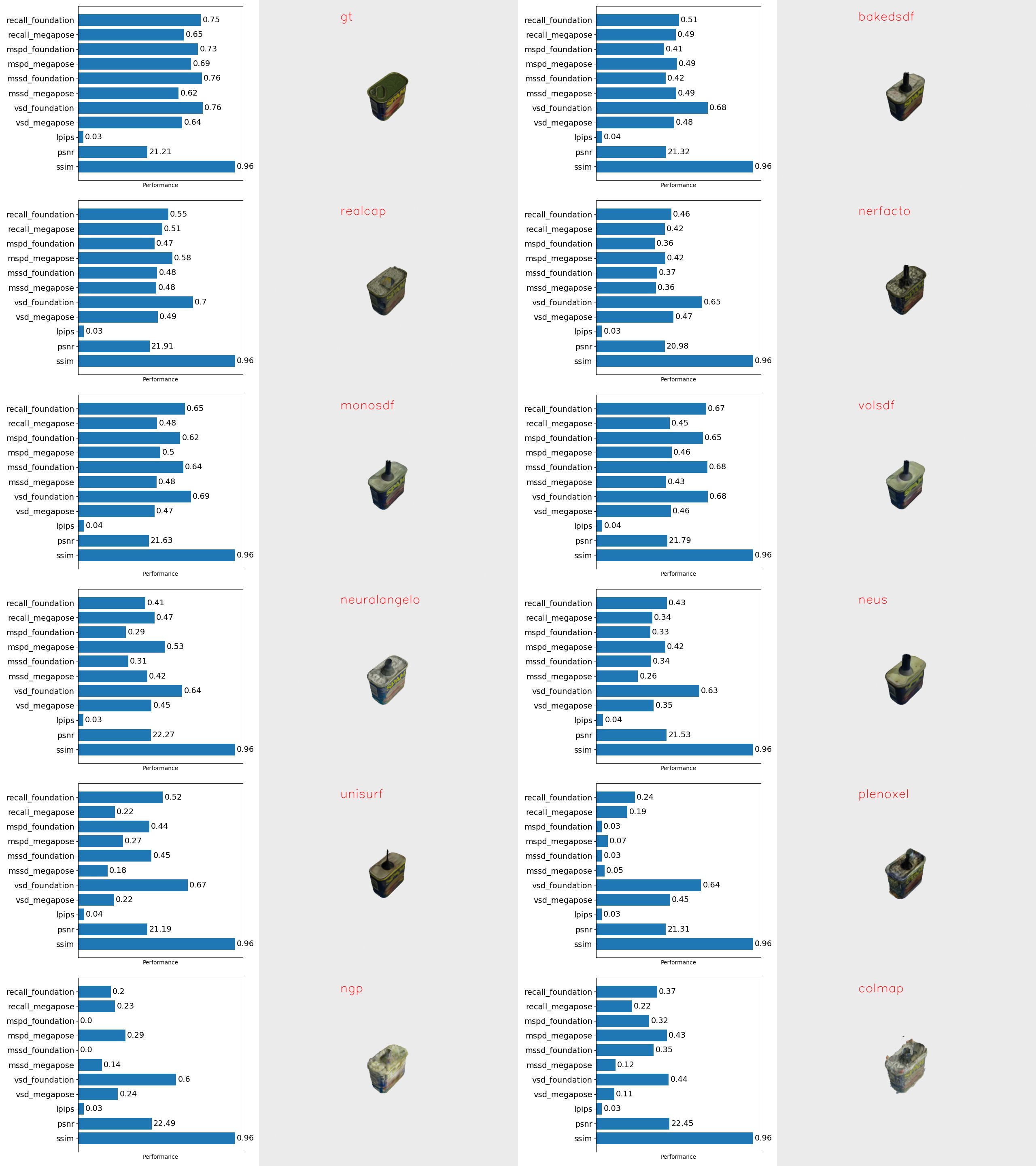}
   \caption{Pose performance and texture scores (left) and object renderings (right) of 09-potted-meat-can reconstructed with various reconstructed methods.}
   \label{fig:09_potted_meat_can}
\end{figure*}

\begin{figure*}
   \centering   \includegraphics[width=\linewidth,height=\textheight,keepaspectratio]{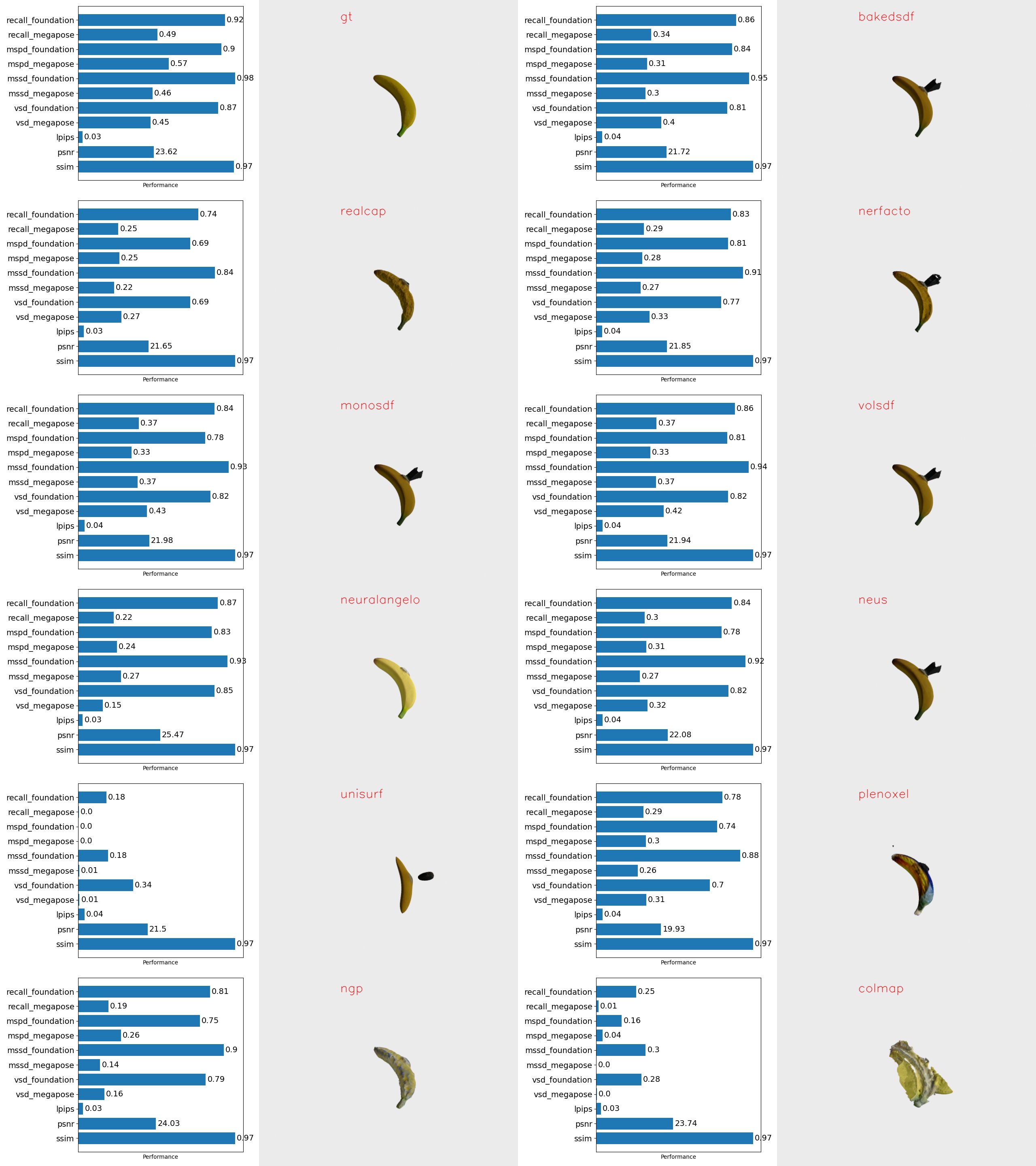}
   \caption{Pose performance and texture scores (left) and object renderings (right) of 10-banana reconstructed with various reconstructed methods.}
   \label{fig:10_banana}
\end{figure*}

\begin{figure*}
   \centering   \includegraphics[width=\linewidth,height=\textheight,keepaspectratio]{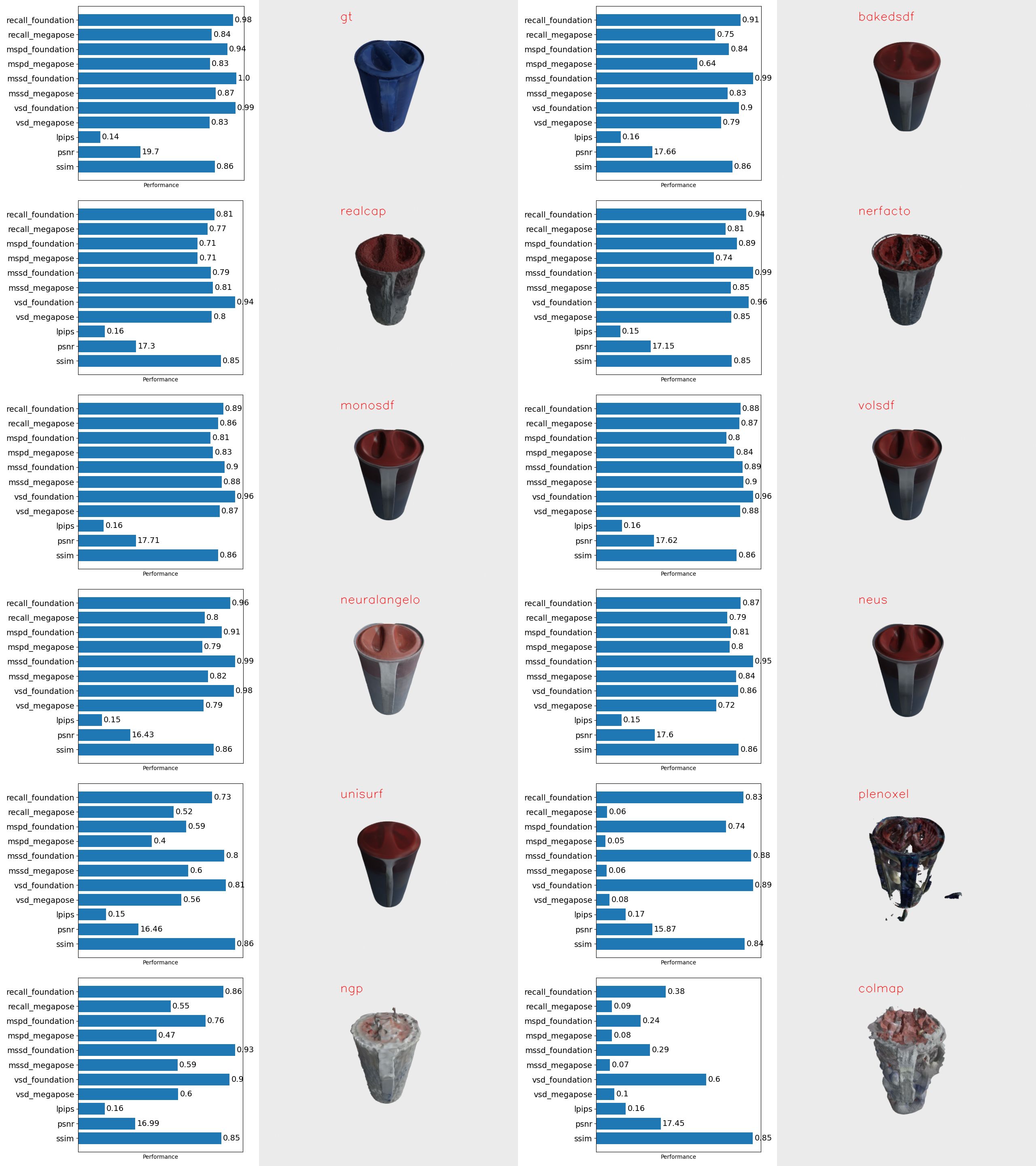}
   \caption{Pose performance and texture scores (left) and object renderings (right) of 11-pitcher-base reconstructed with various reconstructed methods.}
   \label{fig:11_pitcher_base}
\end{figure*}

\begin{figure*}
   \centering   \includegraphics[width=\linewidth,height=\textheight,keepaspectratio]{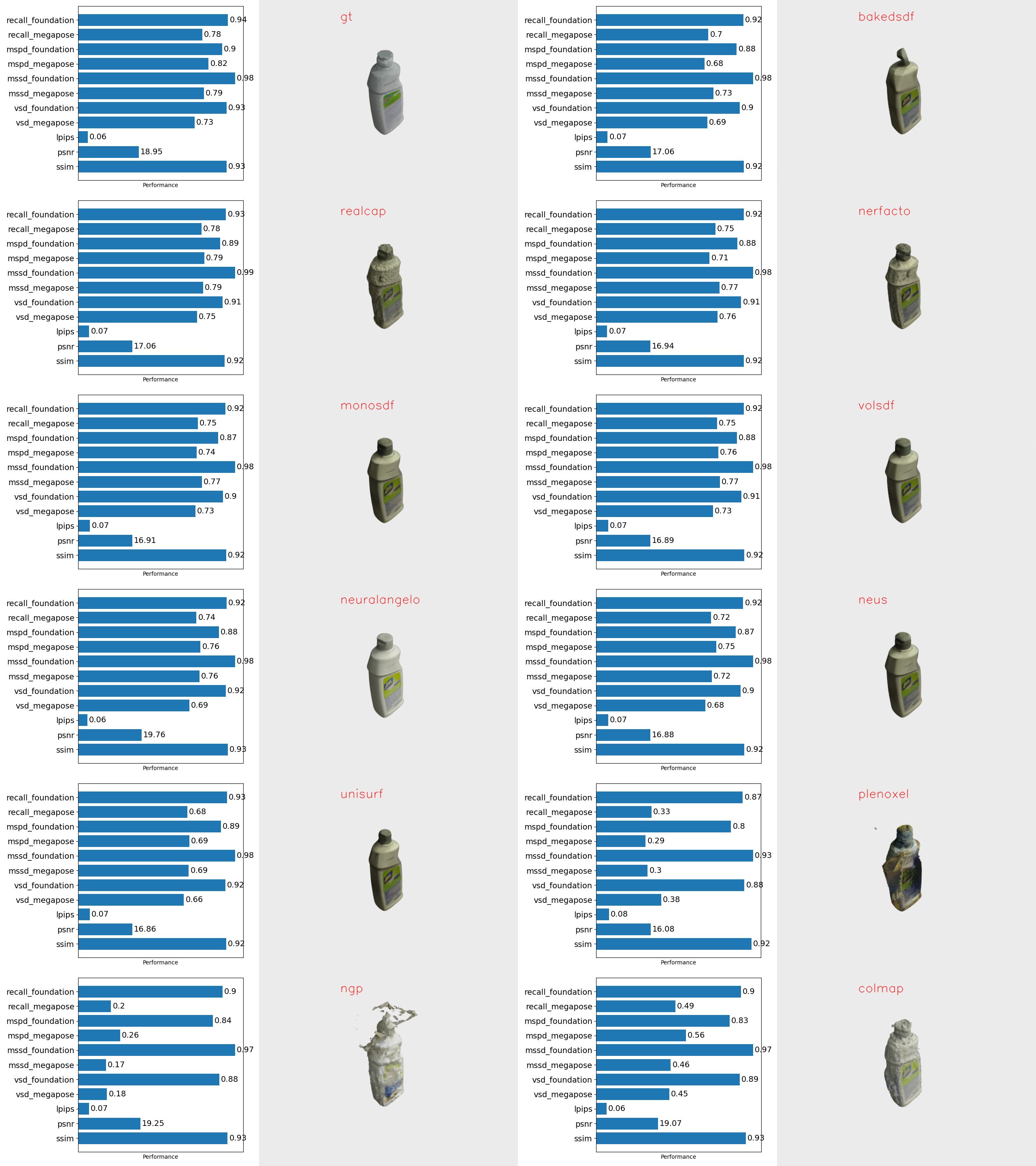}
   \caption{Pose performance and texture scores (left) and object renderings (right) of 12-bleach-cleanser reconstructed with various reconstructed methods.}
   \label{fig:12_bleach_cleanser}
\end{figure*}

\begin{figure*}
   \centering   \includegraphics[width=\linewidth,height=\textheight,keepaspectratio]{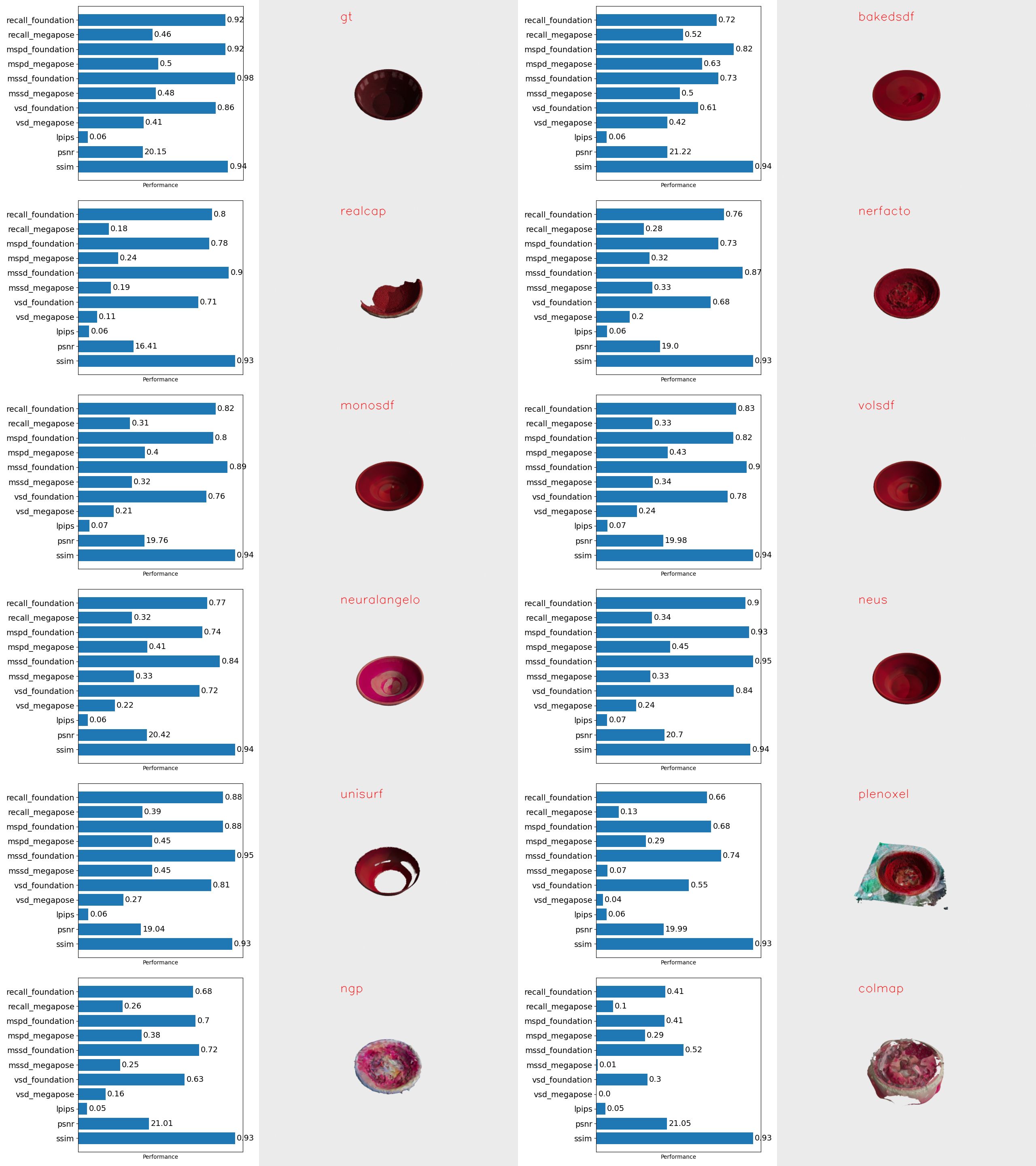}
   \caption{Pose performance and texture scores (left) and object renderings (right) of 13-bowl reconstructed with various reconstructed methods.}
   \label{fig:13_bowl}
\end{figure*}

\begin{figure*}
   \centering   \includegraphics[width=\linewidth,height=\textheight,keepaspectratio]{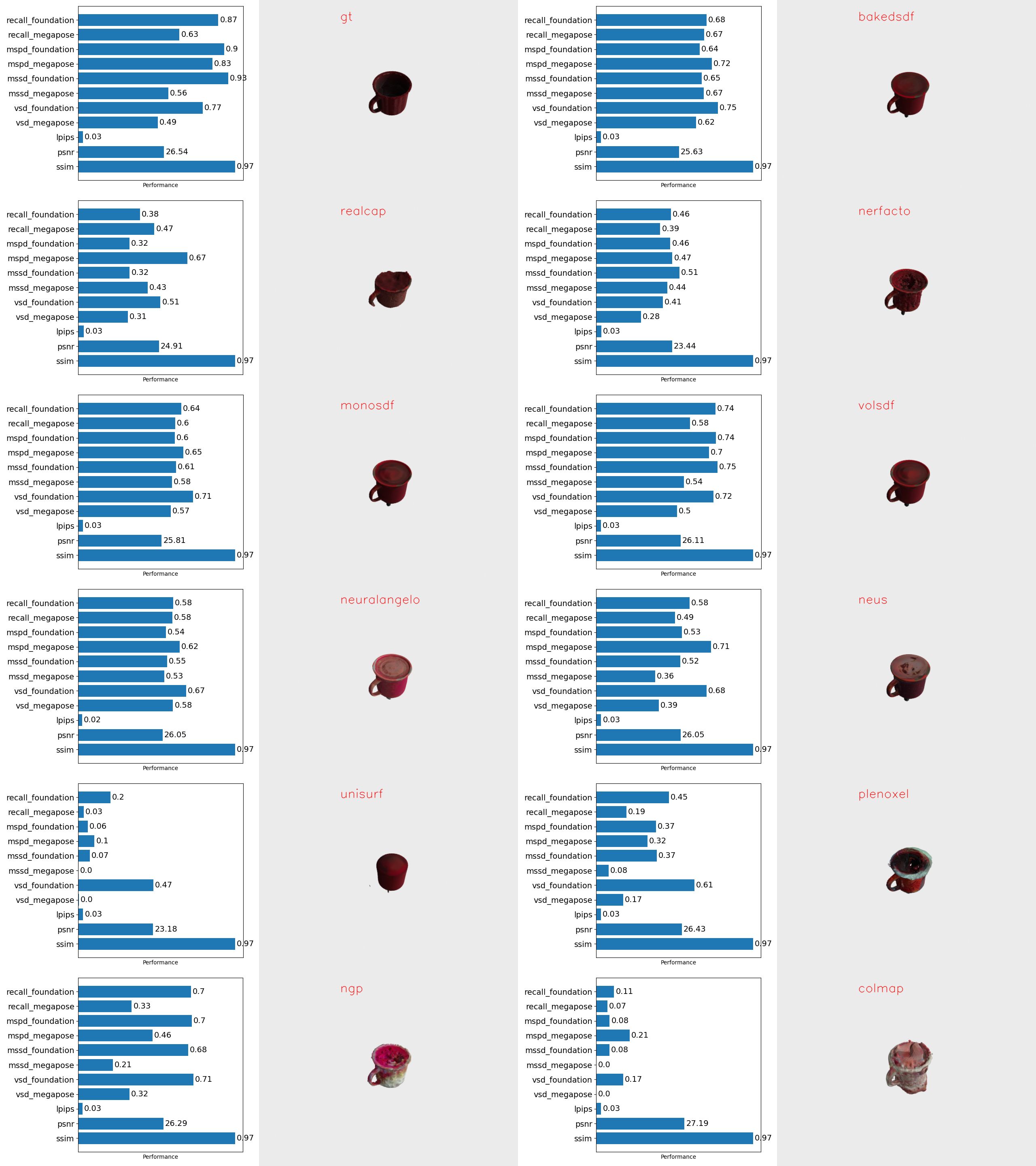}
   \caption{Pose performance and texture scores (left) and object renderings (right) of 14-mug reconstructed with various reconstructed methods.}
   \label{fig:14_mug}
\end{figure*}

\begin{figure*}
   \centering   \includegraphics[width=\linewidth,height=\textheight,keepaspectratio]{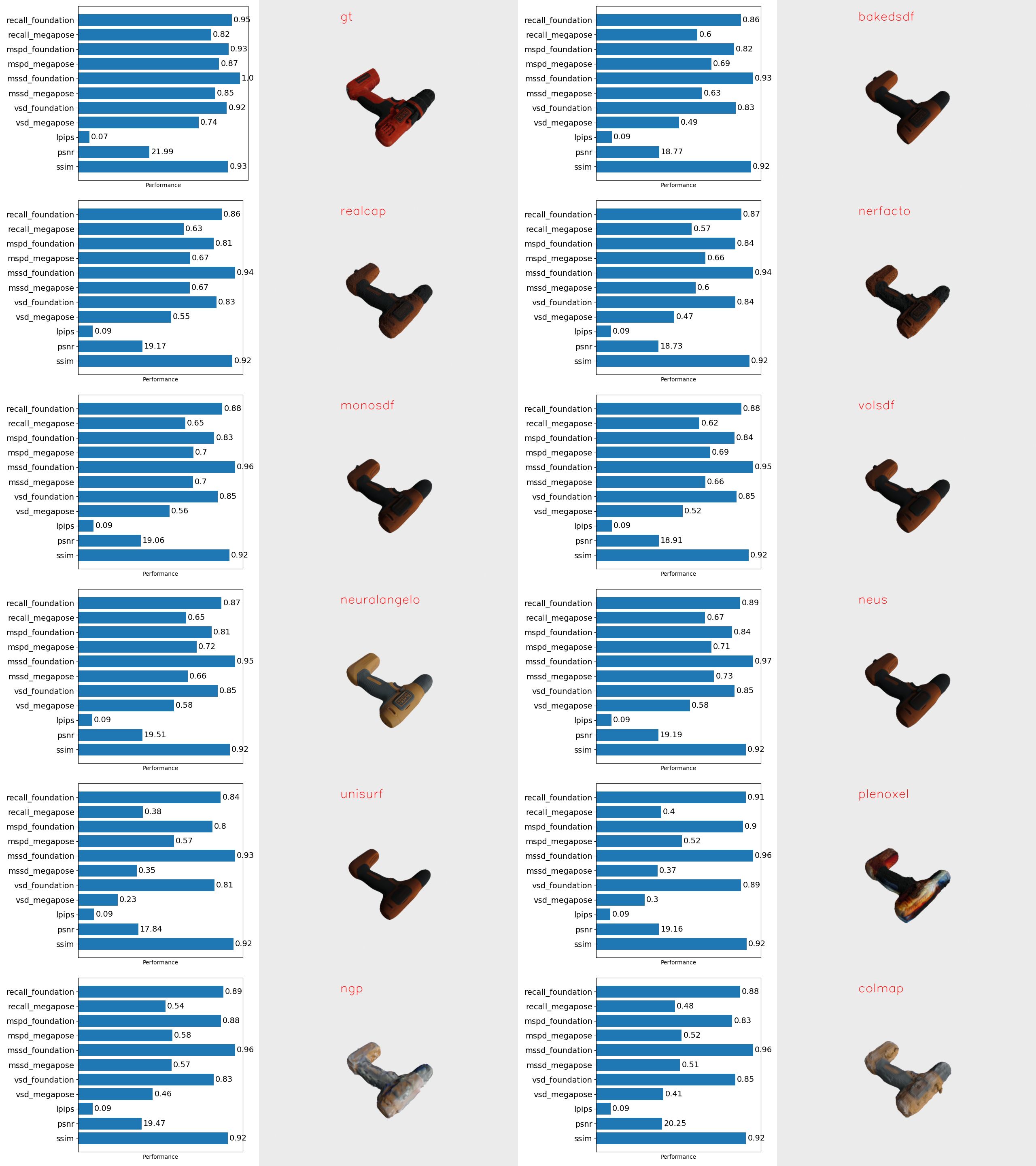}
   \caption{Pose performance and texture scores (left) and object renderings (right) of 15-power-drill reconstructed with various reconstructed methods.}
   \label{fig:15_power_drill}
\end{figure*}

\begin{figure*}
   \centering   \includegraphics[width=\linewidth,height=\textheight,keepaspectratio]{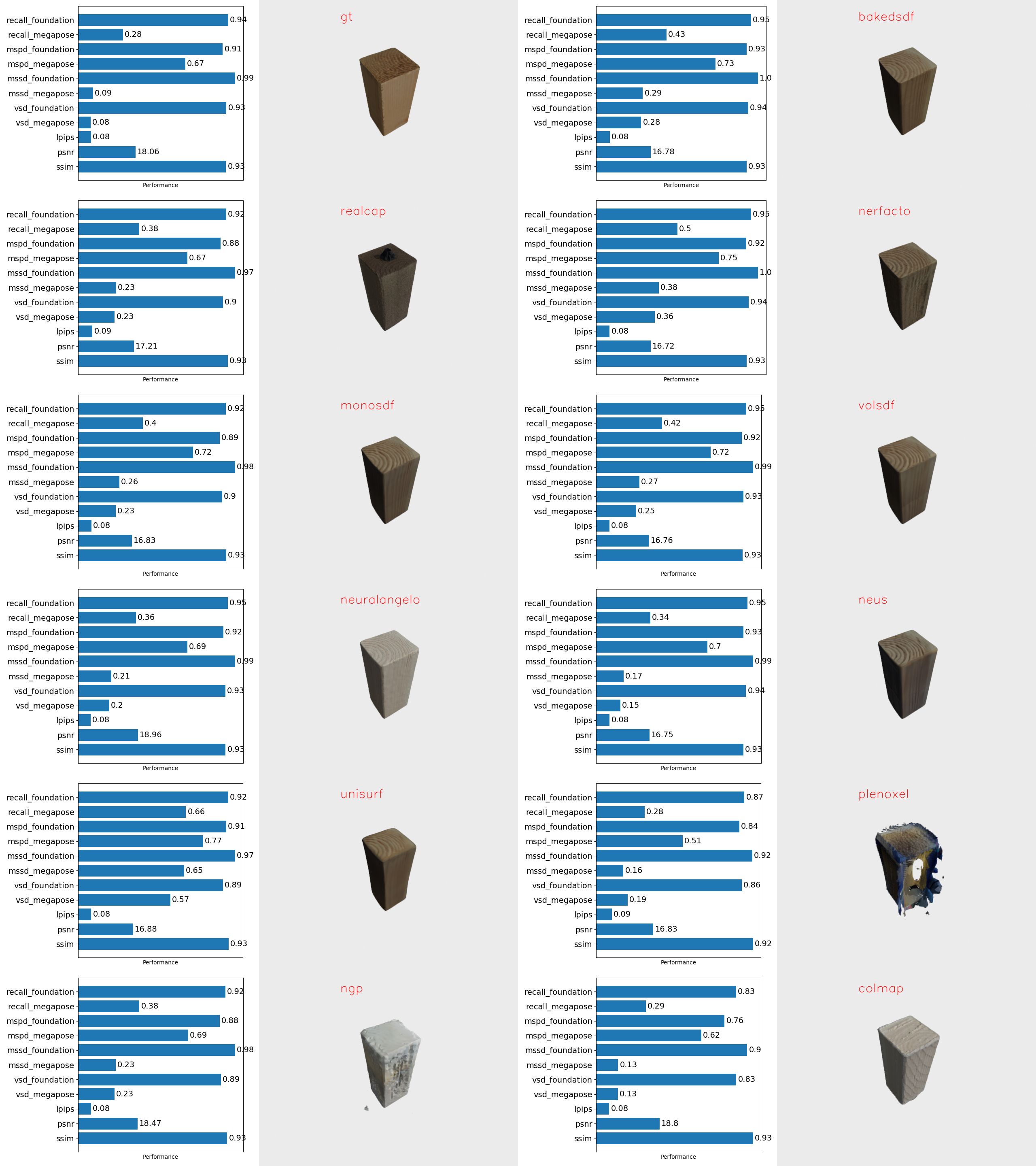}
   \caption{Pose performance and texture scores (left) and object renderings (right) of 16-wood-block reconstructed with various reconstructed methods.}
   \label{fig:16_wood_block}
\end{figure*}

\begin{figure*}
   \centering   \includegraphics[width=\linewidth,height=\textheight,keepaspectratio]{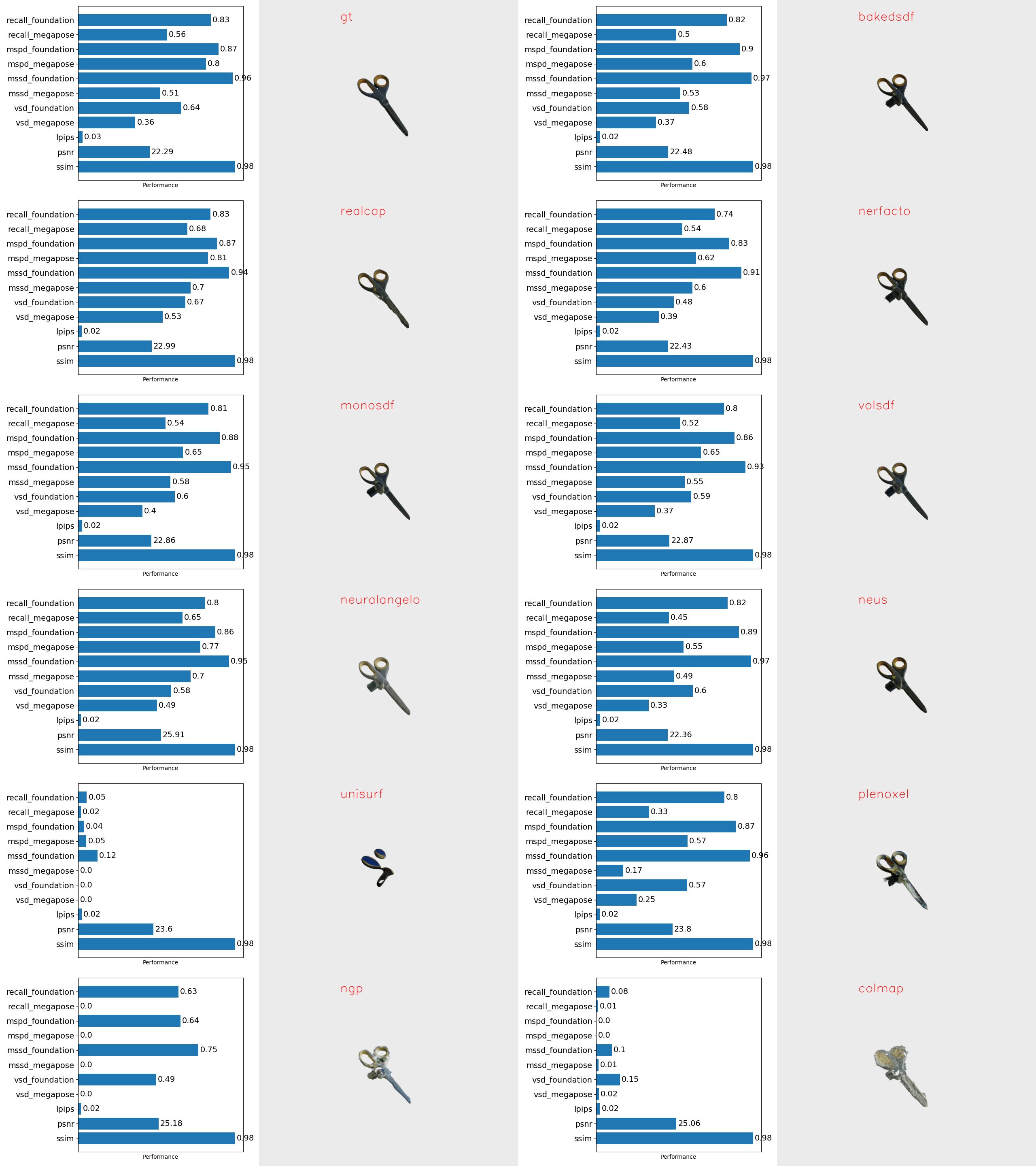}
   \caption{Pose performance and texture scores (left) and object renderings (right) of 17-scissors reconstructed with various reconstructed methods.}
   \label{fig:17_scissors}
\end{figure*}

\begin{figure*}
   \centering   \includegraphics[width=\linewidth,height=\textheight,keepaspectratio]{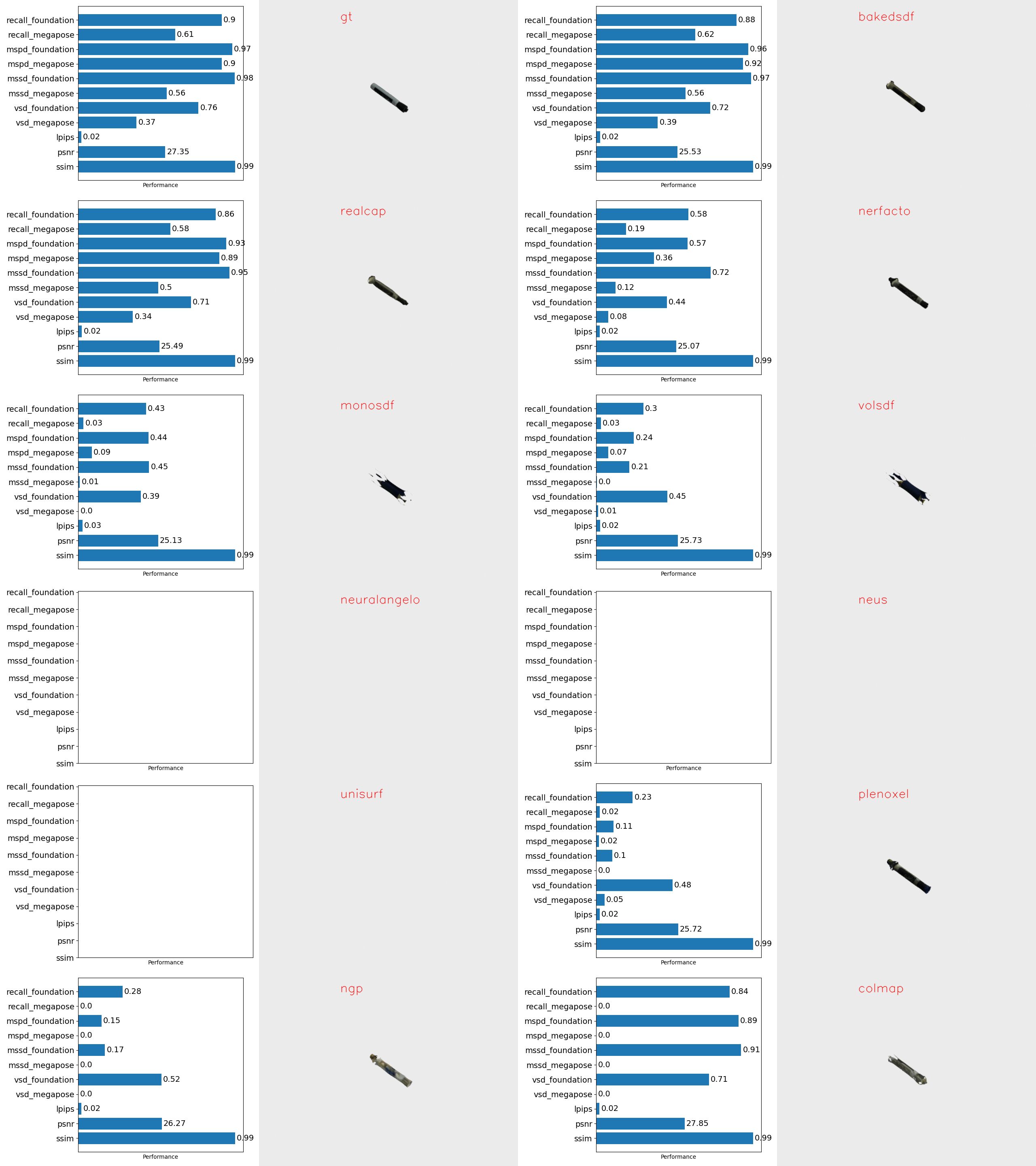}
   \caption{Pose performance and texture scores (left) and object renderings (right) of 18-large-marker reconstructed with various reconstructed methods.}
   \label{fig:18_large_marker}
\end{figure*}

\begin{figure*}
   \centering   \includegraphics[width=\linewidth,height=\textheight,keepaspectratio]{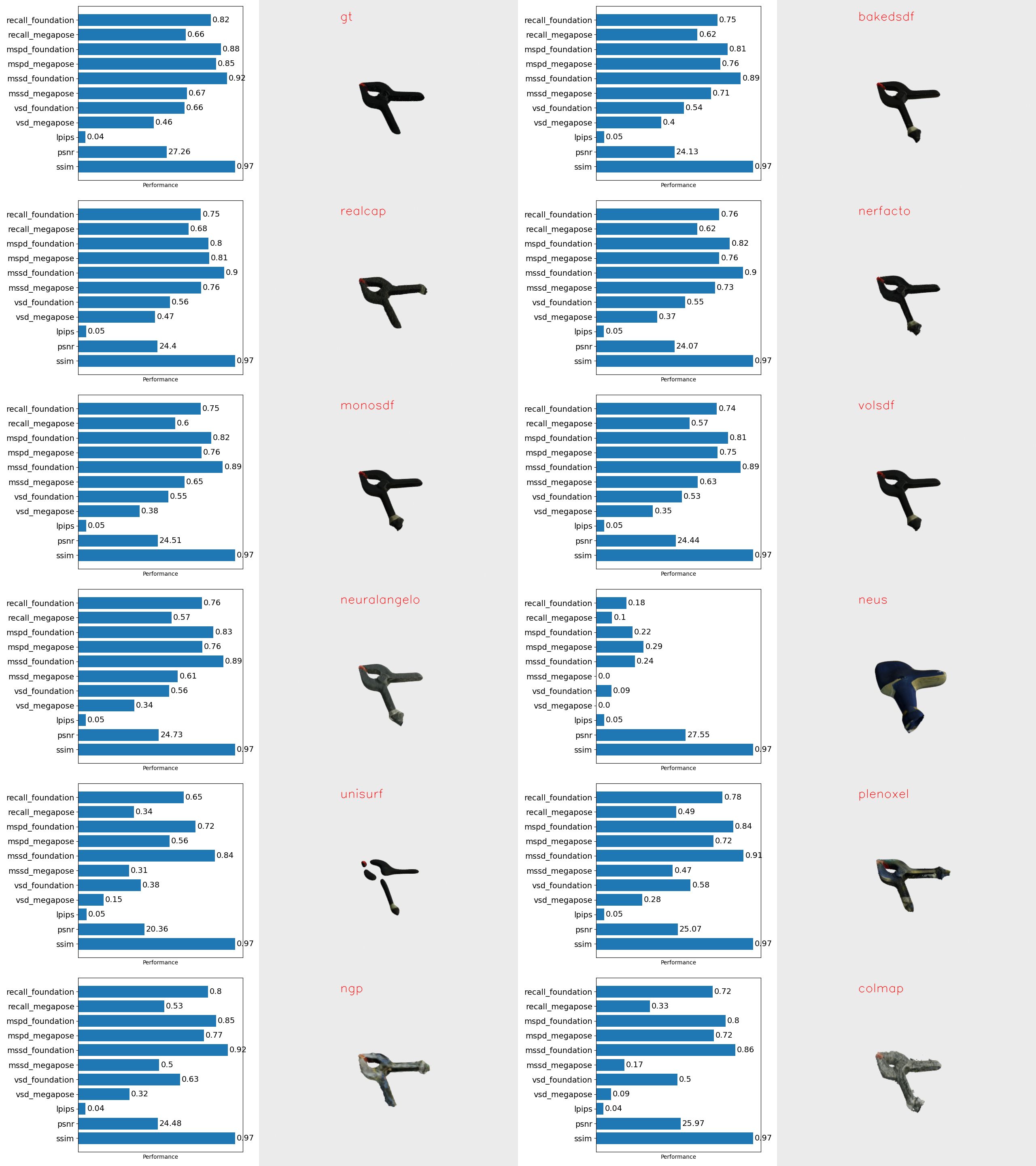}
   \caption{Pose performance and texture scores (left) and object renderings (right) of 19-large-clamp reconstructed with various reconstructed methods.}
   \label{fig:19_large_clamp}
\end{figure*}

\begin{figure*}
   \centering   \includegraphics[width=\linewidth,height=\textheight,keepaspectratio]{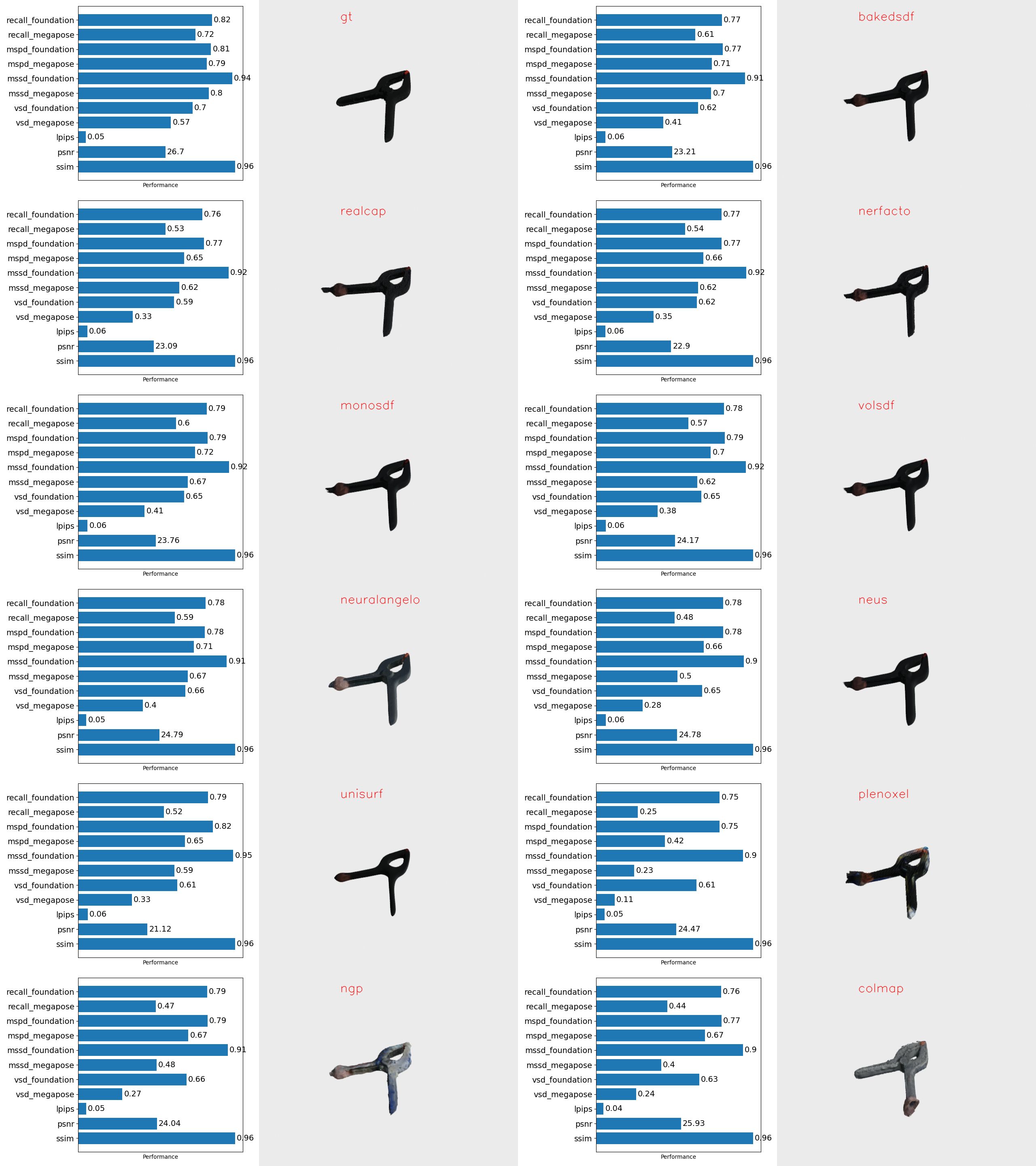}
   \caption{Pose performance and texture scores (left) and object renderings (right) of 20-extra-large-clamp reconstructed with various reconstructed methods.}
   \label{fig:20_extra_large_clamp}
\end{figure*}

\begin{figure*}
   \centering   \includegraphics[width=\linewidth,height=\textheight,keepaspectratio]{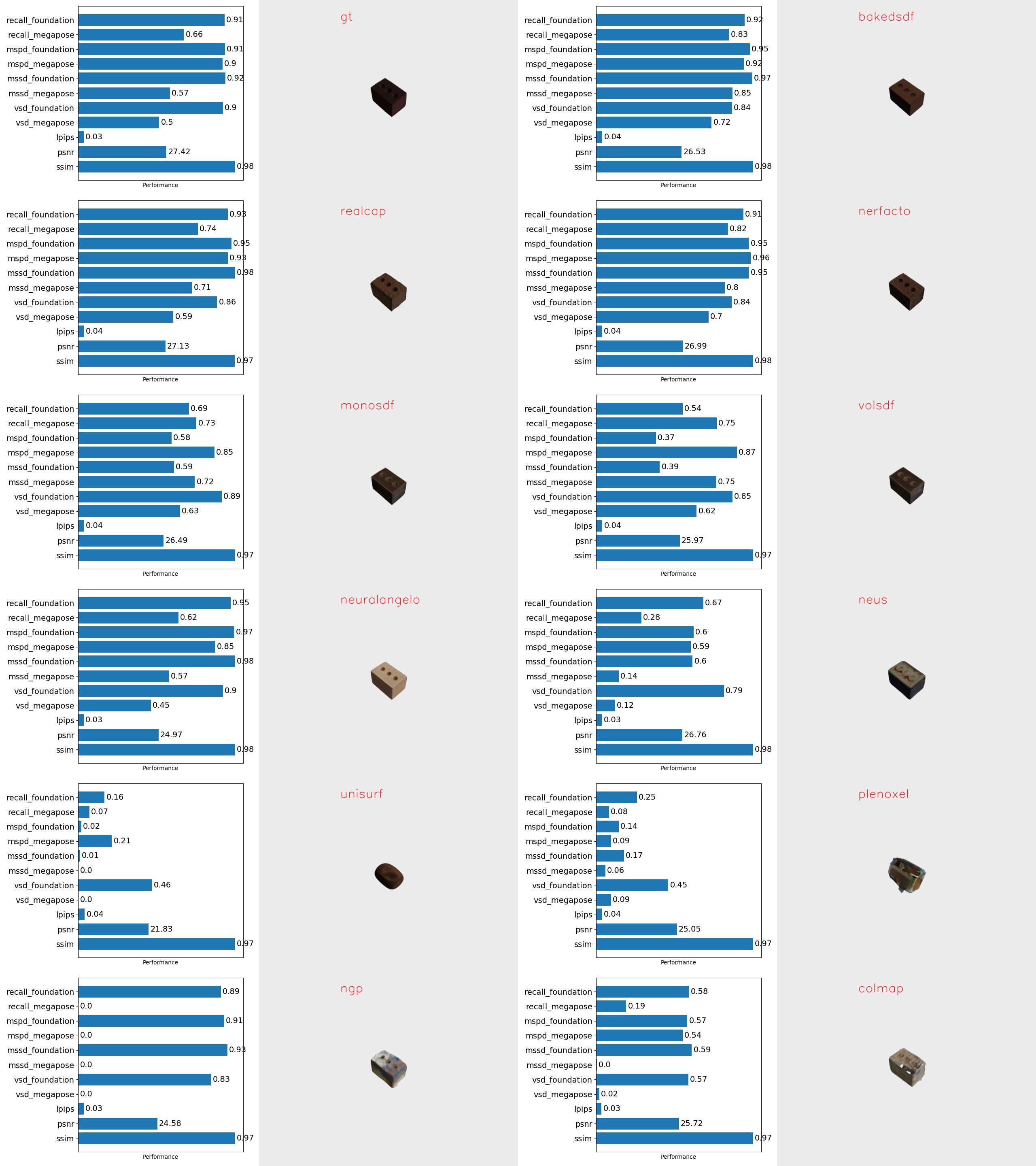}
   \caption{Pose performance and texture scores (left) and object renderings (right) of 21-foam-brick reconstructed with various reconstructed methods.}
   \label{fig:21_foam_brick}
\end{figure*}

The 3D reconstructions are displayed in Figure \ref{fig:01_master_chef_can}, \ref{fig:02_cracker_box},  \ref{fig:03_sugar_box}, \ref{fig:04_tomatoe_soup_can}, \ref{fig:05_mustard_bottle}, \ref{fig:06_tuna_fish_can}, \ref{fig:07_pudding_box}, \ref{fig:08_gelatin_box}, \ref{fig:09_potted_meat_can}, \ref{fig:10_banana}, \ref{fig:11_pitcher_base}, \ref{fig:12_bleach_cleanser}, \ref{fig:13_bowl}, \ref{fig:14_mug}, \ref{fig:15_power_drill}, \ref{fig:16_wood_block}, \ref{fig:17_scissors}, \ref{fig:18_large_marker}, \ref{fig:19_large_clamp}, \ref{fig:20_extra_large_clamp}, \ref{fig:21_foam_brick}. Each figure showcases the object's reconstructions, the pose performances and the texture quality for various reconstruction methods.

\clearpage

{\small
\bibliographystyle{ieee_fullname}
\bibliography{egbib}
}

%% file: combined.bbl
\begin{thebibliography}{100}\itemsep=-1pt

\bibitem{basler}
Basler ace camera aca2440-20gc.
\newblock https://docs.baslerweb.com/aca2440-20gc.

\bibitem{bopbenchmark}
Bop: Benchmark for 6d object pose estimation.
\newblock https://bop.felk.cvut.cz/home/.

\bibitem{boptoolkit}
Bop toolkit.
\newblock https://github.com/thodan/bop\_toolkit/.

\bibitem{moveitcalibration}
Moveit hand-eye calibration.
\newblock https://github.com/ros-planning/moveit\_calibrationc.

\bibitem{ycbbench}
Ycb benchmark.
\newblock https://www.ycbbenchmarks.com/object-set/.

\bibitem{lin2024sam}
Sam-6d: Segment anything model meets zero-shot 6d object pose estimation.
\newblock In {\em Proceedings of the IEEE/CVF Conference on Computer Vision and Pattern Recognition}, 2024.

\bibitem{arandjelovic2012three}
Relja Arandjelovi{\'c} and Andrew Zisserman.
\newblock Three things everyone should know to improve object retrieval.
\newblock In {\em 2012 IEEE conference on computer vision and pattern recognition}, pages 2911--2918. IEEE, 2012.

\bibitem{aravecchia2024comparing}
St{\'e}phanie Aravecchia, Marianne Clausel, and C{\'e}dric Pradalier.
\newblock Comparing metrics for evaluating 3d map quality in natural environments.
\newblock {\em Robotics and Autonomous Systems}, 173:104617, 2024.

\bibitem{besl1992method}
Paul~J Besl and Neil~D McKay.
\newblock Method for registration of 3-d shapes.
\newblock In {\em Sensor fusion IV: control paradigms and data structures}, volume 1611, pages 586--606. Spie, 1992.

\bibitem{bleyer2011patchmatch}
Michael Bleyer, Christoph Rhemann, and Carsten Rother.
\newblock Patchmatch stereo-stereo matching with slanted support windows.
\newblock In {\em Bmvc}, volume~11, pages 1--11, 2011.

\bibitem{brachmann2014learning}
Eric Brachmann, Alexander Krull, Frank Michel, Stefan Gumhold, Jamie Shotton, and Carsten Rother.
\newblock Learning 6d object pose estimation using 3d object coordinates.
\newblock In {\em Computer Vision--ECCV 2014: 13th European Conference, Zurich, Switzerland, September 6-12, 2014, Proceedings, Part II 13}, pages 536--551. Springer, 2014.

\bibitem{brachmann2016uncertainty}
Eric Brachmann, Frank Michel, Alexander Krull, Michael~Ying Yang, Stefan Gumhold, et~al.
\newblock Uncertainty-driven 6d pose estimation of objects and scenes from a single rgb image.
\newblock In {\em Proceedings of the IEEE conference on computer vision and pattern recognition}, pages 3364--3372, 2016.

\bibitem{bradski2000opencv}
Gary Bradski, Adrian Kaehler, et~al.
\newblock Opencv.
\newblock {\em Dr. Dobb’s journal of software tools}, 3(2), 2000.

\bibitem{cai2020reconstruct}
Ming Cai and Ian Reid.
\newblock Reconstruct locally, localize globally: A model free method for object pose estimation.
\newblock In {\em Proceedings of the IEEE/CVF Conference on Computer Vision and Pattern Recognition}, pages 3153--3163, 2020.

\bibitem{calli2015ycb}
Berk Calli, Arjun Singh, Aaron Walsman, Siddhartha Srinivasa, Pieter Abbeel, and Aaron~M Dollar.
\newblock The ycb object and model set: Towards common benchmarks for manipulation research.
\newblock In {\em 2015 international conference on advanced robotics (ICAR)}, pages 510--517. IEEE, 2015.

\bibitem{caraffa2025freeze}
Andrea Caraffa, Davide Boscaini, Amir Hamza, and Fabio Poiesi.
\newblock Freeze: Training-free zero-shot 6d pose estimation with geometric and vision foundation models.
\newblock In {\em European Conference on Computer Vision}, pages 414--431. Springer, 2024.

\bibitem{chen2023texpose}
Hanzhi Chen, Fabian Manhardt, Nassir Navab, and Benjamin Busam.
\newblock Texpose: Neural texture learning for self-supervised 6d object pose estimation.
\newblock In {\em Proceedings of the IEEE/CVF Conference on Computer Vision and Pattern Recognition}, pages 4841--4852, 2023.

\bibitem{chen2020category}
Xu Chen, Zijian Dong, Jie Song, Andreas Geiger, and Otmar Hilliges.
\newblock Category level object pose estimation via neural analysis-by-synthesis.
\newblock In {\em Computer Vision--ECCV 2020: 16th European Conference, Glasgow, UK, August 23--28, 2020, Proceedings, Part XXVI 16}, pages 139--156. Springer, 2020.

\bibitem{chitta2012moveit}
Sachin Chitta, Ioan Sucan, and Steve Cousins.
\newblock Moveit![ros topics].
\newblock {\em IEEE Robotics \& Automation Magazine}, 19(1):18--19, 2012.

\bibitem{LocalChapterEvents:ItalChap:ItalianChapConf2008:129-136}
Paolo Cignoni, Marco Callieri, Massimiliano Corsini, Matteo Dellepiane, Fabio Ganovelli, and Guido Ranzuglia.
\newblock {MeshLab: an Open-Source Mesh Processing Tool}.
\newblock In Vittorio Scarano, Rosario~De Chiara, and Ugo Erra, editors, {\em Eurographics Italian Chapter Conference}. The Eurographics Association, 2008.

\bibitem{collet2009object}
Alvaro Collet, Dmitry Berenson, Siddhartha~S Srinivasa, and Dave Ferguson.
\newblock Object recognition and full pose registration from a single image for robotic manipulation.
\newblock In {\em 2009 IEEE International Conference on Robotics and Automation}, pages 48--55. IEEE, 2009.

\bibitem{curless1996volumetric}
Brian Curless and Marc Levoy.
\newblock A volumetric method for building complex models from range images.
\newblock In {\em Proceedings of the 23rd annual conference on Computer graphics and interactive techniques}, pages 303--312, 1996.

\bibitem{trimesh}
{Dawson-Haggerty et al.}
\newblock trimesh.

\bibitem{doumanoglou2016recovering}
Andreas Doumanoglou, Rigas Kouskouridas, Sotiris Malassiotis, and Tae-Kyun Kim.
\newblock Recovering 6d object pose and predicting next-best-view in the crowd.
\newblock In {\em Proceedings of the IEEE conference on computer vision and pattern recognition}, pages 3583--3592, 2016.

\bibitem{downs2022google}
Laura Downs, Anthony Francis, Nate Koenig, Brandon Kinman, Ryan Hickman, Krista Reymann, Thomas~B McHugh, and Vincent Vanhoucke.
\newblock Google scanned objects: A high-quality dataset of 3d scanned household items.
\newblock In {\em 2022 International Conference on Robotics and Automation (ICRA)}, pages 2553--2560. IEEE, 2022.

\bibitem{drost2017introducing}
Bertram Drost, Markus Ulrich, Paul Bergmann, Philipp Hartinger, and Carsten Steger.
\newblock Introducing mvtec itodd-a dataset for 3d object recognition in industry.
\newblock In {\em Proceedings of the IEEE international conference on computer vision workshops}, pages 2200--2208, 2017.

\bibitem{drost2010model}
Bertram Drost, Markus Ulrich, Nassir Navab, and Slobodan Ilic.
\newblock Model globally, match locally: Efficient and robust 3d object recognition.
\newblock In {\em 2010 IEEE computer society conference on computer vision and pattern recognition}, pages 998--1005. Ieee, 2010.

\bibitem{faugeras1997level}
Olivier Faugeras and Renaud Keriven.
\newblock Level set methods and the stereo problem.
\newblock In {\em Scale-Space Theory in Computer Vision: First International Conference, Scale-Space'97 Utrecht, The Netherlands, July 2--4, 1997 Proceedings 1}, pages 272--283. Springer, 1997.

\bibitem{fridovich2022plenoxels}
Sara Fridovich-Keil, Alex Yu, Matthew Tancik, Qinhong Chen, Benjamin Recht, and Angjoo Kanazawa.
\newblock Plenoxels: Radiance fields without neural networks.
\newblock In {\em Proceedings of the IEEE/CVF Conference on Computer Vision and Pattern Recognition}, pages 5501--5510, 2022.

\bibitem{furukawa2006carved}
Yasutaka Furukawa and Jean Ponce.
\newblock Carved visual hulls for image-based modeling.
\newblock In {\em Computer Vision--ECCV 2006: 9th European Conference on Computer Vision, Graz, Austria, May 7-13, 2006. Proceedings, Part I 9}, pages 564--577. Springer, 2006.

\bibitem{furukawa2010accurate}
Y Furukawa and J Ponce.
\newblock Accurate, dense, and robust multiview stereopsis.
\newblock {\em IEEE Transactions on Pattern Analysis and Machine Intelligence}, 32(8):1362--1376, 2010.

\bibitem{gonzalez2010measurement}
{\'A}lvaro Gonz{\'a}lez.
\newblock Measurement of areas on a sphere using fibonacci and latitude--longitude lattices.
\newblock {\em Mathematical Geosciences}, 42:49--64, 2010.

\bibitem{gropp2020implicit}
Amos Gropp, Lior Yariv, Niv Haim, Matan Atzmon, and Yaron Lipman.
\newblock Implicit geometric regularization for learning shapes.
\newblock In {\em Proceedings of the 37th International Conference on Machine Learning}, pages 3789--3799, 2020.

\bibitem{haugaard2022surfemb}
Rasmus~Laurvig Haugaard and Anders~Glent Buch.
\newblock Surfemb: Dense and continuous correspondence distributions for object pose estimation with learnt surface embeddings.
\newblock In {\em Proceedings of the IEEE/CVF Conference on Computer Vision and Pattern Recognition}, pages 6749--6758, 2022.

\bibitem{he2022onepose++}
Xingyi He, Jiaming Sun, Yuang Wang, Di Huang, Hujun Bao, and Xiaowei Zhou.
\newblock Onepose++: Keypoint-free one-shot object pose estimation without cad models.
\newblock {\em Advances in Neural Information Processing Systems}, 35:35103--35115, 2022.

\bibitem{he2022fs6d}
Yisheng He, Yao Wang, Haoqiang Fan, Jian Sun, and Qifeng Chen.
\newblock Fs6d: Few-shot 6d pose estimation of novel objects.
\newblock In {\em Proceedings of the IEEE/CVF Conference on Computer Vision and Pattern Recognition}, pages 6814--6824, 2022.

\bibitem{hennersperger2017towards}
Christoph Hennersperger, Bernhard Fuerst, Salvatore Virga, Oliver Zettinig, Benjamin Frisch, Thomas Neff, and Nassir Navab.
\newblock Towards mri-based autonomous robotic us acquisitions: a first feasibility study.
\newblock {\em IEEE transactions on medical imaging}, 36(2):538--548, 2017.

\bibitem{hinterstoisser2011gradient}
Stefan Hinterstoisser, Cedric Cagniart, Slobodan Ilic, Peter Sturm, Nassir Navab, Pascal Fua, and Vincent Lepetit.
\newblock Gradient response maps for real-time detection of textureless objects.
\newblock {\em IEEE transactions on pattern analysis and machine intelligence}, 34(5):876--888, 2011.

\bibitem{hinterstoisser2011multimodal}
Stefan Hinterstoisser, Stefan Holzer, Cedric Cagniart, Slobodan Ilic, Kurt Konolige, Nassir Navab, and Vincent Lepetit.
\newblock Multimodal templates for real-time detection of texture-less objects in heavily cluttered scenes.
\newblock In {\em 2011 international conference on computer vision}, pages 858--865. IEEE, 2011.

\bibitem{hinterstoisser2013model}
Stefan Hinterstoisser, Vincent Lepetit, Slobodan Ilic, Stefan Holzer, Gary Bradski, Kurt Konolige, and Nassir Navab.
\newblock Model based training, detection and pose estimation of texture-less 3d objects in heavily cluttered scenes.
\newblock In {\em Computer Vision--ACCV 2012: 11th Asian Conference on Computer Vision, Daejeon, Korea, November 5-9, 2012, Revised Selected Papers, Part I 11}, pages 548--562. Springer, 2013.

\bibitem{hinterstoisser2016going}
Stefan Hinterstoisser, Vincent Lepetit, Naresh Rajkumar, and Kurt Konolige.
\newblock Going further with point pair features.
\newblock In {\em Computer Vision--ECCV 2016: 14th European Conference, Amsterdam, The Netherlands, October 11-14, 2016, Proceedings, Part III 14}, pages 834--848. Springer, 2016.

\bibitem{hirschmuller2007evaluation}
Heiko Hirschmuller and Daniel Scharstein.
\newblock Evaluation of cost functions for stereo matching.
\newblock In {\em 2007 IEEE conference on computer vision and pattern recognition}, pages 1--8. IEEE, 2007.

\bibitem{hodan2020epos}
Tomas Hodan, Daniel Barath, and Jiri Matas.
\newblock Epos: Estimating 6d pose of objects with symmetries.
\newblock In {\em Proceedings of the IEEE/CVF conference on computer vision and pattern recognition}, pages 11703--11712, 2020.

\bibitem{hodan2017t}
Tom{\'a}{\v{s}} Hodan, Pavel Haluza, {\v{S}}tep{\'a}n Obdr{\v{z}}{\'a}lek, Jiri Matas, Manolis Lourakis, and Xenophon Zabulis.
\newblock T-less: An rgb-d dataset for 6d pose estimation of texture-less objects.
\newblock In {\em 2017 IEEE Winter Conference on Applications of Computer Vision (WACV)}, pages 880--888. IEEE, 2017.

\bibitem{hodan2018bop}
Tomas Hodan, Frank Michel, Eric Brachmann, Wadim Kehl, Anders GlentBuch, Dirk Kraft, Bertram Drost, Joel Vidal, Stephan Ihrke, Xenophon Zabulis, et~al.
\newblock Bop: Benchmark for 6d object pose estimation.
\newblock In {\em Proceedings of the European conference on computer vision (ECCV)}, pages 19--34, 2018.

\bibitem{hodavn2020bop}
Tom{\'a}{\v{s}} Hoda{\v{n}}, Martin Sundermeyer, Bertram Drost, Yann Labb{\'e}, Eric Brachmann, Frank Michel, Carsten Rother, and Ji{\v{r}}{\'\i} Matas.
\newblock Bop challenge 2020 on 6d object localization.
\newblock In {\em Computer Vision--ECCV 2020 Workshops: Glasgow, UK, August 23--28, 2020, Proceedings, Part II 16}, pages 577--594. Springer, 2020.

\bibitem{jensen2014large}
Rasmus Jensen, Anders Dahl, George Vogiatzis, Engin Tola, and Henrik Aan{\ae}s.
\newblock Large scale multi-view stereopsis evaluation.
\newblock In {\em Proceedings of the IEEE conference on computer vision and pattern recognition}, pages 406--413, 2014.

\bibitem{kajiya1984ray}
James~T Kajiya and Brian~P Von~Herzen.
\newblock Ray tracing volume densities.
\newblock {\em ACM SIGGRAPH computer graphics}, 18(3):165--174, 1984.

\bibitem{kanade1995development}
Takeo Kanade, Hiroshi Kano, Shigeru Kimura, Atsushi Yoshida, and Kazuo Oda.
\newblock Development of a video-rate stereo machine.
\newblock In {\em Proceedings 1995 IEEE/RSJ International Conference on Intelligent Robots and Systems. Human Robot Interaction and Cooperative Robots}, volume~3, pages 95--100. IEEE, 1995.

\bibitem{kaskman2019homebreweddb}
Roman Kaskman, Sergey Zakharov, Ivan Shugurov, and Slobodan Ilic.
\newblock Homebreweddb: Rgb-d dataset for 6d pose estimation of 3d objects.
\newblock In {\em Proceedings of the IEEE/CVF International Conference on Computer Vision Workshops}, pages 0--0, 2019.

\bibitem{kazhdan2013screened}
Michael Kazhdan and Hugues Hoppe.
\newblock Screened poisson surface reconstruction.
\newblock {\em ACM Transactions on Graphics (ToG)}, 32(3):1--13, 2013.

\bibitem{knapitsch2017tanks}
Arno Knapitsch, Jaesik Park, Qian-Yi Zhou, and Vladlen Koltun.
\newblock Tanks and temples: Benchmarking large-scale scene reconstruction.
\newblock {\em ACM Transactions on Graphics (ToG)}, 36(4):1--13, 2017.

\bibitem{kutulakos2000theory}
Kiriakos~N Kutulakos and Steven~M Seitz.
\newblock A theory of shape by space carving.
\newblock {\em International journal of computer vision}, 38:199--218, 2000.

\bibitem{labbe2020cosypose}
Yann Labb{\'e}, Justin Carpentier, Mathieu Aubry, and Josef Sivic.
\newblock Cosypose: Consistent multi-view multi-object 6d pose estimation.
\newblock In {\em Computer Vision--ECCV 2020: 16th European Conference, Glasgow, UK, August 23--28, 2020, Proceedings, Part XVII 16}, pages 574--591. Springer, 2020.

\bibitem{labbe2022megapose}
Yann Labb\'e, Lucas Manuelli, Arsalan Mousavian, Stephen Tyree, Stan Birchfield, Jonathan Tremblay, Justin Carpentier, Mathieu Aubry, Dieter Fox, and Josef Sivic.
\newblock {{MegaPose}}: {{6D Pose Estimation}} of {{Novel Objects}} via {{Render}} \& {{Compare}}.
\newblock In {\em CoRL}.

\bibitem{lhuillier2005quasi}
Maxime Lhuillier and Long Quan.
\newblock A quasi-dense approach to surface reconstruction from uncalibrated images.
\newblock {\em IEEE transactions on pattern analysis and machine intelligence}, 27(3):418--433, 2005.

\bibitem{li2018deepim}
Yi Li, Gu Wang, Xiangyang Ji, Yu Xiang, and Dieter Fox.
\newblock Deepim: Deep iterative matching for 6d pose estimation.
\newblock In {\em Proceedings of the European Conference on Computer Vision (ECCV)}, pages 683--698, 2018.

\bibitem{li2023neuralangelo}
Zhaoshuo Li, Thomas M{\"u}ller, Alex Evans, Russell~H Taylor, Mathias Unberath, Ming-Yu Liu, and Chen-Hsuan Lin.
\newblock Neuralangelo: High-fidelity neural surface reconstruction.
\newblock In {\em Proceedings of the IEEE/CVF Conference on Computer Vision and Pattern Recognition}, pages 8456--8465, 2023.

\bibitem{li2019cdpn}
Zhigang Li, Gu Wang, and Xiangyang Ji.
\newblock Cdpn: Coordinates-based disentangled pose network for real-time rgb-based 6-dof object pose estimation.
\newblock In {\em Proceedings of the IEEE/CVF International Conference on Computer Vision}, pages 7678--7687, 2019.

\bibitem{lin2022single}
Yunzhi Lin, Jonathan Tremblay, Stephen Tyree, Patricio~A Vela, and Stan Birchfield.
\newblock Single-stage keypoint-based category-level object pose estimation from an rgb image.
\newblock In {\em 2022 International Conference on Robotics and Automation (ICRA)}, pages 1547--1553. IEEE, 2022.

\bibitem{liu2022gen6d}
Yuan Liu, Yilin Wen, Sida Peng, Cheng Lin, Xiaoxiao Long, Taku Komura, and Wenping Wang.
\newblock Gen6d: Generalizable model-free 6-dof object pose estimation from rgb images.
\newblock In {\em European Conference on Computer Vision}, pages 298--315. Springer, 2022.

\bibitem{lombardi2019neural}
Stephen Lombardi, Tomas Simon, Jason Saragih, Gabriel Schwartz, Andreas Lehrmann, and Yaser Sheikh.
\newblock Neural volumes: learning dynamic renderable volumes from images.
\newblock {\em ACM Transactions on Graphics (TOG)}, 38(4):1--14, 2019.

\bibitem{long2024wonder3d}
Xiaoxiao Long, Yuan-Chen Guo, Cheng Lin, Yuan Liu, Zhiyang Dou, Lingjie Liu, Yuexin Ma, Song-Hai Zhang, Marc Habermann, Christian Theobalt, et~al.
\newblock Wonder3d: Single image to 3d using cross-domain diffusion.
\newblock In {\em Proceedings of the IEEE/CVF Conference on Computer Vision and Pattern Recognition}, pages 9970--9980, 2024.

\bibitem{lorensen1987marching}
William~E Lorensen and Harvey~E Cline.
\newblock Marching cubes: A high resolution 3d surface construction algorithm.
\newblock {\em ACM SIGGRAPH Computer Graphics}, 21(4):163--169, 1987.

\bibitem{Lowe04IJCV}
David~G. Lowe.
\newblock {Distinctive Image Features from Scale-Invariant Keypoints}.
\newblock {\em IJCV}, 2004.

\bibitem{manuelli2019kpam}
Lucas Manuelli, Wei Gao, Peter Florence, and Russ Tedrake.
\newblock kpam: Keypoint affordances for category-level robotic manipulation.
\newblock In {\em The International Symposium of Robotics Research}, pages 132--157. Springer, 2019.

\bibitem{martinez2010moped}
Manuel Martinez, Alvaro Collet, and Siddhartha~S Srinivasa.
\newblock Moped: A scalable and low latency object recognition and pose estimation system.
\newblock In {\em 2010 IEEE International Conference on Robotics and Automation}, pages 2043--2049. IEEE, 2010.

\bibitem{mescheder2019occupancy}
Lars Mescheder, Michael Oechsle, Michael Niemeyer, Sebastian Nowozin, and Andreas Geiger.
\newblock Occupancy networks: Learning 3d reconstruction in function space.
\newblock In {\em Proceedings of the IEEE/CVF conference on computer vision and pattern recognition}, pages 4460--4470, 2019.

\bibitem{mildenhall2020nerf}
B Mildenhall, PP Srinivasan, M Tancik, JT Barron, R Ramamoorthi, and R Ng.
\newblock Nerf: Representing scenes as neural radiance fields for view synthesis.
\newblock In {\em European conference on computer vision}, 2020.

\bibitem{moon2024genflow}
Sungphill Moon, Hyeontae Son, Dongcheol Hur, and Sangwook Kim.
\newblock Genflow: Generalizable recurrent flow for 6d pose refinement of novel objects.
\newblock In {\em Proceedings of the IEEE/CVF Conference on Computer Vision and Pattern Recognition}, 2024.

\bibitem{muller2022instant}
Thomas M{\"u}ller, Alex Evans, Christoph Schied, and Alexander Keller.
\newblock Instant neural graphics primitives with a multiresolution hash encoding.
\newblock {\em ACM Transactions on Graphics (ToG)}, 41(4):1--15, 2022.

\bibitem{newcombe2011kinectfusion}
Richard~A Newcombe, Shahram Izadi, Otmar Hilliges, David Molyneaux, David Kim, Andrew~J Davison, Pushmeet Kohi, Jamie Shotton, Steve Hodges, and Andrew Fitzgibbon.
\newblock Kinectfusion: Real-time dense surface mapping and tracking.
\newblock In {\em 2011 10th IEEE international symposium on mixed and augmented reality}, pages 127--136. IEEE, 2011.

\bibitem{nguyen2024gigaPose}
Van~Nguyen Nguyen, Thibault Groueix, Mathieu Salzmann, and Vincent Lepetit.
\newblock Gigapose: Fast and robust novel object pose estimation via one correspondence.
\newblock In {\em Proceedings of the IEEE/CVF Conference on Computer Vision and Pattern Recognition}, 2024.

\bibitem{niemeyer2020differentiable}
Michael Niemeyer, Lars Mescheder, Michael Oechsle, and Andreas Geiger.
\newblock Differentiable volumetric rendering: Learning implicit 3d representations without 3d supervision.
\newblock In {\em Proceedings of the IEEE/CVF Conference on Computer Vision and Pattern Recognition}, pages 3504--3515, 2020.

\bibitem{oechsle2021unisurf}
Michael Oechsle, Songyou Peng, and Andreas Geiger.
\newblock Unisurf: Unifying neural implicit surfaces and radiance fields for multi-view reconstruction.
\newblock In {\em Proceedings of the IEEE/CVF International Conference on Computer Vision}, pages 5589--5599, 2021.

\bibitem{ornek2023foundpose}
Evin~P{\i}nar {\"O}rnek, Yann Labb{\'e}, Bugra Tekin, Lingni Ma, Cem Keskin, Christian Forster, and Tomas Hodan.
\newblock Foundpose: Unseen object pose estimation with foundation features.
\newblock In {\em Proceedings of the IEEE/CVF Conference on Computer Vision and Pattern Recognition}, 2024.

\bibitem{park2019deepsdf}
Jeong~Joon Park, Peter Florence, Julian Straub, Richard Newcombe, and Steven Lovegrove.
\newblock Deepsdf: Learning continuous signed distance functions for shape representation.
\newblock In {\em Proceedings of the IEEE/CVF conference on computer vision and pattern recognition}, pages 165--174, 2019.

\bibitem{park2020latentfusion}
Keunhong Park, Arsalan Mousavian, Yu Xiang, and Dieter Fox.
\newblock Latentfusion: End-to-end differentiable reconstruction and rendering for unseen object pose estimation.
\newblock In {\em Proceedings of the IEEE/CVF conference on computer vision and pattern recognition}, pages 10710--10719, 2020.

\bibitem{rad2017bb8}
Mahdi Rad and Vincent Lepetit.
\newblock Bb8: A scalable, accurate, robust to partial occlusion method for predicting the 3d poses of challenging objects without using depth.
\newblock In {\em Proceedings of the IEEE international conference on computer vision}, pages 3828--3836, 2017.

\bibitem{ratliff2009chomp}
Nathan Ratliff, Matt Zucker, J~Andrew Bagnell, and Siddhartha Srinivasa.
\newblock Chomp: Gradient optimization techniques for efficient motion planning.
\newblock In {\em 2009 IEEE international conference on robotics and automation}, pages 489--494. IEEE, 2009.

\bibitem{ravi2020pytorch3d}
Nikhila Ravi, Jeremy Reizenstein, David Novotny, Taylor Gordon, Wan-Yen Lo, Justin Johnson, and Georgia Gkioxari.
\newblock Accelerating 3d deep learning with pytorch3d.
\newblock {\em arXiv:2007.08501}, 2020.

\bibitem{capturereality}
RealityCapture2023.
\newblock {RealityCapture}, 4 2023.

\bibitem{rennie2016dataset}
Colin Rennie, Rahul Shome, Kostas~E Bekris, and Alberto~F De~Souza.
\newblock A dataset for improved rgbd-based object detection and pose estimation for warehouse pick-and-place.
\newblock {\em IEEE Robotics and Automation Letters}, 1(2):1179--1185, 2016.

\bibitem{rothganger20063d}
Fred Rothganger, Svetlana Lazebnik, Cordelia Schmid, and Jean Ponce.
\newblock 3d object modeling and recognition using local affine-invariant image descriptors and multi-view spatial constraints.
\newblock {\em International journal of computer vision}, 66:231--259, 2006.

\bibitem{scharstein2014high}
Daniel Scharstein, Heiko Hirschm{\"u}ller, York Kitajima, Greg Krathwohl, Nera Ne{\v{s}}i{\'c}, Xi Wang, and Porter Westling.
\newblock High-resolution stereo datasets with subpixel-accurate ground truth.
\newblock In {\em Pattern Recognition: 36th German Conference, GCPR 2014, M{\"u}nster, Germany, September 2-5, 2014, Proceedings 36}, pages 31--42. Springer, 2014.

\bibitem{scharstein2002taxonomy}
Daniel Scharstein and Richard Szeliski.
\newblock A taxonomy and evaluation of dense two-frame stereo correspondence algorithms.
\newblock {\em International journal of computer vision}, 47:7--42, 2002.

\bibitem{schonberger2016structure}
Johannes~L Schonberger and Jan-Michael Frahm.
\newblock Structure-from-motion revisited.
\newblock In {\em Proceedings of the IEEE conference on computer vision and pattern recognition}, pages 4104--4113, 2016.

\bibitem{schonberger2016pixelwise}
Johannes~L Sch{\"o}nberger, Enliang Zheng, Jan-Michael Frahm, and Marc Pollefeys.
\newblock Pixelwise view selection for unstructured multi-view stereo.
\newblock In {\em Computer Vision--ECCV 2016: 14th European Conference, Amsterdam, The Netherlands, October 11-14, 2016, Proceedings, Part III 14}, pages 501--518. Springer, 2016.

\bibitem{ETH3D_bench}
Thomas Sch\"ops, Torsten Sattler, and Marc Pollefeys.
\newblock {BAD SLAM}: Bundle adjusted direct {RGB-D SLAM}.
\newblock In {\em Conference on Computer Vision and Pattern Recognition (CVPR)}, 2019.

\bibitem{schops2017multi}
Thomas Schops, Johannes~L Schonberger, Silvano Galliani, Torsten Sattler, Konrad Schindler, Marc Pollefeys, and Andreas Geiger.
\newblock A multi-view stereo benchmark with high-resolution images and multi-camera videos.
\newblock In {\em Proceedings of the IEEE conference on computer vision and pattern recognition}, pages 3260--3269, 2017.

\bibitem{seitz2006comparison}
Steven~M Seitz, Brian Curless, James Diebel, Daniel Scharstein, and Richard Szeliski.
\newblock A comparison and evaluation of multi-view stereo reconstruction algorithms.
\newblock In {\em 2006 IEEE computer society conference on computer vision and pattern recognition (CVPR'06)}, volume~1, pages 519--528. IEEE, 2006.

\bibitem{seitz1999photorealistic}
Steven~M Seitz and Charles~R Dyer.
\newblock Photorealistic scene reconstruction by voxel coloring.
\newblock {\em International Journal of Computer Vision}, 35:151--173, 1999.

\bibitem{shugurov2022osop}
Ivan Shugurov, Fu Li, Benjamin Busam, and Slobodan Ilic.
\newblock Osop: A multi-stage one shot object pose estimation framework.
\newblock In {\em Proceedings of the IEEE/CVF Conference on Computer Vision and Pattern Recognition}, pages 6835--6844, 2022.

\bibitem{sinha2007multi}
Sudipta~N Sinha, Philippos Mordohai, and Marc Pollefeys.
\newblock Multi-view stereo via graph cuts on the dual of an adaptive tetrahedral mesh.
\newblock In {\em 2007 IEEE 11th international conference on computer vision}, pages 1--8. IEEE, 2007.

\bibitem{sinha2004visual}
Sudipta~N Sinha and Marc Pollefeys.
\newblock Visual-hull reconstruction from uncalibrated and unsynchronized video streams.
\newblock In {\em Proceedings. 2nd International Symposium on 3D Data Processing, Visualization and Transmission, 2004. 3DPVT 2004.}, pages 349--356. IEEE, 2004.

\bibitem{sitzmann2019scene}
Vincent Sitzmann, Michael Zollh{\"o}fer, and Gordon Wetzstein.
\newblock Scene representation networks: Continuous 3d-structure-aware neural scene representations.
\newblock {\em Advances in Neural Information Processing Systems}, 32, 2019.

\bibitem{strecha2006combined}
Christoph Strecha, Rik Fransens, and Luc Van~Gool.
\newblock Combined depth and outlier estimation in multi-view stereo.
\newblock In {\em 2006 IEEE Computer Society Conference on Computer Vision and Pattern Recognition (CVPR'06)}, volume~2, pages 2394--2401. IEEE, 2006.

\bibitem{strecha2008benchmarking}
Christoph Strecha, Wolfgang Von~Hansen, Luc Van~Gool, Pascal Fua, and Ulrich Thoennessen.
\newblock On benchmarking camera calibration and multi-view stereo for high resolution imagery.
\newblock In {\em 2008 IEEE conference on computer vision and pattern recognition}, pages 1--8. Ieee, 2008.

\bibitem{sun2022onepose}
Jiaming Sun, Zihao Wang, Siyu Zhang, Xingyi He, Hongcheng Zhao, Guofeng Zhang, and Xiaowei Zhou.
\newblock Onepose: One-shot object pose estimation without cad models.
\newblock In {\em Proceedings of the IEEE/CVF Conference on Computer Vision and Pattern Recognition}, pages 6825--6834, 2022.

\bibitem{nerfstudio}
Matthew Tancik, Ethan Weber, Evonne Ng, Ruilong Li, Brent Yi, Justin Kerr, Terrance Wang, Alexander Kristoffersen, Jake Austin, Kamyar Salahi, Abhik Ahuja, David McAllister, and Angjoo Kanazawa.
\newblock Nerfstudio: A modular framework for neural radiance field development.
\newblock In {\em ACM SIGGRAPH 2023 Conference Proceedings}, SIGGRAPH '23, 2023.

\bibitem{tejani2014latent}
Alykhan Tejani, Danhang Tang, Rigas Kouskouridas, and Tae-Kyun Kim.
\newblock Latent-class hough forests for 3d object detection and pose estimation.
\newblock In {\em Computer Vision--ECCV 2014: 13th European Conference, Zurich, Switzerland, September 6-12, 2014, Proceedings, Part VI 13}, pages 462--477. Springer, 2014.

\bibitem{tola2012efficient}
Engin Tola, Christoph Strecha, and Pascal Fua.
\newblock Efficient large-scale multi-view stereo for ultra high-resolution image sets.
\newblock {\em Machine Vision and Applications}, 23:903--920, 2012.

\bibitem{tremblay2018deep}
Jonathan Tremblay, Thang To, Balakumar Sundaralingam, Yu Xiang, Dieter Fox, and Stan Birchfield.
\newblock Deep object pose estimation for semantic robotic grasping of household objects.
\newblock In {\em Conference on Robot Learning}, pages 306--316. PMLR, 2018.

\bibitem{tyree20226}
Stephen Tyree, Jonathan Tremblay, Thang To, Jia Cheng, Terry Mosier, Jeffrey Smith, and Stan Birchfield.
\newblock 6-dof pose estimation of household objects for robotic manipulation: An accessible dataset and benchmark.
\newblock In {\em 2022 IEEE/RSJ International Conference on Intelligent Robots and Systems (IROS)}, pages 13081--13088. IEEE, 2022.

\bibitem{ulusoy2017semantic}
Ali~Osman Ulusoy, Michael~J Black, and Andreas Geiger.
\newblock Semantic multi-view stereo: Jointly estimating objects and voxels.
\newblock In {\em 2017 IEEE Conference on Computer Vision and Pattern Recognition (CVPR)}, pages 4531--4540. IEEE, 2017.

\bibitem{Waechter2014Texturing}
Michael Waechter, Nils Moehrle, and Michael Goesele.
\newblock Let there be color! --- {L}arge-scale texturing of {3D} reconstructions.
\newblock In {\em Proceedings of the European Conference on Computer Vision}. Springer, 2014.

\bibitem{wang2021gdr}
Gu Wang, Fabian Manhardt, Federico Tombari, and Xiangyang Ji.
\newblock Gdr-net: Geometry-guided direct regression network for monocular 6d object pose estimation.
\newblock In {\em Proceedings of the IEEE/CVF Conference on Computer Vision and Pattern Recognition}, pages 16611--16621, 2021.

\bibitem{wang2019normalized}
He Wang, Srinath Sridhar, Jingwei Huang, Julien Valentin, Shuran Song, and Leonidas~J Guibas.
\newblock Normalized object coordinate space for category-level 6d object pose and size estimation.
\newblock In {\em Proceedings of the IEEE/CVF Conference on Computer Vision and Pattern Recognition}, pages 2642--2651, 2019.

\bibitem{wang2021neus}
Peng Wang, Lingjie Liu, Yuan Liu, Christian Theobalt, Taku Komura, and Wenping Wang.
\newblock Neus: Learning neural implicit surfaces by volume rendering for multi-view reconstruction.
\newblock {\em Advances in Neural Information Processing Systems}, 34:27171--27183, 2021.

\bibitem{foundationposewen2024}
Bowen Wen, Wei Yang, Jan Kautz, and Stan Birchfield.
\newblock {FoundationPose}: Unified 6d pose estimation and tracking of novel objects.
\newblock In {\em CVPR}, 2024.

\bibitem{wohlhart2015learning}
Paul Wohlhart and Vincent Lepetit.
\newblock Learning descriptors for object recognition and 3d pose estimation.
\newblock In {\em Proceedings of the IEEE conference on computer vision and pattern recognition}, pages 3109--3118, 2015.

\bibitem{xiang2018posecnn}
Yu Xiang, Tanner Schmidt, Venkatraman Narayanan, and Dieter Fox.
\newblock Posecnn: A convolutional neural network for 6d object pose estimation in cluttered scenes.
\newblock 2018.

\bibitem{yariv2021volume}
Lior Yariv, Jiatao Gu, Yoni Kasten, and Yaron Lipman.
\newblock Volume rendering of neural implicit surfaces.
\newblock {\em Advances in Neural Information Processing Systems}, 34:4805--4815, 2021.

\bibitem{yariv2023bakedsdf}
Lior Yariv, Peter Hedman, Christian Reiser, Dor Verbin, Pratul~P Srinivasan, Richard Szeliski, Jonathan~T Barron, and Ben Mildenhall.
\newblock Bakedsdf: Meshing neural sdfs for real-time view synthesis.
\newblock {\em ACM Transactions on Graphics}, 2023.

\bibitem{yariv2020multiview}
Lior Yariv, Yoni Kasten, Dror Moran, Meirav Galun, Matan Atzmon, Basri Ronen, and Yaron Lipman.
\newblock Multiview neural surface reconstruction by disentangling geometry and appearance.
\newblock {\em Advances in Neural Information Processing Systems}, 33:2492--2502, 2020.

\bibitem{Yu2022SDFStudio}
Zehao Yu, Anpei Chen, Bozidar Antic, Songyou Peng, Apratim Bhattacharyya, Michael Niemeyer, Siyu Tang, Torsten Sattler, and Andreas Geiger.
\newblock Sdfstudio: A unified framework for surface reconstruction, 2022.

\bibitem{yu2022monosdf}
Zehao Yu, Songyou Peng, Michael Niemeyer, Torsten Sattler, and Andreas Geiger.
\newblock Monosdf: Exploring monocular geometric cues for neural implicit surface reconstruction.
\newblock {\em Advances in neural information processing systems}, 35:25018--25032, 2022.

\bibitem{zhang2018perceptual}
Richard Zhang, Phillip Isola, Alexei~A Efros, Eli Shechtman, and Oliver Wang.
\newblock The unreasonable effectiveness of deep features as a perceptual metric.
\newblock In {\em CVPR}, 2018.

\bibitem{Zhou2018}
Qian-Yi Zhou, Jaesik Park, and Vladlen Koltun.
\newblock {Open3D}: {A} modern library for {3D} data processing.
\newblock {\em arXiv:1801.09847}, 2018.

\end{thebibliography}
